%% file: main.tex
\documentclass[11pt]{article}

\usepackage{naacl2021}

\usepackage{times}
\usepackage{latexsym}

\usepackage[T1]{fontenc}

\usepackage[utf8]{inputenc}

\usepackage{microtype}

\usepackage{amsmath}
\usepackage[all=normal, bibbreaks=normal, paragraphs=tight, floats=tight, mathspacing=normal, wordspacing=normal, tracking=normal]{savetrees}
\usepackage{nameref, hyperref}

\DeclareMathOperator*{\pmi}{pmi}
\DeclareMathOperator*{\pcmi}{pcmi}
\DeclareMathOperator*{\pmibf}{\textbf{pmi}}
\DeclareMathOperator*{\pcmibf}{\textbf{pcmi}}
\usepackage{tabularx, booktabs}
\usepackage{footmisc}

\makeatletter
\renewcommand{\paragraph}{%
    \@startsection {paragraph}{4}{\z@ }{0.5ex plus 0.5ex minus .1ex}{-0.5em}{\normalsize \bf }
}
\makeatother
\usepackage[disable]{todonotes}

\newcommand{\ashwin}[1]{\todo[color=teal!40]{\tiny Ashwin: #1\par}}
\newcommand{\kaitlyn}[1]{\todo[color=purple!40]{\tiny Kaitlyn: #1\par}} 

\newcommand{\kw}[1]{\textbf{\textit{#1}}}
\newcommand{\maxpmi}{Max-PMI}

\usepackage{adjustbox}
\newcolumntype{R}[2]{%
    >{\adjustbox{angle=#1,lap=\width-(#2)}\bgroup}%
    l%
    <{\egroup}%
}

\title{Human-like informative conversations: Better acknowledgements using conditional mutual information}

\author{Ashwin Paranjape \\
  Stanford University \\
  \texttt{ashwinp@cs.stanford.edu} \\\And
  Christopher D. Manning \\
  Stanford University \\
  \texttt{manning@cs.stanford.edu} \\}

\begin{document}
\maketitle
\begin{abstract}
This work aims to build a dialogue agent that can weave new factual content into conversations as naturally as humans.
    We draw insights from linguistic principles of conversational analysis and annotate human-human conversations from the Switchboard Dialog Act Corpus to examine humans strategies for \textit{acknowledgement}, \textit{transition}, \textit{detail selection} and \textit{presentation}. 
    When current chatbots (explicitly provided with new factual content) introduce facts into a conversation, their generated responses do not \textbf{\textit{acknowledge}} the prior turns. 
This is because models trained with two contexts -- new factual content and conversational history -- generate responses that are non-specific w.r.t.\ one of the contexts, typically the conversational history.
We show that specificity w.r.t.\ conversational history is better captured by \textit{pointwise \textbf{conditional} mutual information} ($\pcmi_h$) than by the established use of \textit{pointwise mutual information} ($\pmi$). 
    Our proposed method, Fused-PCMI, trades off $\pmi$ for $\pcmi_h$ and is preferred by humans for overall quality over the {\maxpmi} baseline 60\% of the time. 
    Human evaluators also judge responses with higher $\pcmi_h$ better at acknowledgement 74\% of the time.
    The results demonstrate that systems mimicking human conversational traits (in this case acknowledgement) improve overall quality and more broadly illustrate the utility of linguistic principles in improving dialogue agents. 

\end{abstract}

\section{Introduction}
Social chatbots are improving in appeal and are being deployed widely to converse with humans \cite{ap2020}.
Advances in neural generation \cite{adiwardana2020towards, roller2020recipes} enable them to handle a wide variety of user turns and to provide fluent bot responses. 
People expect their interactions with these dialogue agents to be similar to real social relationships \citep{ReevesNass1996}.
In particular, they expect social chatbots to both use information that is already known and separately add new information to the conversation, in line with the \textit{given-new} contract \cite{clark1977discourse}.

\begin{figure}
    \centering
        \includegraphics[width=\columnwidth]{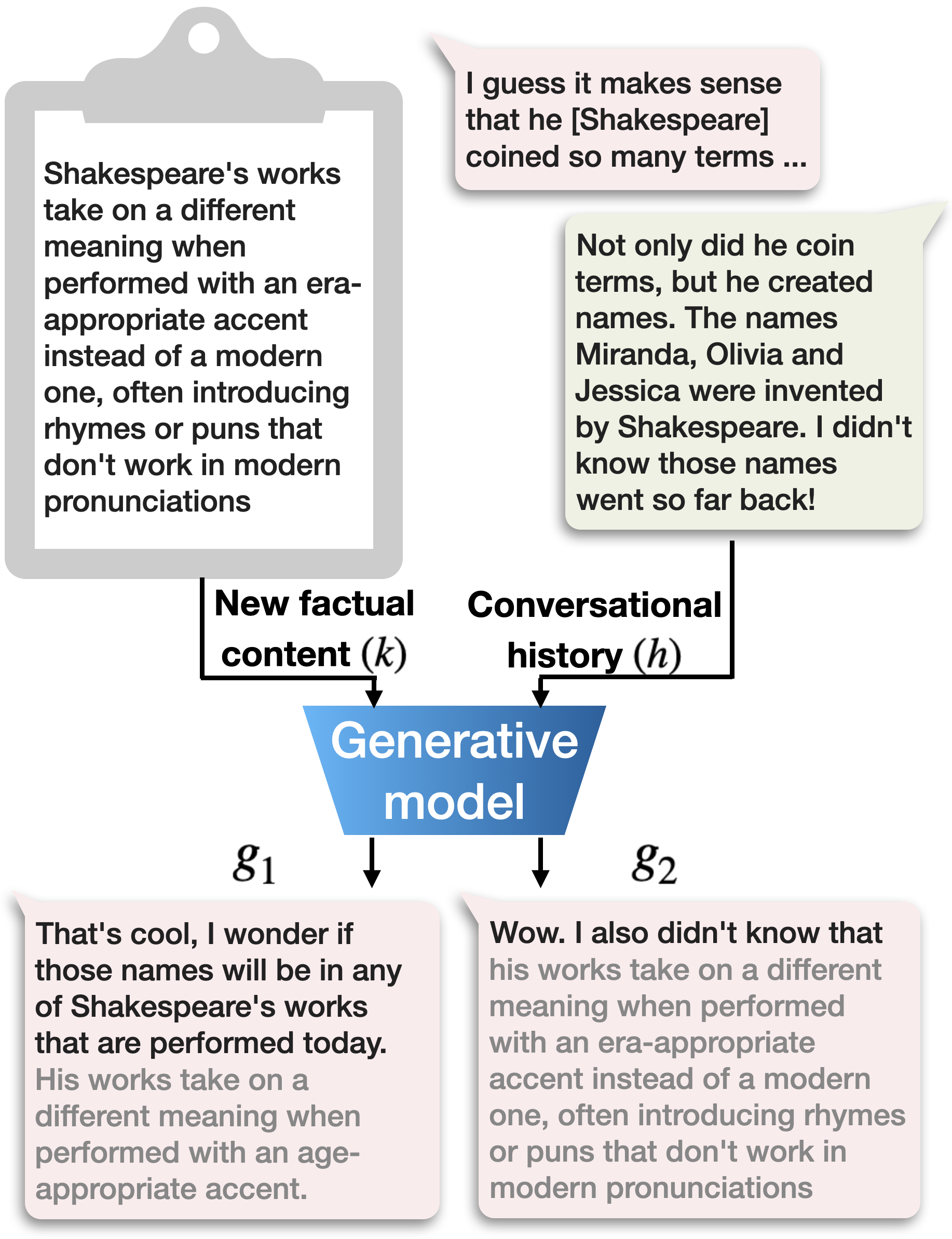}
    \caption{\textbf{The setting for conversational rephrasing.} Conversational history ($\textbf{h}$) and new factual content ($\textbf{k}$), two largely independent contexts, are used to sample responses ($\textbf{g}_1$, $\textbf{g}_2$) from a generative model. The samples differ qualitatively. While almost all of $\textbf{g}_2$ is verbatim from $\textbf{k}$ (in gray), the first sentence in $\textbf{g}_1$ (in black) acknowledges using $\textbf{h}$ and bridges to $\textbf{k}$.}
    \label{fig:convpara_example}
\end{figure}

Neural generation methods for adding new information \cite{dinan2018wizard, Gopalakrishnan2019,ghazvininejad2018knowledge,zhang-etal-2018-personalizing} measure progress using metrics like ``engagingness'', ``appropriateness'' and ``informativeness''.
But these metrics are too broad and provide little actionable insight to drive improvements in these systems.  
On the other hand, psycholinguists and sociolinguists have studied human conversations in depth and have identified fine-grained conventions, principles and contracts \cite{grice1975logic, clark_1996, krauss1996social}.

\textbf{Our first contribution is a linguistic analysis of how human conversations incorporate world knowledge.}
We manually annotate conversations from the Switchboard corpus to identify key traits.
In particular, we find that people apply four kinds of strategies: (1) \textbf{acknowledgement} of each other's utterances, (2) \textbf{transition} to new information, (3) appropriate level of \textbf{detail selection} and (4) \textbf{presentation} of factual content in forms such as opinions or experiences. 

To identify deficiencies of the above types in machine-learned models, we consider a simplified task of \textbf{conversational rephrasing} (Figure~\ref{fig:convpara_example}), in which the factual content to be added is not left latent but is provided as a text input to the model (as in \citet{dinan2018wizard}), along with conversational history.
Just as humans do not recite a fact verbatim in a conversation, we expect the model to rephrase the factual content by taking conversational context into account. 
We derive the data for this task using the Topical Chat dataset \cite{Gopalakrishnan2019} and fine-tune a large pre-trained language model on it.

\citet{li-etal-2016-diversity, zhang2019dialogpt} use  maximum pointwise mutual information ({\maxpmi}) to filter out bad and unspecific responses sampled from a generative language model. 
However, we observe that {\maxpmi} responses lack in  acknowledgement, an essential human trait. 
This is because a generated response that simply copies over the new factual content while largely ignoring the conversational history can have high mutual information (MI) with the overall input. 

\textbf{Our second contribution is a method to select responses that exhibit human-like acknowledgement.}
To quantify the amount of information drawn from the two contexts of new factual content and conversational history, we propose using \textbf{pointwise conditional mutual information (PCMI)}.
\ashwin{Should I shorted the following by removing exact results? Apparently Introduction isn't supposed to contain results. }
We show that responses with a higher PCMI w.r.t conversational history given factual content ($\pcmi_h$) are judged by humans to be better at acknowledging prior turns 74\% of the time.%
\footnote{
  Statistically significant with $p < 0.05$ (Binomial Test).\label{fn:stat_sig}}
Then, we use $\pcmi_h$ to identify {\maxpmi} responses that lack acknowledgement and find alternative responses (Fused-PCMI) that trade off $\pmi$ for $\pcmi_h$. 
Despite a lower PMI, human annotators prefer the Fused-PCMI alternative over the {\maxpmi} response 60\% of the time.\footref{fn:stat_sig}
We release annotated conversations from the Switchboard corpus (with guidelines), code for fine-tuning and calculating scores and human evaluations.\footnote{\href{https://github.com/AshwinParanjape/human-like-informative-conversations}{https://github.com/AshwinParanjape/human-like-informative-conversations}}

\section{Strategies for informative conversations}
\label{sec:strategies}
\begin{figure}
    \centering
        \includegraphics[width=\columnwidth]{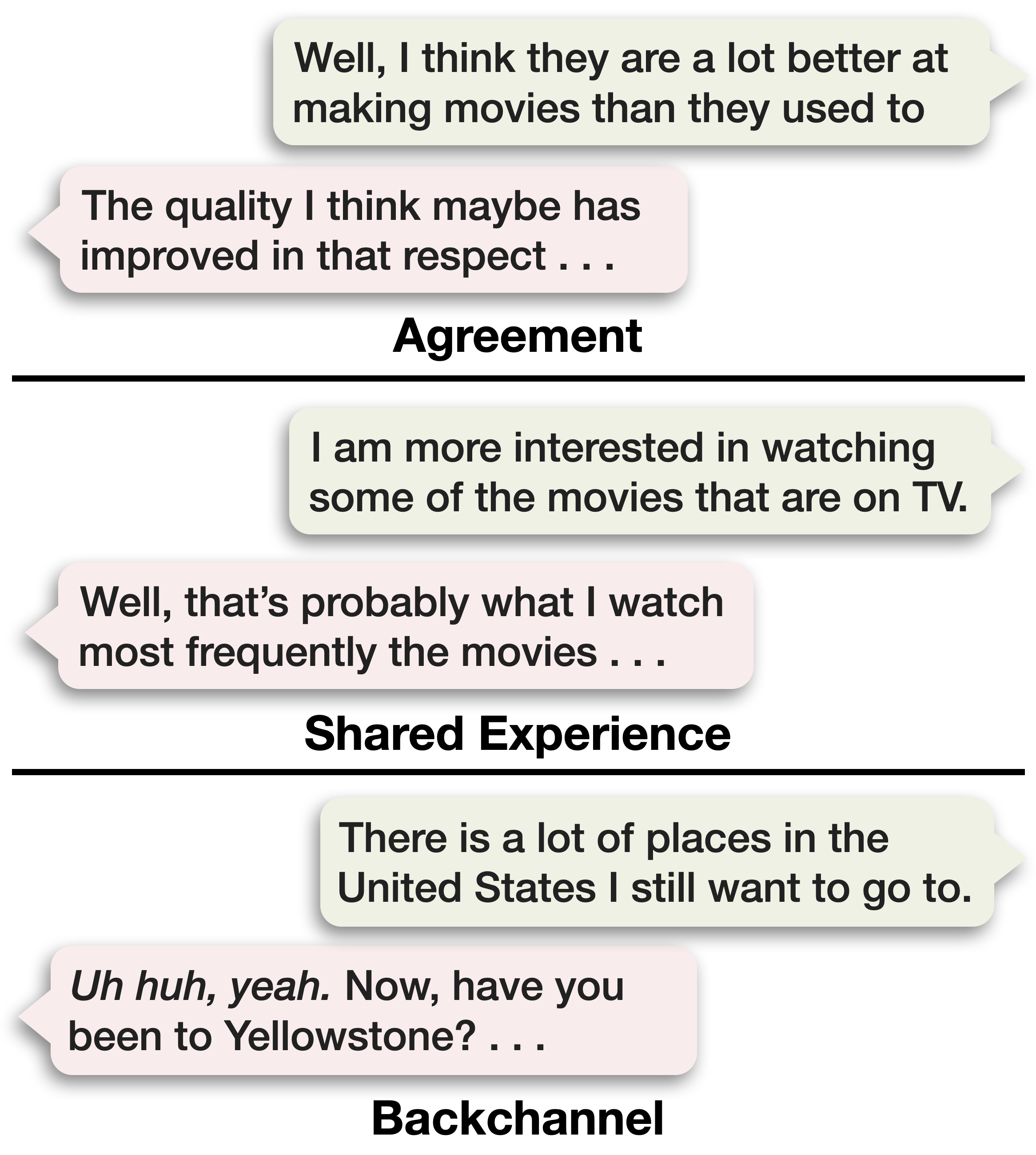}
    \caption{Examples for \textbf{Acknowledgement Strategies} from Switchboard (parts omitted for brevity).} \label{fig:acknowledgement_examples}
\end{figure}

\begin{figure}
    \centering
        \includegraphics[width=\columnwidth]{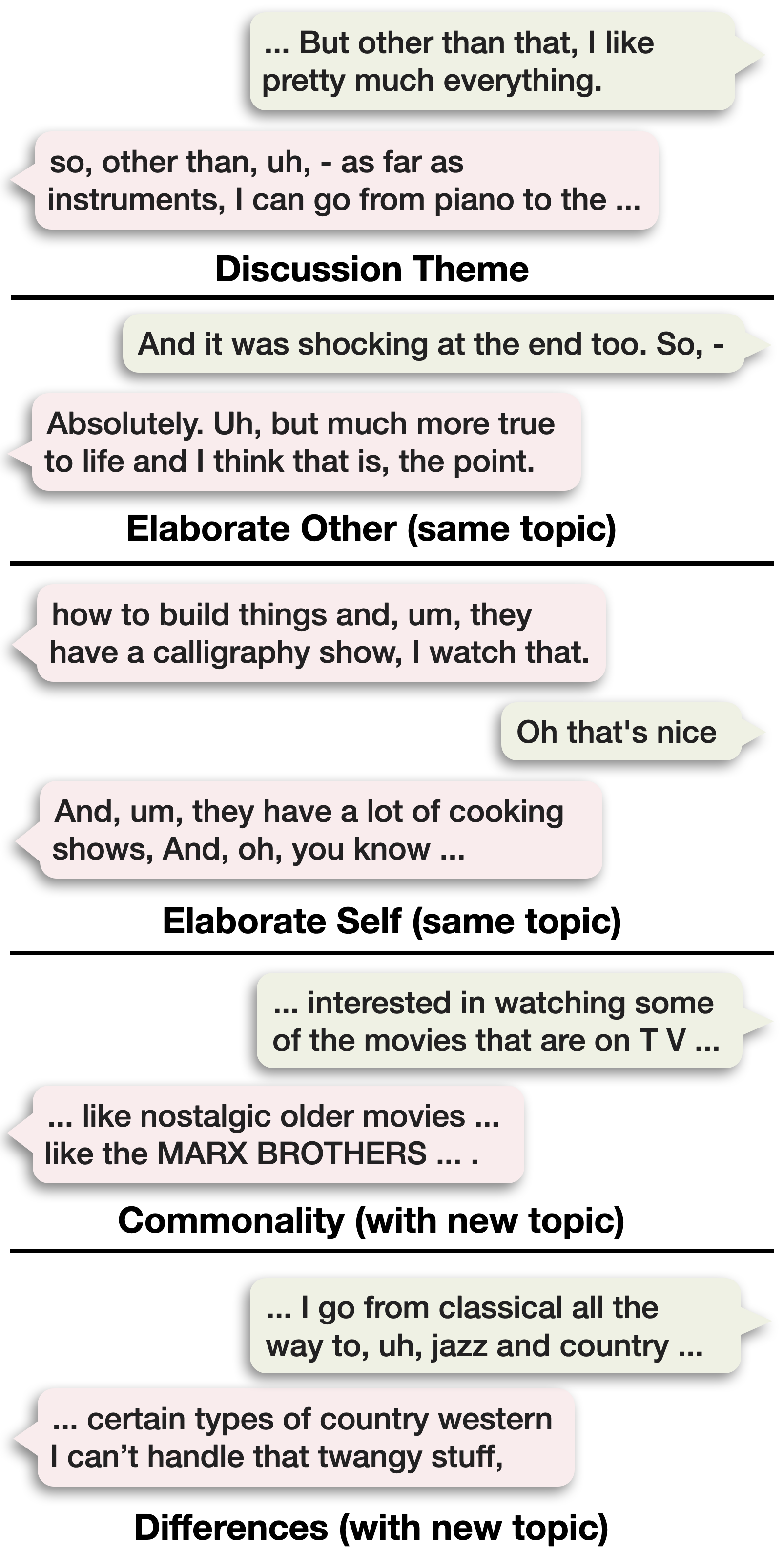}
    \caption{Examples for popular \textbf{Transition Strategies} from Switchboard (parts omitted for brevity).}
    \label{fig:transition_examples}
\end{figure}
    To understand strategies used by humans while talking about factual knowledge, we annotate 
    turns in human-human conversations. 
    We adopt and extend Herbert Clark's approach to conversational analysis. 
    According to his \textit{given-new} contract \cite{clark1977discourse}, the speaker connects their utterances with the given information (assumed to be known to the listener) and adds new information. 
This builds up \textit{common ground} \cite{stalnaker2002common} between the two participants, defined to be the sum of their mutual, common or joint knowledge, beliefs and suppositions.
We identify the following four aspects to the process of adding new information to a conversation.

\textbf{Acknowledgement strategies} According to \citet{clark1991grounding}, the listener provides positive evidence for grounding.
           We classify all mentions of prior context into various acknowledgement strategies. 
           
\textbf{Transition strategies}\hspace{0.5em} According to \citet{sacks1971stepwise}, topical changes happen step by step, connecting the given, stated information to new information. We annotate the semantic justifications for topical changes as different transition strategies. 

\textbf{Detail selection strategies}\hspace{0.35em} According to \citet{isaacs1987references}, speakers in a conversation inevitably know varying amounts of information about the discussion topic and must assess each other's expertise to accommodate their differences. 
        We posit that each speaker applies detail selection strategies to select the right level of detail to be presented. 
        
 \textbf{Presentation strategies}\hspace{0.35em} According to \citet{smith1993course}, presentation of responses is guided by two social goals -- exchange of information and self-presentation. While we do not consider social goals in this work, we hypothesize that people talk about factual information in non-factual forms (e.g., opinions, experiences, recommendations) which we classify as various presentation strategies.

\subsection{Analysis of strategies} 

\paragraph{Dataset}
We annotate part of the The Switchboard Dialog Act Corpus \citep{stolcke-etal-2000-dialogue}, an extension of the Switchboard Telephone Speech Corpus \citep{switchboard} with turn-level dialog-act tags. 
The corpus was created by pairing speakers across the US over telephone and introducing a topic for discussion. 
This dataset is uniquely useful because as a speech dataset, it is more intimate and realistic than text-based conversations between strangers. 
We annotate conversations on social topics which might include specific knowledge (like Books, Vacations, etc.)\ but leave out ones about subjective or personal experiences. 

\paragraph{Specific knowledge} We define \textit{specific knowledge} as knowledge that can be ``looked up'' but isn't widely known (as opposed to \textit{general knowledge} that everybody is expected to know and \textit{experiential knowledge} that can only be derived from embodied experiences). 
In this work, we are interested only in specific knowledge because it serves as a source of new information in a conversation that is hard for a language model to learn implicitly but is likely available as text that can be supplied to the system.
Out of 408 annotated turns, 111 (27\%) incorporate specific knowledge and account for 56\% of the tokens. 

\begin{figure*}
    \centering
        \input{figures/strategies.pgf} \caption{Distribution of acknowledgement, transition and presentation strategies} \label{fig:strategies_breakdown}
\end{figure*}
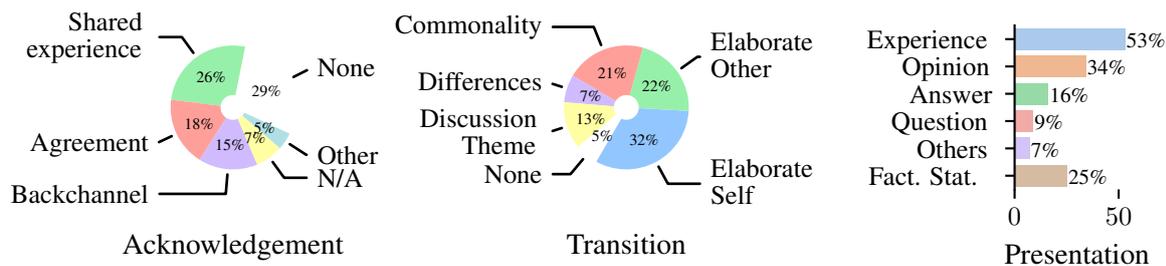

Next, we analyze various strategies employed in turns containing specific knowledge:
\paragraph{Acknowledgement Strategies} 
In 70\% of the turns, the speaker acknowledges the prior turn corroborating \citet{clark1991grounding}. 
Three main strategies (Figure~\ref{fig:acknowledgement_examples}): \kw{agreement} (or disagreement), \kw{shared experiences} (or differing experience) and \kw{backchanneling} account for 60\% of the turns (Figure~\ref{fig:strategies_breakdown}).
In certain cases, explicit acknowledgement isn't necessary. For example, the answer to a question demonstrates grounding and serves as an implicit acknowledgement. These are categorized as \kw{N/A}.
\kaitlyn{Neat, if there's room, an example could be great }

\paragraph{Transition Strategies} 
At the beginning of a conversation, the participants use the \kw{discussion theme} to pick a topic (various transition strategies are shown in Figure~\ref{fig:transition_examples}). 
The decision to stay on the topic or to transition to a new one is an implicit form of negotiation and depends on the interest and ability of both speakers to participate. 
Nearly half the time, people elaborate upon the current topic (Figure~\ref{fig:strategies_breakdown}). 
With a supportive listener, they might elaborate upon their own prior utterance (\kw{self-elaboration}). 
Or they might signal interest in continuing the topic by elaborating the other speaker's utterance (\kw{other-elaboration}). 
However, in a quarter of the turns, a participant loses interest or both participants run out of material. In that case, they transition to a new topic, implicitly justified by \kw{commonalities} or \kw{differences} with the current topic. 
If all else fails, they fall back to the \kw{discussion theme} to pick a new topic. 
\paragraph{Detail-selection strategies} 
People probe the other speaker's knowledge about an entity before diving into details. 
As a probing mechanism, people introduce an entity without any details (\kw{introduce-entity}) 50\% of the time. Depending on the response, \kw{details} are laid out 66\% of the time.
Note that a turn can have both labels, i.e., it can introduce an entity for the first time or it can have details of one entity while also introducing another entity.
Interestingly, in 7\% of turns, an entity's name is omitted but some details are presented, creating an opening for the other speaker to chime in.
\paragraph{Presentation strategies} A single utterance can have multiple modes of presentation. 
A \kw{factual} (objective) statement of specific knowledge is uncommon (25\%) in comparison with a subjective rendering in the form of an \kw{experience} (53\%) or an \kw{opinion} (34\%) (Figure \ref{fig:strategies_breakdown}). 
The other common modes of presentation are \kw{questions} (9\%) and \kw{answers} (16\%), which often occur as adjacency pairs. 
We also found a few \kw{other} uncommon modes (7\%) such as recommendations or hypotheses based on specific knowledge. 
\subsection{Implications for dialogue agents}
The four aspects -- acknowledgement, transition, detail selection and presentation -- are essential ingredients and indicative of quality conversation. 
They provide us with finer-grained questions amenable to human evaluation: \textit{``How does the agent acknowledge?''}, \textit{``Was it a smooth transition?''}, \textit{``Does the utterance contain the right level of detail?''}, and \textit{``Was the information presented as experience or an opinion?''}.

These four aspects are also more actionable than the evaluation metrics used in prior work. 
They can inspire new techniques that are purposefully built to emulate these strategies.
For instance, transitions can be improved with purpose-built information retrieval methods that use commonalities and differences to choose a new topic. 
To improve detail-selection, an agent could keep track of user knowledge and pragmatically select the right level of detail.
Moreover, in their datasets, \citet{dinan2018wizard} and \citet{Gopalakrishnan2019} asked people to reply using knowledge snippets, but that can lead to factual statements dominating the presentation strategies. 
We hope that newer datasets either suggest ways to reduce this bias or not provide knowledge snippets to humans in the first place but instead post facto match utterances to knowledge snippets. 

In the rest of the paper, we focus on generating responses with better acknowledgements.
This is because current neural generation methods perform poorly in this regard when compared with the other aspects.
They fail to acknowledge prior turns and even when they do, the acknowledgements are shallow and generic (e.g., backchannel). 
We hypothesize that the bottleneck is not the modeling capacity, but rather our inability to extract acknowledgements. 
The responses are not specific w.r.t.\ conversational context, a prerequisite for richer acknowledgements (e.g., shared experiences). 
We show that selecting responses specific to conversational context improves acknowledgements and overall quality. 
More broadly, we are able to demonstrate the utility of our linguistic analysis in evaluating and improving a dialogue agent. 

\section{A method for richer acknowledgements} \label{sec:methods}

Current neural generation methods typically offer short and formulaic phrases as acknowledgements: ``That's interesting'', ``I like that'', ``Yeah, I agree''. 
Such phrases are appropriate almost everywhere and convey little positive evidence for understanding or grounding. 
The training corpus, on the other hand, contains richer acknowledgements, which generated responses should be able to emulate. 

We assume that the representational capacity of current neural models is sufficient and that out of all the sampled responses, some do indeed contain a richer form of acknowledgement.
We posit that non-existent or poor sample selection strategies are to blame and that without a good sample selection strategy, improvements to the dataset, model or token-wise sampling methods are unlikely to help. 

\textit{We hypothesize that responses that are more specific to conversational history provide better evidence for understanding and hence contain richer acknowledgements. }
As a baseline sample selection strategy, we first consider maximum pointwise mutual information ({\maxpmi}) (as used by \citet{zhang2019dialogpt}) between the generated response and the conversational contexts (i.e., new factual content and conversational history).
However, this is insufficient because it is an imprecise measure of specificity w.r.t.\ conversational history. 
Instead, we use pointwise conditional mutual information (PCMI) to maintain specificity with individual contexts and propose a combination of PMI and PCMI scores to select overall better quality responses than {\maxpmi}.

\paragraph{Conversational rephrasing} 
    The choice of new factual content is a confounding factor for analysis. Hence, we define a simplified task, \textit{conversational rephrasing}, where content is provided as an input. 
Thus, conversational rephrasing is a generation task where conversational history ($\textbf{h}$) and new factual content ($\textbf{k}$) are given as inputs and a response ($\textbf{g}$) is generated as the output (Figure~\ref{fig:convpara_example}). 
We expect the generation $\textbf{g}$ to paraphrase the new factual content $\textbf{k}$ in a conversational manner by utilizing the conversational history $\textbf{h}$.  

\paragraph{Base generator}
We fix the sequence-to-sequence model and token-wise sampling method and vary the sample selection strategy.
The model is trained to take $\textbf{h}$ and $\textbf{k}$ as input and to generate $\textbf{g}$ as the output with the language modelling loss, i.e., we minimize the token-wise negative log likelihood.
During generation, tokens are sampled autoregressively from left-to-right. 
While sampling each token, the probability distribution is truncated using nucleus sampling \cite{Holtzman2020The} but the truncation is kept to a minimum with a high value of $p$ for top-p sampling. 
Multiple diverse candidates are sampled from the base generator and now the best candidate needs to be selected. 

\begin{figure*}
    \centering
        \input{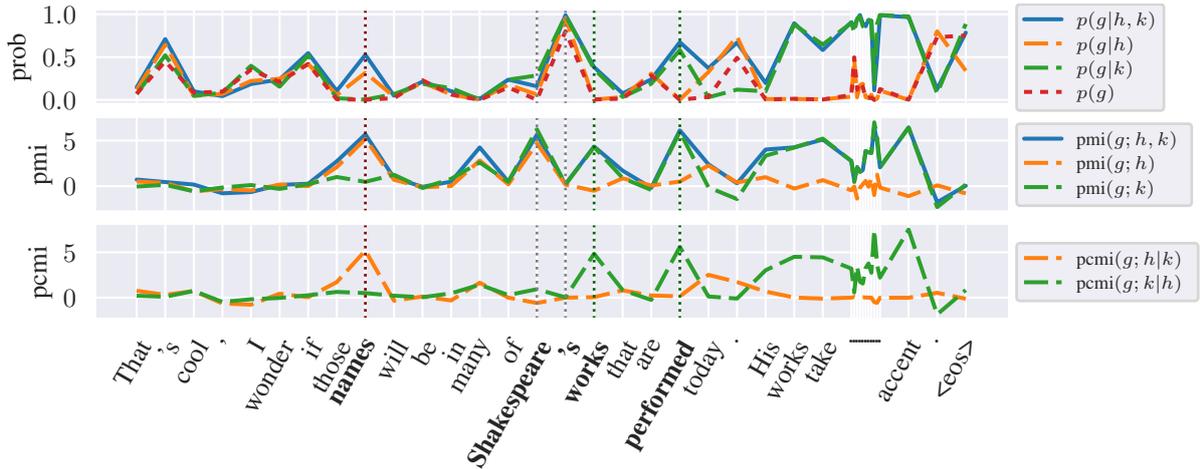}
    \caption{\textbf{Token-wise probabilities (top), $\pmibf$ (middle) and $\pcmibf$ (bottom) scores for the generated response $\textbf{g}$ from Figure~\ref{fig:convpara_example}.}  The $\pcmi$ graph is computed from the $\pmi$ graph which in turn is computed from the probability graph.
    The probabilities by themselves are unreliable measures of contextual specificity; the tokens predictable without  $\textbf{h}$, $\textbf{k}$ (e.g., \textit{'s}) have high probability but low $\pmi$.
    $\pmi$ cannot differentiate between the two contexts; tokens coming from both contexts (e.g., \textit{Shakespeare}) have high $\pmi$ but low $\pcmi$.
    $\pcmi$ differentiates the two contexts; tokens unique to conversational history $\textbf{h}$ (e.g., \textit{names}, \textit{today}) have high $\pcmi_h$, Tokens unique to new factual content $\textbf{k}$ (e.g., \textit{works}, \textit{performed}, all of last sentence) have high $\pcmi_k$.
    }
    \label{fig:probs_plot}
\end{figure*}
\paragraph{PMI for overall specificity}
\citet{li-etal-2016-diversity} suggest selecting the response with maximum PMI (referred to as MMI in their work) to maintain specificity and get rid of bland or low-quality samples.
Pointwise Mutual Information (PMI) between two events ($x$, $y$) is a measure of change in the probability of one event $x$, given another event $y$:  $\pmi(x ; y) \equiv \log \frac{p(x|y)}{p(x)}$.  
We use $\pmi$ to determine the increase in likelihood of $\textbf{g}$, given $\textbf{h}$ and $\textbf{k}$.
$$\pmi(\textbf{g}; \textbf{h}, \textbf{k}) = \log \frac{p(\textbf{g}|\textbf{h}, \textbf{k})}{p(\textbf{g})}$$
A candidate generation $\textbf{g}$ with higher PMI is more likely given the two contexts $\textbf{h}$ and $\textbf{k}$ than otherwise and is therefore considered more specific to the contexts. 
A low PMI value for a candidate response implies non-specificity to either context providing a clear signal for discarding it. 
A high PMI is necessary but not sufficient for a candidate to be specific to both the contexts simultaneously, since mutual information could come from either context.
For example, $\textbf{g}_2$ (Figure~\ref{fig:convpara_example})  merely copies \textbf{k} but gets a high PMI score (Table~\ref{table:pmi_pcmi_differentiation}). Whereas $\textbf{g}_1$ acknowledges prior turn and uses \textbf{k} but gets a lower PMI score. 

\begin{table}[t] 
\centering
    \begin{tabularx}{\columnwidth}{l|rrr}
\toprule
    \textbf{Response} & $\pmibf(\mathbf{g}; \mathbf{h}, \mathbf{k})$ & $\pmibf(\mathbf{g}; \mathbf{h})$ &  $\pcmibf_h$\\ \midrule
    $\textbf{g}_1$  & 87 & 18 & 14 \\ 
    $\textbf{g}_2$  & 150 & 18 & 4 \\ 
        \bottomrule
     
\end{tabularx}

    \caption{Measures of mutual information for generated responses from Figure 1.
    $\textbf{g}_2$ largely copies \textbf{k}, has high $\pmi(\textbf{g}; \textbf{h}, \textbf{k})$ and would be chosen by \maxpmi. $\textbf{g}_1$'s first sentence acknowledges using \textbf{h} and bridges to \textbf{k}; it would be chosen by Fused-PCMI on the basis of $\pcmi_h$. $\pmi(\textbf{g}; \textbf{h})$ cannot differentiate the two. }
\label{table:pmi_pcmi_differentiation}
\end{table}

\paragraph{PCMI for contextual specificity} Pointwise Conditional Mutual Information (PCMI) considers a third variable ($z$) and
removes information due to $z$ from $\pmi(x;y,z)$ to keep only the information uniquely attributable to $y$.
$$\pcmi(x;y|z) = \pmi(x; y, z) - \pmi(x; z)$$
We propose using $\pcmi$ for contextual specificity, i.e., $\pcmi_h = \pcmi(\textbf{g};\textbf{h}|\textbf{k})$ for specificity w.r.t.\ to conversational history $\textbf{h}$, and $\pcmi_k =\allowbreak \pcmi(\textbf{g};\textbf{k}|\textbf{h})$ for specificity w.r.t.\ new factual content $\textbf{k}$.

{\em Since acknowledgement strategies are primarily based on the history of the conversation thus far, we would expect candidates with higher $\pcmi_h$ to exhibit more human-like acknowledgement strategies.}

As a point of comparison, consider using $\pmi(\textbf{g};\textbf{h})$ instead of $\pcmi_h$.
In our setting of conversational rephrasing for informative dialogue, \textbf{k} topically overlaps with \textbf{h}. 
If $\textbf{g}$ merely copied over the new factual content $\textbf{k}$ without any reference to $\textbf{h}$, it would still have a high $\pmi(\textbf{g};\textbf{h})$ due to topical overlap but a low $\pcmi_h$. 
Going back to Table~\ref{table:pmi_pcmi_differentiation}, we can see that $\pmi(\textbf{g};\textbf{h})$ is unable to distinguish between the two examples but $\pcmi_h$ is. 

In Figure~\ref{fig:probs_plot}, the above quantities are broken down to token-level granularity. 
We can see that specific words that are uniquely attributable to each context are cleanly separated by both $\pcmi_h$ and $\pcmi_k$. 

\paragraph{Combining PMI \& PCMI for overall quality} \label{para:PCMI_overall}
To show the utility of $\pcmi_h$ in improving overall quality, we propose a heuristic method to find a more balanced response (\textbf{Fused-PCMI}) than the {\maxpmi} response. 
{\em For every {\maxpmi} response with a low $\pcmi_h$, we consider an alternative that has both high $\pcmi_h$ and an acceptable PMI.}
If such an alternative is found, we select that as the Fused-PCMI response; otherwise we default to the {\maxpmi} response as the Fused-PCMI response.
We consider a PMI score in the top 50\% of the candidate set as acceptable.  
To compute $\pcmi$ thresholds, we calculate quantiles based on the entire validation set and consider $\pcmi_h$ in the first quartile to be low and $\pcmi_h$ in the fourth quartile to be high.
This approach is less susceptible to outliers, more interpretable and easier to calibrate than a weighted arithmetic or geometric mean.

\section{Evaluation Setup}
We derive the data for our conversational rephrasing task from the Topical Chat dataset \cite{Gopalakrishnan2019}. 
We use it to fine-tune a large pre-trained neural language model. 
This forms the base model as described in Section \ref{sec:methods}.
To evaluate our proposed methods, we design three experiments and perform a comparative study with human annotators. 

\paragraph{Topical Chat Dataset} This is a human-human chat dataset where crowd-workers were asked to chat with each other around certain topics. 
They were provided with relevant interesting facts from the ``Today I learned'' (TIL) subreddit which they could use during the conversation.
TILs are are short (1--3 sentences), self-contained, interesting facts, most of them from Wikipedia articles. 
When an utterance can be matched to a TIL (based on a TF-IDF threshold of 0.12), we create an instance for the conversational rephrasing task: with the utterance as $\textbf{g}$, the two previous utterances as $\textbf{h}$ and the corresponding TIL as $\textbf{k}$.
We split the instances into training, validation and test sets (sizes in Section ~\ref{sec:experimental_details}) such that all utterances related an entity belong to the same set. 

\paragraph{Base Model}
\label{sec:models}
We use the GPT2-medium model (24-layer; 345M params) pretrained on the English WebText dataset  \cite{radford2019language}, as implemented in HuggingFace's TransferTransfo \cite{TransferTransfo, Wolf2019HuggingFacesTS} framework.
Fine-tuning is performed using the language modelling objective on the training set with default hyperparameters until lowest perplexity is reached on the validation set. 
During generation, we sample tokens using nucleus sampling \cite{Holtzman2020The} with $p = 0.9$ and temperature $\tau=0.9$ and get candidate responses.
To compute auxiliary probabilities $\{ p(\textbf{g}|\textbf{h})$, $p(\textbf{g}|\textbf{k})$, $p(\textbf{g}) \}$ for these candidates, we use separate ablation models. 
The ablation models are trained similar to the base model but after removing respective contexts from the training inputs.

\subsection{Experimental Design}
To validate our proposed methods, we do a paired comparison (on Amazon Mechanical Turk) where human annotators are shown two prior turns of conversational history and asked to choose between two candidate responses. 
Annotators are allowed to mark both candidates as nonsensical if the responses don't make sense.
In Section~\ref{sec:mturk_interfaces}, we show the interfaces used to collect annotations from Amazon Mechanical Turk.
Each pair of responses was compared by three annotators -- we consider a candidate to be better than the other when at least two of them (majority) agree upon it.
For each of the following three experiments, we compare 100 pairs of candidates generated using instances from the test set. 
The null hypothesis ($H_0$) for the three experiments is that there is no difference between the methods used to generate the candidates and we hope to reject the null hypothesis in favor of the alternate hypothesis ($H_1$) at a significance level ($\alpha$) of 0.05. 

\paragraph{Exp 1: PMI and overall quality} First, we want to confirm that {\em high PMI responses are overall better quality than randomly chosen candidates} ($H_1$). 
To do so, we first generate 10 responses for each instance and compare the response having maximum $\pmi(\textbf{g};\textbf{h}, \textbf{k})$ ({\maxpmi})  with a randomly chosen response from the remaining 9. 
We ask human annotators to pick the overall better candidate response.

    \paragraph{Exp 2: $\pcmibf_\textbf{h}$ and acknowledgement} We test if {\em responses having high $\pcmi_h$ provide better acknowledgement} ($H_1$). 
To do so, we first sample 100 responses (larger than previous experiment) and out of all possible pairs keep those with $|\Delta\pcmi_h|>15$ (larger than population interquartile range; Figure~\ref{fig:PCMI_skew_univariate}). 
To control for the amount of new information being added, we pick pairs with closest values of $\pcmi_k$ (recall that $\pcmi_k$ denotes information uniquely attributable to $\textbf{k}$). 
Such selected pairs have $\text{Median}|\Delta\pcmi_k|=0.42$. 
We ask annotators to pick the response that provides better acknowledgement and select an acknowlegement span to support their claim.

    \paragraph{Exp 3: Fused-PCMI vs.\ {\maxpmi}} We test if {\em the proposed method, Fused-PCMI (that combines PMI and PCMI) selects better responses than {\maxpmi}} ($H_1$). 
For Fused-PCMI, we set low and high $\pcmi_h$ thresholds to be 5 and 14 respectively based on population quartiles.
For instances where the Fused-PCMI response is different from the {\maxpmi} response, we compare the two. 
We consider 10 candidate responses for each test instance and find that for around 10\% of the instances the Fused-PCMI candidate is different from the {\maxpmi} candidate. 
Human annotators are then asked to pick the overall better response of the two.

\section{Results \& Analyses}
    Based on human annotations, we are able to reject $H_0$ in favor of $H_1$ in all three experiments (Table~\ref{table:results})\footnote{Statistically significant with $p < 0.05$ (Binomial Test). \label{fn:stat_sig_2}}: high PMI responses are overall better quality than randomly chosen candidates, responses having high $\pcmi_h$ provide better acknowledgement, and Fused-PCMI selects better responses than {\maxpmi}. 
    
\begin{table}[h]
\centering
    \begin{tabular}{lrrrr} \toprule
        \textbf{Exp} & \textbf{n} & \textbf{K} & \textbf{p} &$\mathbf{\kappa}$\\ \midrule
        \textbf{1}  & 87 & 55 (63\%) & 0.009 & 0.18\\ 
        \textbf{2}  & 95 & 70 (74\%) & $3\mathrm{e}{-6}$ & 0.48\\ 
        \textbf{3}  & 99 & 59 (60\%) & 0.035 & 0.11\\
        \bottomrule
     
    \end{tabular}
    \caption{\textbf{Human annotation results.}
    Out of 100 instances, majority agreement was reached in \textbf{n} instances. The majority rejects the null-hypothesis ($H_0$) in favor of the alternate hypothesis ($H_1$) in \textbf{K} instances. \textbf{p} denotes the p-value and $\mathbf{\kappa}$ denotes Fliess kappa for Inter-annotator agreement.  }
\label{table:results}
\end{table}

While according to Exp~1, high PMI responses are overall better quality, upon further analysis we find that \textit{PMI is useful for filtering out bad samples, but not necessarily for selecting between the good samples.}
When paired with a random response from the top 50\% of the candidates (ranked according to their PMI), people prefer the {\maxpmi} response only 52\% of the time (not significant). 
On the other hand, if the random response was in the bottom 50\%, then the {\maxpmi} response is preferred 74\% of the time.\footref{fn:stat_sig_2}

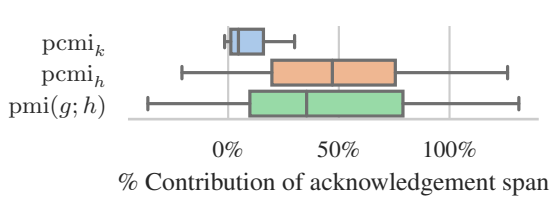
\begin{figure}
    \centering
    \input{figures/PCMI_acknowledgement_ratio.pgf}
    \caption{\textbf{Attribution to acknowledgement span.} A larger fraction of $\pcmi_h$ can be attributed to human annotated acknowledgement spans compared to $\pmi(g;h)$ and $\pcmi_k$.  
    }
    \label{fig:PCMI_acknowledgement_span}
\end{figure}
\begin{table*}[h]
    \small
\centering
    \begin{tabularx}{\textwidth}{XX}
\toprule
        \textbf{Previous Utterance} & \textbf{High $\pcmibf_h$ candidate} \\ \midrule
        No I did not! Han Solo was apparently also an imperial lieutenant before meeting up with Chewbacca. & \textbf{that is very interesting and I wonder if he was one of the first or the first one to meet Chewba.} I was just reading that George Lucas originally intended Han to be a green alien\\
        I did not. That sounds interesting. Much like the "tip of the hat" originates from Knights raising their visors. & \textbf{Yes, that sounds like a very cool tradition.} I wonder how they enforced that law, it sounds as ridiculous as the prohibition of wearing a hat in 1912, that was a law that would prevent the use of hat pins by suffragettes. \\
        Yeah. When Apollo 11 was preparing for their voyage, they had discussion as to whether they should plant a United Nations flag instead of a US flag. If I planted a UN flar and someone asked me "whose flag is that?" I would answer, mine… & lol, no kidding. \textbf{I wonder if they are still discussing the issue with the moon?} I can't believe the moon is moving away from us at about the rate that your nails grow. That seems really slow to me.\\
        \bottomrule
\end{tabularx}

    \caption{Human annotated spans of text that indicate acknowledgement (in bold) in candidate responses with higher $\pcmibf_h$ from Exp 2.}

\label{table:exp_2_examples}
\end{table*}
In Exp 2, we ask annotators to mark text-spans that indicate acknowledgement (Table~\ref{table:exp_2_examples}). If token-level $\pcmi_h$ is concentrated in these spans, we have further proof that $\pcmi_h$ indicates acknowledgement. Indeed, in Figure~\ref{fig:PCMI_acknowledgement_span}, we see that $\pcmi_h$ is most attributable to the acknowledgement spans, followed by $\pmi(g;h)$ and $\pcmi_k$. 
\textit{Thus, $\pcmi_h$ captures acknowledgements with greater specificty than $\pmi(g;h)$.}

\begin{figure}
    \centering
    \input{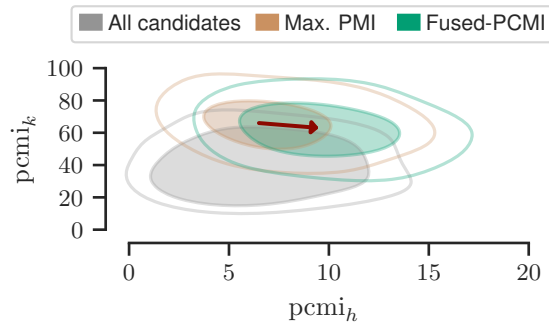}
    \caption{\textbf{Bivariate Kernel Density Estimate plot All candidates, {\maxpmi} responses and Fused-PCMI responses.} Bivariate kernel density estimate plot w.r.t. $\pcmibf_k$ and $\pcmibf_h$ at levels 0.5 and 0.75. We see that Fused-PCMI responses compared with {\maxpmi} trade off little $\pcmi_k$ for a large relative gain in $\pcmi_h$. See Figure \ref{fig:PCMI_skew_univariate} in Section \ref{sec:appendix_univariate} for univariate box plots. 
    }
    \label{fig:PCMI_skew}
\end{figure}

\begin{table*}[h]
    \small
\centering
    \begin{tabularx}{\textwidth}{rXXX}
\toprule
        \# & \textbf{Previous Utterance} & \textbf{\maxpmi\ candidate} & \textbf{Fused-PCMI candidate} \\ \toprule
        1 & I live in the southwest USA desert. Not much in the way of water here either, but we've got some nice artificial pools. What kind of swimming did you do? &\textcolor{gray}{There is a type of competitive swimming where the athlete has their hands and legs bound, and swims like a dolphin}. I have never heard that before. It seems fishy to me. I'm not sure I can swim like that. & I went competitively but was having a tough time. I liked that kind of \textcolor{gray} {competition where the swimmer can swim with their hands bound}. \\ \midrule
        2 & It is interesting. It's also interesting that they are trying to plan a Mars colonization project that will be funded by a reality show, I really want to see that! & \textcolor{gray} {A satellite was lost on Mars because of the use of the Imperial system and NASA used the metric system.} & Sounds like a weird reality show.  I'm going to look it up. \textcolor{gray} {A satellite was lost on Mars because it was used by NASA using the metric system}. \\  \midrule
        3 & Yeah I definitely like clothes. I think it's interesting how many uses hats have, like for protection, ceremonial reasons, religious reasons, safety, and fashion. & \textcolor{gray} {Did you know that until 1998, British politicians had to wear a top hat if making a point of order in the house of commons?} & \textcolor{gray} {In 1998, British politicians were required to wear a top hat if making a point of order in the house of commons}. I guess that's kind of messed up and counterproductive.\\
        \toprule
        4 & Yeah it's super funny that he had a rap battle in the lunchroom. Seems like something straight out of a movie! & Ha, I agree. Speaking of rappers, \textcolor{gray} {did you know that "California Gurls" was actually a West Coast vs East Coast answer song to Alicia Keys and Jay Z's "Empire State of Mind"} & Yeah, and how about the fact that the \textcolor{gray}{California Gurls song was actually a  west coast vs east coast  answer song to a pair of famous rappers}. That is surprising \\  \midrule
        5 & im not surprised. rats are quite clever, and i guess if scientists understand the rat brain, then they can wire up the interface. cool. & It's amazing how they can be clever and change forms. I like how they can now \textcolor{gray}{Debrate a chicken with surgical precision}. & \textcolor{gray}{Robots can also debone chicken with precision}. I guess that would help avoid chicken pogis. \\ \bottomrule
\end{tabularx}

    \caption{\textbf{Illustrative samples of selected responses used in Exp 3.}
    For samples 1,2 and 3 people prefer Fused-PCMI and for samples 4 and 5 they prefer {\maxpmi}.
    Factual content copied largely verbatim by the model is in gray. 
    Specifically, the Fused-PCMI candidate in 1 answers the question ({\maxpmi} does not) and in 3 refers back to contradict utility of hats.}
    \label{table:exp_3_examples}
\end{table*}
To understand the mechanism behind the improvement in Exp 3, we look at the distribution of samples w.r.t.\ $\pcmi_k$ and $\pcmi_h$ in Figure~\ref{fig:PCMI_skew}. We observe that {\maxpmi} responses heavily skew the distribution towards higher $\pcmi_k$, whereas Fused-PCMI responses show a more balanced improvement along both $\pcmi_h$ and $\pcmi_k$. \textit{Fused-PCMI increases both $\pcmi_h$ and $\pcmi_k$ (medians cross 75\% quartiles), indicating that the responses are simultaneously specific to both
$\textbf{h}$ and $\textbf{k}$.}

\section{Discussion} 
\label{sec:related-work}

\todo{New caption}

We show that samples with higher $\pcmi_h$ provide better acknowledgement and Fused-PCMI improves overall quality compared to {\maxpmi}. 
Thus, by improving acknowledgements -- an aspect we identified during our analysis of human strategies -- we were able to improve overall quality. 
This demonstrates the utility of linguistic analysis for finding interpretable and actionable metrics. 

While we show that our learnings apply to informative dialogue which adds factual knowledge \cite{dinan2018wizard, parthasarathi-pineau-2018-extending,  Gopalakrishnan2019}, we expect it to generalize to any dialog setting that adds new content, e.g., experiences \cite{ghazvininejad2018knowledge} and personas \cite{zhang-etal-2018-personalizing}. Any dual-context language generation task where the two contexts are asymmetric in their information content can potentially benefit from PCMI. 

There is scope for improvement: {\maxpmi} still selects better responses than Fused-PCMI in 40\% of the instances. 
    This could be because it is easy for the model to copy over \textbf{k} and generate a high PMI response that is also fluent and accurate. 
    Fused-PCMI encourages synthesis of acknowledgement using \textbf{h} and abstraction over \textbf{k} and it could therefore be prone to disfluencies and inaccuracies. 
We hope that orthogonal modeling improvements \cite{meng2019refnet} reduce such effects. 

A cause for concern with the human evaluation is low inter-annotator agreement for Exp 1 and 3 where we ask them to pick responses with ``overall better quality and suitability''. 
However, quality measurements are inherently subjective; people differ in the importance they place on different aspects such as engagement, informativeness, fluency etc., as corroborated by prior work \cite{finch-choi-2020-towards} that shows low Cohen's kappa (0.13, 0.22) for overall quality judgements. 
In this work, diverse expectations from multiple annotators are captured yet subsequently averaged into ``overall quality''. We leave it to future work to find finer-grained metrics that have high inter-annotator agreement and derive empirical weights to combine them into ``overall quality''.

In this work, we looked at acknowledgements and their impact on quality in isolation, but in a real system, the performance of the model also depends on other factors like user compliance and the retrieval model. 
In practice, we think the interplay between the four linguistic aspects is critical and needs to be explored. 
For instance, preliminary experiments with live conversations and an off-the-shelf retriever suggested that a bad choice of \textbf{k} with tenuous connections to \textbf{h} can make synthesis harder and lead to lower quality Fused-PCMI responses. 
Better retrieval models \cite{Ren2020ThinkingGA} that make use of transition strategies to determine \textbf{k} can lead to better acknowledgements. 
 
In this work, we identified salient aspects of human-human informative conversations and found deficiencies in current neural dialogue systems. We proposed a PCMI-based selection strategy that selected responses with acknowledgements and higher overall quality. We hope that our work provides actionable insights and metrics for future work and more generally inspires the use of linguistic literature for grounding conversational research. 

\section{Acknowledgements}
We are grateful to Amelia Hardy, Nandita Bhaskhar, Omar Khattab, Kaitlyn Zhou, Abigail See, other Stanford NLP group members and the anonymous reviewers for helpful comments. 
This research is funded in part by Samsung Electronics Co., Ltd.\ and in part by DARPA CwC under ARO prime contract no.~W911NF-15-1-0462. 
This article solely reflects the opinions and conclusions of its authors. 
Christopher Manning is a CIFAR Fellow.

\bibliography{anthology,convpara2020}
\bibliographystyle{acl_natbib}

\clearpage
\appendix
\section{Appendix}
\subsection{Model training details}
\label{sec:experimental_details}
Each model (main and ablation) was trained on a single NVIDIA Titan Xp GPU for 5 epochs and took approximately 8 hours to train. 
The training dataset had 51407 instances, validation 2491 and test 2728. 
The Topical Chat dataset and Switchboard corpus are in English language. 
The main model used for response generation had a validation loss (average negative log liklihood) of 2.05 which it reached after 2 epochs. 

\subsection{Univariate distribution}
\label{sec:appendix_univariate}
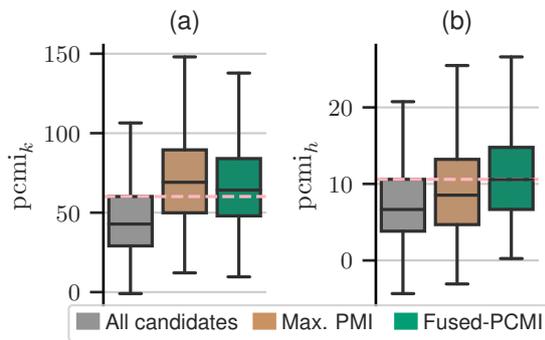
\begin{figure}[h!]
    \centering
    \input{figures/PCMI_skew_only_univariate.pgf}
    \caption{\textbf{Univariate box plots for All candidates, {\maxpmi} responses and Fused-PCMI responses.} (a) is w.r.t. $\pcmi_k$ and (b) w.r.t $\pcmi_h$. Pink horizontal lines indicate 75\% quartile for All candidates. 
    {\maxpmi} responses (orange) have high $\pcmi_k$ (median above pink line), but low $\pcmi_h$. 
    Fused-PCMI responses (green) show balanced yet high $\pcmi_h$ and $\pcmi_k$ (medians cross pink lines). }
    \label{fig:PCMI_skew_univariate}
\end{figure}

\subsection{Annotation Details}
\label{sec:mturk_interfaces}
We had 9, 19 and 19 unique annotators for experiments 1, 2 and 3 respectively. 
All three annotators agreed in 32/87 instances for experiment 1, 52/87 instances for experiment 2 and 32/99 instances for experiment 3. 
\begin{figure*}
    \centering
        \includegraphics[width=\textwidth]{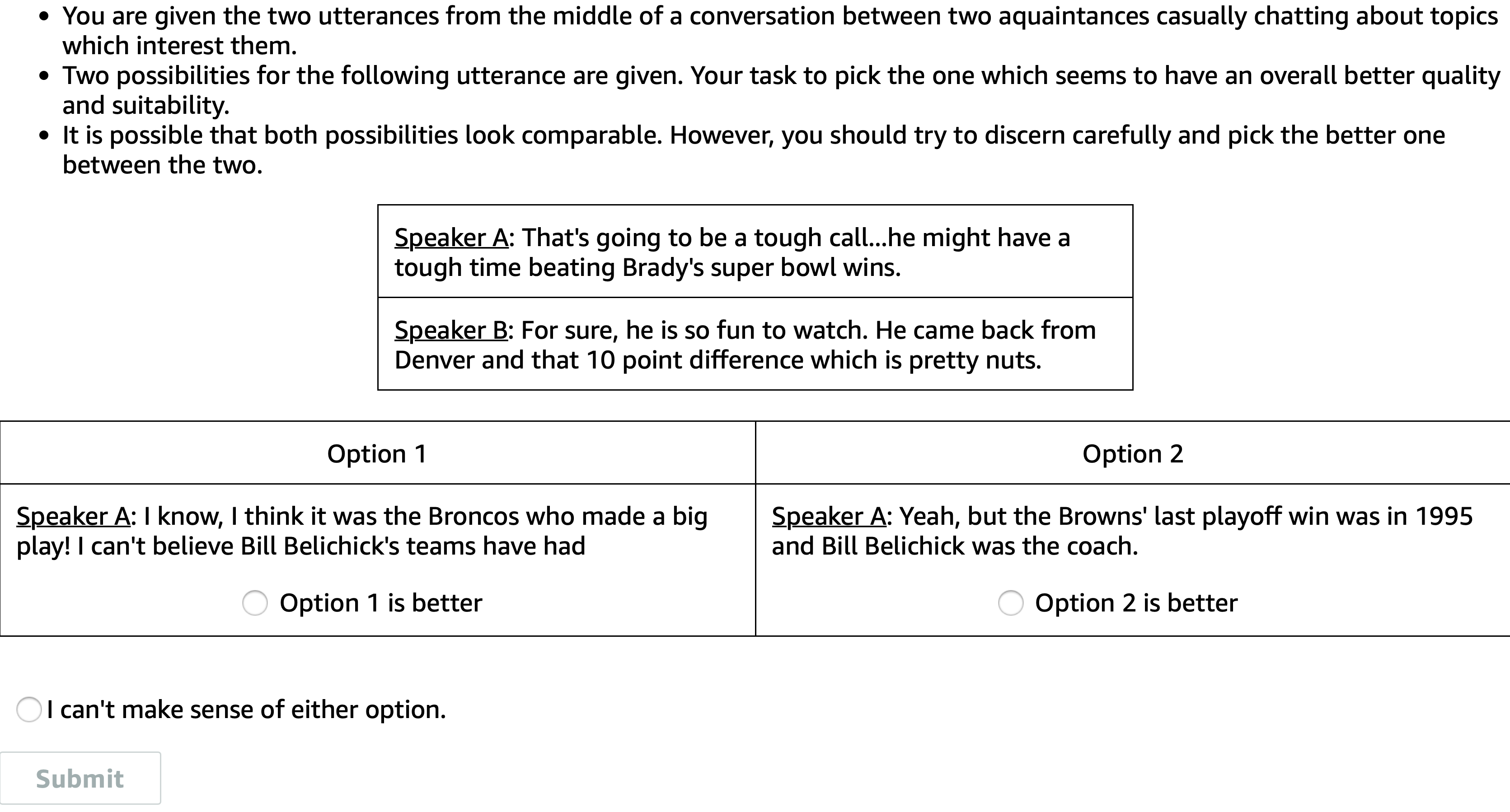}
    \caption{Annotation interface for Best PMI v/s rest}
    \label{fig:mturk_interfaces_better}
\end{figure*}
\begin{figure*}
    \centering
        \includegraphics[width=\textwidth]{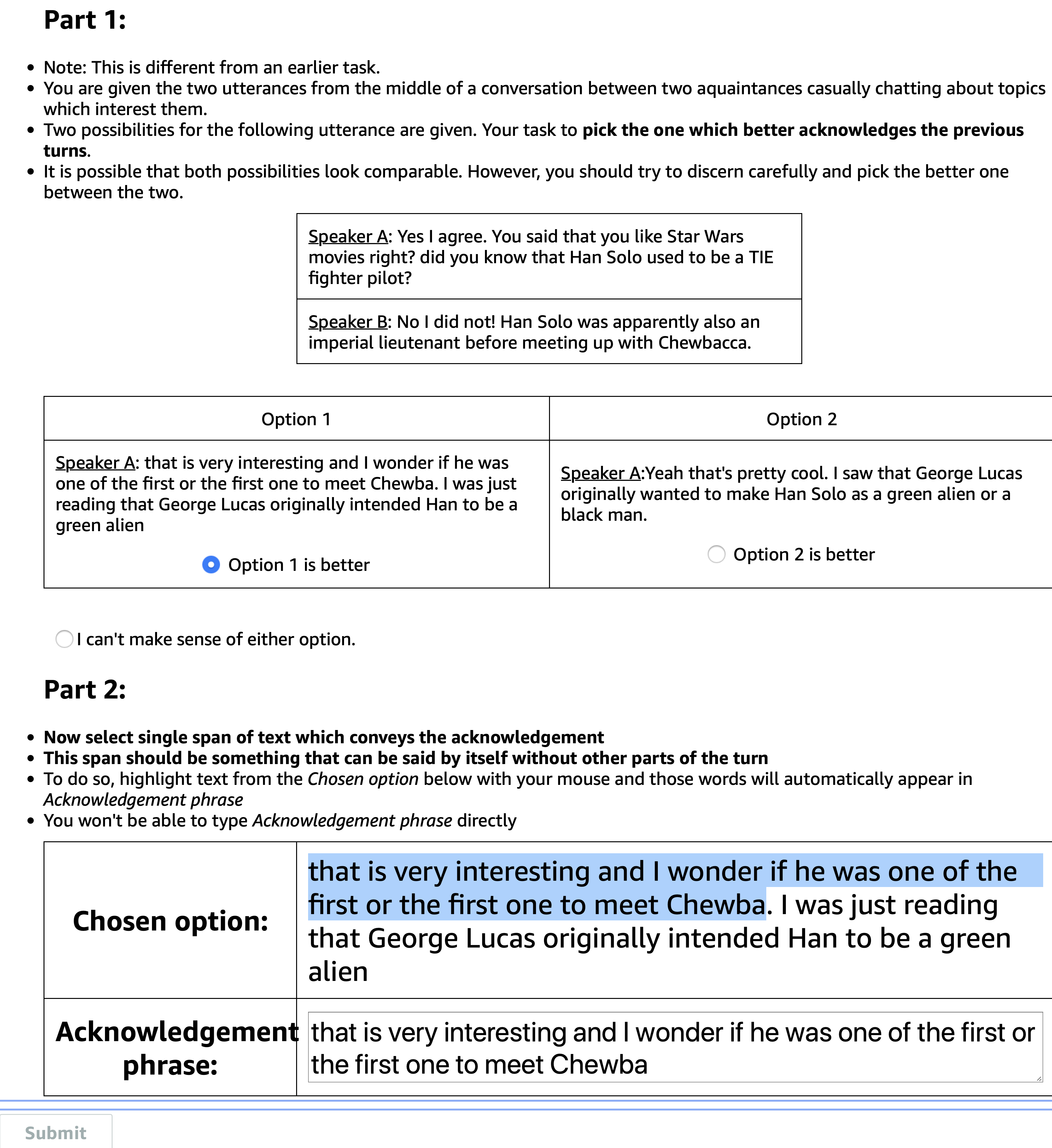}
    \caption{Annotation interface for acknowledgement differences due to $\pcmi_h$}
    \label{fig:mturk_interfaces_acknowledgement}
\end{figure*}

\end{document}

%% file: figures/strategies.pgf
\begingroup%
\makeatletter%
\begin{pgfpicture}%
\pgfpathrectangle{\pgfpointorigin}{\pgfqpoint{6.299212in}{1.557251in}}%
\pgfusepath{use as bounding box, clip}%
\begin{pgfscope}%
\pgfsetbuttcap%
\pgfsetmiterjoin%
\pgfsetlinewidth{0.000000pt}%
\definecolor{currentstroke}{rgb}{1.000000,1.000000,1.000000}%
\pgfsetstrokecolor{currentstroke}%
\pgfsetstrokeopacity{0.000000}%
\pgfsetdash{}{0pt}%
\pgfpathmoveto{\pgfqpoint{0.000000in}{0.000000in}}%
\pgfpathlineto{\pgfqpoint{6.299212in}{0.000000in}}%
\pgfpathlineto{\pgfqpoint{6.299212in}{1.557251in}}%
\pgfpathlineto{\pgfqpoint{0.000000in}{1.557251in}}%
\pgfpathclose%
\pgfusepath{}%
\end{pgfscope}%
\begin{pgfscope}%
\pgfsetroundcap%
\pgfsetroundjoin%
\pgfsetlinewidth{1.003750pt}%
\definecolor{currentstroke}{rgb}{0.000000,0.000000,0.000000}%
\pgfsetstrokecolor{currentstroke}%
\pgfsetdash{}{0pt}%
\pgfpathmoveto{\pgfqpoint{1.605827in}{1.089248in}}%
\pgfpathlineto{\pgfqpoint{1.535862in}{1.053766in}}%
\pgfusepath{stroke}%
\end{pgfscope}%
\begin{pgfscope}%
\definecolor{textcolor}{rgb}{0.000000,0.000000,0.000000}%
\pgfsetstrokecolor{textcolor}%
\pgfsetfillcolor{textcolor}%
\pgftext[x=1.659296in,y=1.099734in,left,]{\color{textcolor}\rmfamily\fontsize{10.000000}{12.000000}\selectfont None}%
\end{pgfscope}%
\begin{pgfscope}%
\pgfsetroundcap%
\pgfsetroundjoin%
\pgfsetlinewidth{1.003750pt}%
\definecolor{currentstroke}{rgb}{0.000000,0.000000,0.000000}%
\pgfsetstrokecolor{currentstroke}%
\pgfsetdash{}{0pt}%
\pgfpathmoveto{\pgfqpoint{0.841602in}{1.262126in}}%
\pgfpathlineto{\pgfqpoint{0.957739in}{1.262126in}}%
\pgfpathlineto{\pgfqpoint{1.017209in}{1.179702in}}%
\pgfusepath{stroke}%
\end{pgfscope}%
\begin{pgfscope}%
\definecolor{textcolor}{rgb}{0.000000,0.000000,0.000000}%
\pgfsetstrokecolor{textcolor}%
\pgfsetfillcolor{textcolor}%
\pgftext[x=0.368998in, y=1.298777in, left, base]{\color{textcolor}\rmfamily\fontsize{10.000000}{12.000000}\selectfont Shared}%
\end{pgfscope}%
\begin{pgfscope}%
\definecolor{textcolor}{rgb}{0.000000,0.000000,0.000000}%
\pgfsetstrokecolor{textcolor}%
\pgfsetfillcolor{textcolor}%
\pgftext[x=0.152949in, y=1.156030in, left, base]{\color{textcolor}\rmfamily\fontsize{10.000000}{12.000000}\selectfont experience}%
\end{pgfscope}%
\begin{pgfscope}%
\pgfsetroundcap%
\pgfsetroundjoin%
\pgfsetlinewidth{1.003750pt}%
\definecolor{currentstroke}{rgb}{0.000000,0.000000,0.000000}%
\pgfsetstrokecolor{currentstroke}%
\pgfsetdash{}{0pt}%
\pgfpathmoveto{\pgfqpoint{0.839009in}{0.713818in}}%
\pgfpathlineto{\pgfqpoint{0.905042in}{0.744990in}}%
\pgfusepath{stroke}%
\end{pgfscope}%
\begin{pgfscope}%
\definecolor{textcolor}{rgb}{0.000000,0.000000,0.000000}%
\pgfsetstrokecolor{textcolor}%
\pgfsetfillcolor{textcolor}%
\pgftext[x=0.786051in,y=0.701639in,right,]{\color{textcolor}\rmfamily\fontsize{10.000000}{12.000000}\selectfont Agreement}%
\end{pgfscope}%
\begin{pgfscope}%
\pgfsetroundcap%
\pgfsetroundjoin%
\pgfsetlinewidth{1.003750pt}%
\definecolor{currentstroke}{rgb}{0.000000,0.000000,0.000000}%
\pgfsetstrokecolor{currentstroke}%
\pgfsetdash{}{0pt}%
\pgfpathmoveto{\pgfqpoint{0.841659in}{0.443680in}}%
\pgfpathlineto{\pgfqpoint{1.185338in}{0.443680in}}%
\pgfpathlineto{\pgfqpoint{1.193719in}{0.544972in}}%
\pgfusepath{stroke}%
\end{pgfscope}%
\begin{pgfscope}%
\definecolor{textcolor}{rgb}{0.000000,0.000000,0.000000}%
\pgfsetstrokecolor{textcolor}%
\pgfsetfillcolor{textcolor}%
\pgftext[x=0.786051in,y=0.443680in,right,]{\color{textcolor}\rmfamily\fontsize{10.000000}{12.000000}\selectfont Backchannel}%
\end{pgfscope}%
\begin{pgfscope}%
\pgfsetroundcap%
\pgfsetroundjoin%
\pgfsetlinewidth{1.003750pt}%
\definecolor{currentstroke}{rgb}{0.000000,0.000000,0.000000}%
\pgfsetstrokecolor{currentstroke}%
\pgfsetdash{}{0pt}%
\pgfpathmoveto{\pgfqpoint{1.603700in}{0.527737in}}%
\pgfpathlineto{\pgfqpoint{1.487608in}{0.527737in}}%
\pgfpathlineto{\pgfqpoint{1.428138in}{0.610161in}}%
\pgfusepath{stroke}%
\end{pgfscope}%
\begin{pgfscope}%
\definecolor{textcolor}{rgb}{0.000000,0.000000,0.000000}%
\pgfsetstrokecolor{textcolor}%
\pgfsetfillcolor{textcolor}%
\pgftext[x=1.659296in,y=0.527737in,left,]{\color{textcolor}\rmfamily\fontsize{10.000000}{12.000000}\selectfont N/A}%
\end{pgfscope}%
\begin{pgfscope}%
\pgfsetroundcap%
\pgfsetroundjoin%
\pgfsetlinewidth{1.003750pt}%
\definecolor{currentstroke}{rgb}{0.000000,0.000000,0.000000}%
\pgfsetstrokecolor{currentstroke}%
\pgfsetdash{}{0pt}%
\pgfpathmoveto{\pgfqpoint{1.603741in}{0.647607in}}%
\pgfpathlineto{\pgfqpoint{1.601952in}{0.647607in}}%
\pgfpathlineto{\pgfqpoint{1.516815in}{0.703124in}}%
\pgfusepath{stroke}%
\end{pgfscope}%
\begin{pgfscope}%
\definecolor{textcolor}{rgb}{0.000000,0.000000,0.000000}%
\pgfsetstrokecolor{textcolor}%
\pgfsetfillcolor{textcolor}%
\pgftext[x=1.659296in,y=0.647607in,left,]{\color{textcolor}\rmfamily\fontsize{10.000000}{12.000000}\selectfont Other}%
\end{pgfscope}%
\begin{pgfscope}%
\pgfsetbuttcap%
\pgfsetmiterjoin%
\definecolor{currentfill}{rgb}{1.000000,1.000000,1.000000}%
\pgfsetfillcolor{currentfill}%
\pgfsetlinewidth{1.003750pt}%
\definecolor{currentstroke}{rgb}{1.000000,1.000000,1.000000}%
\pgfsetstrokecolor{currentstroke}%
\pgfsetdash{}{0pt}%
\pgfpathmoveto{\pgfqpoint{1.515795in}{0.758247in}}%
\pgfpathcurveto{\pgfqpoint{1.536515in}{0.802681in}}{\pgfqpoint{1.546863in}{0.851250in}}{\pgfqpoint{1.546053in}{0.900271in}}%
\pgfpathcurveto{\pgfqpoint{1.545244in}{0.949292in}}{\pgfqpoint{1.533298in}{0.997493in}}{\pgfqpoint{1.511123in}{1.041219in}}%
\pgfpathcurveto{\pgfqpoint{1.488947in}{1.084945in}}{\pgfqpoint{1.457117in}{1.123061in}}{\pgfqpoint{1.418045in}{1.152677in}}%
\pgfpathcurveto{\pgfqpoint{1.378974in}{1.182294in}}{\pgfqpoint{1.333675in}{1.202642in}}{\pgfqpoint{1.285583in}{1.212178in}}%
\pgfpathlineto{\pgfqpoint{1.235256in}{0.958381in}}%
\pgfpathcurveto{\pgfqpoint{1.244874in}{0.956474in}}{\pgfqpoint{1.253934in}{0.952404in}}{\pgfqpoint{1.261748in}{0.946481in}}%
\pgfpathcurveto{\pgfqpoint{1.269562in}{0.940558in}}{\pgfqpoint{1.275928in}{0.932934in}}{\pgfqpoint{1.280363in}{0.924189in}}%
\pgfpathcurveto{\pgfqpoint{1.284799in}{0.915444in}}{\pgfqpoint{1.287188in}{0.905804in}}{\pgfqpoint{1.287350in}{0.896000in}}%
\pgfpathcurveto{\pgfqpoint{1.287511in}{0.886195in}}{\pgfqpoint{1.285442in}{0.876481in}}{\pgfqpoint{1.281298in}{0.867595in}}%
\pgfpathlineto{\pgfqpoint{1.515795in}{0.758247in}}%
\pgfpathclose%
\pgfusepath{stroke,fill}%
\end{pgfscope}%
\begin{pgfscope}%
\pgfsetbuttcap%
\pgfsetmiterjoin%
\definecolor{currentfill}{rgb}{0.592157,0.941176,0.666667}%
\pgfsetfillcolor{currentfill}%
\pgfsetlinewidth{0.000000pt}%
\definecolor{currentstroke}{rgb}{0.000000,0.000000,0.000000}%
\pgfsetstrokecolor{currentstroke}%
\pgfsetstrokeopacity{0.000000}%
\pgfsetdash{}{0pt}%
\pgfpathmoveto{\pgfqpoint{1.285583in}{1.212178in}}%
\pgfpathcurveto{\pgfqpoint{1.242034in}{1.220814in}}{\pgfqpoint{1.197168in}{1.220391in}}{\pgfqpoint{1.153789in}{1.210935in}}%
\pgfpathcurveto{\pgfqpoint{1.110410in}{1.201479in}}{\pgfqpoint{1.069439in}{1.183191in}}{\pgfqpoint{1.033435in}{1.157213in}}%
\pgfpathcurveto{\pgfqpoint{0.997430in}{1.131236in}}{\pgfqpoint{0.967158in}{1.098120in}}{\pgfqpoint{0.944507in}{1.059935in}}%
\pgfpathcurveto{\pgfqpoint{0.921856in}{1.021750in}}{\pgfqpoint{0.907309in}{0.979307in}}{\pgfqpoint{0.901773in}{0.935255in}}%
\pgfpathlineto{\pgfqpoint{1.158494in}{0.902996in}}%
\pgfpathcurveto{\pgfqpoint{1.159601in}{0.911807in}}{\pgfqpoint{1.162510in}{0.920295in}}{\pgfqpoint{1.167040in}{0.927932in}}%
\pgfpathcurveto{\pgfqpoint{1.171570in}{0.935569in}}{\pgfqpoint{1.177625in}{0.942192in}}{\pgfqpoint{1.184826in}{0.947388in}}%
\pgfpathcurveto{\pgfqpoint{1.192027in}{0.952583in}}{\pgfqpoint{1.200221in}{0.956241in}}{\pgfqpoint{1.208897in}{0.958132in}}%
\pgfpathcurveto{\pgfqpoint{1.217573in}{0.960023in}}{\pgfqpoint{1.226546in}{0.960108in}}{\pgfqpoint{1.235256in}{0.958381in}}%
\pgfpathlineto{\pgfqpoint{1.285583in}{1.212178in}}%
\pgfpathclose%
\pgfusepath{fill}%
\end{pgfscope}%
\begin{pgfscope}%
\pgfsetbuttcap%
\pgfsetmiterjoin%
\definecolor{currentfill}{rgb}{1.000000,0.623529,0.603922}%
\pgfsetfillcolor{currentfill}%
\pgfsetlinewidth{0.000000pt}%
\definecolor{currentstroke}{rgb}{0.000000,0.000000,0.000000}%
\pgfsetstrokecolor{currentstroke}%
\pgfsetstrokeopacity{0.000000}%
\pgfsetdash{}{0pt}%
\pgfpathmoveto{\pgfqpoint{0.901773in}{0.935255in}}%
\pgfpathcurveto{\pgfqpoint{0.894114in}{0.874305in}}{\pgfqpoint{0.903977in}{0.812416in}}{\pgfqpoint{0.930200in}{0.756866in}}%
\pgfpathcurveto{\pgfqpoint{0.956424in}{0.701315in}}{\pgfqpoint{0.997939in}{0.654368in}}{\pgfqpoint{1.049864in}{0.621546in}}%
\pgfpathlineto{\pgfqpoint{1.188112in}{0.840254in}}%
\pgfpathcurveto{\pgfqpoint{1.177727in}{0.846819in}}{\pgfqpoint{1.169424in}{0.856208in}}{\pgfqpoint{1.164179in}{0.867318in}}%
\pgfpathcurveto{\pgfqpoint{1.158934in}{0.878429in}}{\pgfqpoint{1.156962in}{0.890806in}}{\pgfqpoint{1.158494in}{0.902996in}}%
\pgfpathlineto{\pgfqpoint{0.901773in}{0.935255in}}%
\pgfpathclose%
\pgfusepath{fill}%
\end{pgfscope}%
\begin{pgfscope}%
\pgfsetbuttcap%
\pgfsetmiterjoin%
\definecolor{currentfill}{rgb}{0.815686,0.733333,1.000000}%
\pgfsetfillcolor{currentfill}%
\pgfsetlinewidth{0.000000pt}%
\definecolor{currentstroke}{rgb}{0.000000,0.000000,0.000000}%
\pgfsetstrokecolor{currentstroke}%
\pgfsetstrokeopacity{0.000000}%
\pgfsetdash{}{0pt}%
\pgfpathmoveto{\pgfqpoint{1.049864in}{0.621546in}}%
\pgfpathcurveto{\pgfqpoint{1.093921in}{0.593697in}}{\pgfqpoint{1.144063in}{0.576907in}}{\pgfqpoint{1.196005in}{0.572609in}}%
\pgfpathcurveto{\pgfqpoint{1.247948in}{0.568312in}}{\pgfqpoint{1.300168in}{0.576633in}}{\pgfqpoint{1.348202in}{0.596862in}}%
\pgfpathlineto{\pgfqpoint{1.247779in}{0.835318in}}%
\pgfpathcurveto{\pgfqpoint{1.238172in}{0.831272in}}{\pgfqpoint{1.227728in}{0.829608in}}{\pgfqpoint{1.217340in}{0.830467in}}%
\pgfpathcurveto{\pgfqpoint{1.206951in}{0.831327in}}{\pgfqpoint{1.196923in}{0.834685in}}{\pgfqpoint{1.188112in}{0.840254in}}%
\pgfpathlineto{\pgfqpoint{1.049864in}{0.621546in}}%
\pgfpathclose%
\pgfusepath{fill}%
\end{pgfscope}%
\begin{pgfscope}%
\pgfsetbuttcap%
\pgfsetmiterjoin%
\definecolor{currentfill}{rgb}{1.000000,0.996078,0.639216}%
\pgfsetfillcolor{currentfill}%
\pgfsetlinewidth{0.000000pt}%
\definecolor{currentstroke}{rgb}{0.000000,0.000000,0.000000}%
\pgfsetstrokecolor{currentstroke}%
\pgfsetstrokeopacity{0.000000}%
\pgfsetdash{}{0pt}%
\pgfpathmoveto{\pgfqpoint{1.348202in}{0.596862in}}%
\pgfpathcurveto{\pgfqpoint{1.370723in}{0.606346in}}{\pgfqpoint{1.392096in}{0.618352in}}{\pgfqpoint{1.411912in}{0.632650in}}%
\pgfpathcurveto{\pgfqpoint{1.431729in}{0.646948in}}{\pgfqpoint{1.449861in}{0.663445in}}{\pgfqpoint{1.465962in}{0.681827in}}%
\pgfpathlineto{\pgfqpoint{1.271331in}{0.852311in}}%
\pgfpathcurveto{\pgfqpoint{1.268111in}{0.848634in}}{\pgfqpoint{1.264485in}{0.845335in}}{\pgfqpoint{1.260521in}{0.842475in}}%
\pgfpathcurveto{\pgfqpoint{1.256558in}{0.839616in}}{\pgfqpoint{1.252283in}{0.837215in}}{\pgfqpoint{1.247779in}{0.835318in}}%
\pgfpathlineto{\pgfqpoint{1.348202in}{0.596862in}}%
\pgfpathclose%
\pgfusepath{fill}%
\end{pgfscope}%
\begin{pgfscope}%
\pgfsetbuttcap%
\pgfsetmiterjoin%
\definecolor{currentfill}{rgb}{0.690196,0.878431,0.901961}%
\pgfsetfillcolor{currentfill}%
\pgfsetlinewidth{0.000000pt}%
\definecolor{currentstroke}{rgb}{0.000000,0.000000,0.000000}%
\pgfsetstrokecolor{currentstroke}%
\pgfsetstrokeopacity{0.000000}%
\pgfsetdash{}{0pt}%
\pgfpathmoveto{\pgfqpoint{1.465962in}{0.681827in}}%
\pgfpathcurveto{\pgfqpoint{1.476019in}{0.693308in}}{\pgfqpoint{1.485250in}{0.705486in}}{\pgfqpoint{1.493587in}{0.718271in}}%
\pgfpathcurveto{\pgfqpoint{1.501924in}{0.731055in}}{\pgfqpoint{1.509345in}{0.744414in}}{\pgfqpoint{1.515795in}{0.758247in}}%
\pgfpathlineto{\pgfqpoint{1.281298in}{0.867595in}}%
\pgfpathcurveto{\pgfqpoint{1.280008in}{0.864828in}}{\pgfqpoint{1.278524in}{0.862156in}}{\pgfqpoint{1.276856in}{0.859599in}}%
\pgfpathcurveto{\pgfqpoint{1.275189in}{0.857043in}}{\pgfqpoint{1.273343in}{0.854607in}}{\pgfqpoint{1.271331in}{0.852311in}}%
\pgfpathlineto{\pgfqpoint{1.465962in}{0.681827in}}%
\pgfpathclose%
\pgfusepath{fill}%
\end{pgfscope}%
\begin{pgfscope}%
\definecolor{textcolor}{rgb}{0.000000,0.000000,0.000000}%
\pgfsetstrokecolor{textcolor}%
\pgfsetfillcolor{textcolor}%
\pgftext[x=1.395743in,y=0.982704in,,]{\color{textcolor}\rmfamily\fontsize{6.000000}{7.200000}\selectfont 29\%}%
\end{pgfscope}%
\begin{pgfscope}%
\definecolor{textcolor}{rgb}{0.000000,0.000000,0.000000}%
\pgfsetstrokecolor{textcolor}%
\pgfsetfillcolor{textcolor}%
\pgftext[x=1.109130in,y=1.052301in,,]{\color{textcolor}\rmfamily\fontsize{6.000000}{7.200000}\selectfont 26\%}%
\end{pgfscope}%
\begin{pgfscope}%
\definecolor{textcolor}{rgb}{0.000000,0.000000,0.000000}%
\pgfsetstrokecolor{textcolor}%
\pgfsetfillcolor{textcolor}%
\pgftext[x=1.047189in,y=0.812092in,,]{\color{textcolor}\rmfamily\fontsize{6.000000}{7.200000}\selectfont 18\%}%
\end{pgfscope}%
\begin{pgfscope}%
\definecolor{textcolor}{rgb}{0.000000,0.000000,0.000000}%
\pgfsetstrokecolor{textcolor}%
\pgfsetfillcolor{textcolor}%
\pgftext[x=1.206673in,y=0.701538in,,]{\color{textcolor}\rmfamily\fontsize{6.000000}{7.200000}\selectfont 15\%}%
\end{pgfscope}%
\begin{pgfscope}%
\definecolor{textcolor}{rgb}{0.000000,0.000000,0.000000}%
\pgfsetstrokecolor{textcolor}%
\pgfsetfillcolor{textcolor}%
\pgftext[x=1.336217in,y=0.737563in,,]{\color{textcolor}\rmfamily\fontsize{6.000000}{7.200000}\selectfont 7\%}%
\end{pgfscope}%
\begin{pgfscope}%
\definecolor{textcolor}{rgb}{0.000000,0.000000,0.000000}%
\pgfsetstrokecolor{textcolor}%
\pgfsetfillcolor{textcolor}%
\pgftext[x=1.385222in,y=0.788935in,,]{\color{textcolor}\rmfamily\fontsize{6.000000}{7.200000}\selectfont 5\%}%
\end{pgfscope}%
\begin{pgfscope}%
\definecolor{textcolor}{rgb}{0.000000,0.000000,0.000000}%
\pgfsetstrokecolor{textcolor}%
\pgfsetfillcolor{textcolor}%
\pgftext[x=1.222674in,y=0.129277in,,base]{\color{textcolor}\rmfamily\fontsize{11.000000}{13.200000}\selectfont Acknowledgement}%
\end{pgfscope}%
\begin{pgfscope}%
\pgfsetroundcap%
\pgfsetroundjoin%
\pgfsetlinewidth{1.003750pt}%
\definecolor{currentstroke}{rgb}{0.000000,0.000000,0.000000}%
\pgfsetstrokecolor{currentstroke}%
\pgfsetdash{}{0pt}%
\pgfpathmoveto{\pgfqpoint{3.641275in}{0.496551in}}%
\pgfpathlineto{\pgfqpoint{3.475443in}{0.496551in}}%
\pgfpathlineto{\pgfqpoint{3.427135in}{0.585976in}}%
\pgfusepath{stroke}%
\end{pgfscope}%
\begin{pgfscope}%
\definecolor{textcolor}{rgb}{0.000000,0.000000,0.000000}%
\pgfsetstrokecolor{textcolor}%
\pgfsetfillcolor{textcolor}%
\pgftext[x=3.696856in, y=0.533202in, left, base]{\color{textcolor}\rmfamily\fontsize{10.000000}{12.000000}\selectfont Elaborate}%
\end{pgfscope}%
\begin{pgfscope}%
\definecolor{textcolor}{rgb}{0.000000,0.000000,0.000000}%
\pgfsetstrokecolor{textcolor}%
\pgfsetfillcolor{textcolor}%
\pgftext[x=3.696856in, y=0.390456in, left, base]{\color{textcolor}\rmfamily\fontsize{10.000000}{12.000000}\selectfont Self}%
\end{pgfscope}%
\begin{pgfscope}%
\pgfsetroundcap%
\pgfsetroundjoin%
\pgfsetlinewidth{1.003750pt}%
\definecolor{currentstroke}{rgb}{0.000000,0.000000,0.000000}%
\pgfsetstrokecolor{currentstroke}%
\pgfsetdash{}{0pt}%
\pgfpathmoveto{\pgfqpoint{3.641320in}{1.158999in}}%
\pgfpathlineto{\pgfqpoint{3.628052in}{1.158999in}}%
\pgfpathlineto{\pgfqpoint{3.545488in}{1.099724in}}%
\pgfusepath{stroke}%
\end{pgfscope}%
\begin{pgfscope}%
\definecolor{textcolor}{rgb}{0.000000,0.000000,0.000000}%
\pgfsetstrokecolor{textcolor}%
\pgfsetfillcolor{textcolor}%
\pgftext[x=3.696856in, y=1.195650in, left, base]{\color{textcolor}\rmfamily\fontsize{10.000000}{12.000000}\selectfont Elaborate}%
\end{pgfscope}%
\begin{pgfscope}%
\definecolor{textcolor}{rgb}{0.000000,0.000000,0.000000}%
\pgfsetstrokecolor{textcolor}%
\pgfsetfillcolor{textcolor}%
\pgftext[x=3.696856in, y=1.052903in, left, base]{\color{textcolor}\rmfamily\fontsize{10.000000}{12.000000}\selectfont Other}%
\end{pgfscope}%
\begin{pgfscope}%
\pgfsetroundcap%
\pgfsetroundjoin%
\pgfsetlinewidth{1.003750pt}%
\definecolor{currentstroke}{rgb}{0.000000,0.000000,0.000000}%
\pgfsetstrokecolor{currentstroke}%
\pgfsetdash{}{0pt}%
\pgfpathmoveto{\pgfqpoint{2.879217in}{1.315074in}}%
\pgfpathlineto{\pgfqpoint{3.091407in}{1.315074in}}%
\pgfpathlineto{\pgfqpoint{3.129304in}{1.220764in}}%
\pgfusepath{stroke}%
\end{pgfscope}%
\begin{pgfscope}%
\definecolor{textcolor}{rgb}{0.000000,0.000000,0.000000}%
\pgfsetstrokecolor{textcolor}%
\pgfsetfillcolor{textcolor}%
\pgftext[x=2.823612in,y=1.315074in,right,]{\color{textcolor}\rmfamily\fontsize{10.000000}{12.000000}\selectfont Commonality}%
\end{pgfscope}%
\begin{pgfscope}%
\pgfsetroundcap%
\pgfsetroundjoin%
\pgfsetlinewidth{1.003750pt}%
\definecolor{currentstroke}{rgb}{0.000000,0.000000,0.000000}%
\pgfsetstrokecolor{currentstroke}%
\pgfsetdash{}{0pt}%
\pgfpathmoveto{\pgfqpoint{2.877869in}{1.017974in}}%
\pgfpathlineto{\pgfqpoint{2.925916in}{1.002513in}}%
\pgfusepath{stroke}%
\end{pgfscope}%
\begin{pgfscope}%
\definecolor{textcolor}{rgb}{0.000000,0.000000,0.000000}%
\pgfsetstrokecolor{textcolor}%
\pgfsetfillcolor{textcolor}%
\pgftext[x=2.823612in,y=1.033633in,right,]{\color{textcolor}\rmfamily\fontsize{10.000000}{12.000000}\selectfont Differences}%
\end{pgfscope}%
\begin{pgfscope}%
\pgfsetroundcap%
\pgfsetroundjoin%
\pgfsetlinewidth{1.003750pt}%
\definecolor{currentstroke}{rgb}{0.000000,0.000000,0.000000}%
\pgfsetstrokecolor{currentstroke}%
\pgfsetdash{}{0pt}%
\pgfpathmoveto{\pgfqpoint{2.877869in}{0.771889in}}%
\pgfpathlineto{\pgfqpoint{2.925916in}{0.787350in}}%
\pgfusepath{stroke}%
\end{pgfscope}%
\begin{pgfscope}%
\definecolor{textcolor}{rgb}{0.000000,0.000000,0.000000}%
\pgfsetstrokecolor{textcolor}%
\pgfsetfillcolor{textcolor}%
\pgftext[x=2.190509in, y=0.792881in, left, base]{\color{textcolor}\rmfamily\fontsize{10.000000}{12.000000}\selectfont Discussion}%
\end{pgfscope}%
\begin{pgfscope}%
\definecolor{textcolor}{rgb}{0.000000,0.000000,0.000000}%
\pgfsetstrokecolor{textcolor}%
\pgfsetfillcolor{textcolor}%
\pgftext[x=2.406945in, y=0.650135in, left, base]{\color{textcolor}\rmfamily\fontsize{10.000000}{12.000000}\selectfont Theme}%
\end{pgfscope}%
\begin{pgfscope}%
\pgfsetroundcap%
\pgfsetroundjoin%
\pgfsetlinewidth{1.003750pt}%
\definecolor{currentstroke}{rgb}{0.000000,0.000000,0.000000}%
\pgfsetstrokecolor{currentstroke}%
\pgfsetdash{}{0pt}%
\pgfpathmoveto{\pgfqpoint{2.879182in}{0.546703in}}%
\pgfpathlineto{\pgfqpoint{2.970823in}{0.546703in}}%
\pgfpathlineto{\pgfqpoint{3.035787in}{0.624870in}}%
\pgfusepath{stroke}%
\end{pgfscope}%
\begin{pgfscope}%
\definecolor{textcolor}{rgb}{0.000000,0.000000,0.000000}%
\pgfsetstrokecolor{textcolor}%
\pgfsetfillcolor{textcolor}%
\pgftext[x=2.823612in,y=0.546703in,right,]{\color{textcolor}\rmfamily\fontsize{10.000000}{12.000000}\selectfont None}%
\end{pgfscope}%
\begin{pgfscope}%
\pgfsetbuttcap%
\pgfsetmiterjoin%
\definecolor{currentfill}{rgb}{0.572549,0.776471,1.000000}%
\pgfsetfillcolor{currentfill}%
\pgfsetlinewidth{0.000000pt}%
\definecolor{currentstroke}{rgb}{0.000000,0.000000,0.000000}%
\pgfsetstrokecolor{currentstroke}%
\pgfsetstrokeopacity{0.000000}%
\pgfsetdash{}{0pt}%
\pgfpathmoveto{\pgfqpoint{3.098522in}{0.614838in}}%
\pgfpathcurveto{\pgfqpoint{3.146342in}{0.587229in}}{\pgfqpoint{3.200445in}{0.572322in}}{\pgfqpoint{3.255657in}{0.571540in}}%
\pgfpathcurveto{\pgfqpoint{3.310870in}{0.570759in}}{\pgfqpoint{3.365373in}{0.584130in}}{\pgfqpoint{3.413955in}{0.610374in}}%
\pgfpathcurveto{\pgfqpoint{3.462537in}{0.636619in}}{\pgfqpoint{3.503598in}{0.674873in}}{\pgfqpoint{3.533212in}{0.721478in}}%
\pgfpathcurveto{\pgfqpoint{3.562825in}{0.768083in}}{\pgfqpoint{3.580016in}{0.821505in}}{\pgfqpoint{3.583140in}{0.876634in}}%
\pgfpathlineto{\pgfqpoint{3.324815in}{0.891272in}}%
\pgfpathcurveto{\pgfqpoint{3.324190in}{0.880246in}}{\pgfqpoint{3.320752in}{0.869562in}}{\pgfqpoint{3.314830in}{0.860241in}}%
\pgfpathcurveto{\pgfqpoint{3.308907in}{0.850920in}}{\pgfqpoint{3.300695in}{0.843269in}}{\pgfqpoint{3.290978in}{0.838020in}}%
\pgfpathcurveto{\pgfqpoint{3.281262in}{0.832771in}}{\pgfqpoint{3.270361in}{0.830097in}}{\pgfqpoint{3.259319in}{0.830253in}}%
\pgfpathcurveto{\pgfqpoint{3.248276in}{0.830410in}}{\pgfqpoint{3.237456in}{0.833391in}}{\pgfqpoint{3.227892in}{0.838913in}}%
\pgfpathlineto{\pgfqpoint{3.098522in}{0.614838in}}%
\pgfpathclose%
\pgfusepath{fill}%
\end{pgfscope}%
\begin{pgfscope}%
\pgfsetbuttcap%
\pgfsetmiterjoin%
\definecolor{currentfill}{rgb}{0.592157,0.941176,0.666667}%
\pgfsetfillcolor{currentfill}%
\pgfsetlinewidth{0.000000pt}%
\definecolor{currentstroke}{rgb}{0.000000,0.000000,0.000000}%
\pgfsetstrokecolor{currentstroke}%
\pgfsetstrokeopacity{0.000000}%
\pgfsetdash{}{0pt}%
\pgfpathmoveto{\pgfqpoint{3.583140in}{0.876634in}}%
\pgfpathcurveto{\pgfqpoint{3.587322in}{0.950442in}}{\pgfqpoint{3.566075in}{1.023499in}}{\pgfqpoint{3.522961in}{1.083551in}}%
\pgfpathcurveto{\pgfqpoint{3.479848in}{1.143604in}}{\pgfqpoint{3.417421in}{1.187097in}}{\pgfqpoint{3.346151in}{1.206735in}}%
\pgfpathlineto{\pgfqpoint{3.277417in}{0.957292in}}%
\pgfpathcurveto{\pgfqpoint{3.291671in}{0.953365in}}{\pgfqpoint{3.304157in}{0.944666in}}{\pgfqpoint{3.312780in}{0.932656in}}%
\pgfpathcurveto{\pgfqpoint{3.321402in}{0.920645in}}{\pgfqpoint{3.325652in}{0.906034in}}{\pgfqpoint{3.324815in}{0.891272in}}%
\pgfpathlineto{\pgfqpoint{3.583140in}{0.876634in}}%
\pgfpathclose%
\pgfusepath{fill}%
\end{pgfscope}%
\begin{pgfscope}%
\pgfsetbuttcap%
\pgfsetmiterjoin%
\definecolor{currentfill}{rgb}{1.000000,0.623529,0.603922}%
\pgfsetfillcolor{currentfill}%
\pgfsetlinewidth{0.000000pt}%
\definecolor{currentstroke}{rgb}{0.000000,0.000000,0.000000}%
\pgfsetstrokecolor{currentstroke}%
\pgfsetstrokeopacity{0.000000}%
\pgfsetdash{}{0pt}%
\pgfpathmoveto{\pgfqpoint{3.346151in}{1.206735in}}%
\pgfpathcurveto{\pgfqpoint{3.277902in}{1.225541in}}{\pgfqpoint{3.205331in}{1.221428in}}{\pgfqpoint{3.139643in}{1.195033in}}%
\pgfpathcurveto{\pgfqpoint{3.073956in}{1.168638in}}{\pgfqpoint{3.018717in}{1.121392in}}{\pgfqpoint{2.982457in}{1.060591in}}%
\pgfpathlineto{\pgfqpoint{3.204679in}{0.928063in}}%
\pgfpathcurveto{\pgfqpoint{3.211931in}{0.940224in}}{\pgfqpoint{3.222978in}{0.949673in}}{\pgfqpoint{3.236116in}{0.954952in}}%
\pgfpathcurveto{\pgfqpoint{3.249253in}{0.960231in}}{\pgfqpoint{3.263768in}{0.961053in}}{\pgfqpoint{3.277417in}{0.957292in}}%
\pgfpathlineto{\pgfqpoint{3.346151in}{1.206735in}}%
\pgfpathclose%
\pgfusepath{fill}%
\end{pgfscope}%
\begin{pgfscope}%
\pgfsetbuttcap%
\pgfsetmiterjoin%
\definecolor{currentfill}{rgb}{0.815686,0.733333,1.000000}%
\pgfsetfillcolor{currentfill}%
\pgfsetlinewidth{0.000000pt}%
\definecolor{currentstroke}{rgb}{0.000000,0.000000,0.000000}%
\pgfsetstrokecolor{currentstroke}%
\pgfsetstrokeopacity{0.000000}%
\pgfsetdash{}{0pt}%
\pgfpathmoveto{\pgfqpoint{2.982457in}{1.060591in}}%
\pgfpathcurveto{\pgfqpoint{2.969941in}{1.039604in}}{\pgfqpoint{2.959843in}{1.017265in}}{\pgfqpoint{2.952358in}{0.994004in}}%
\pgfpathcurveto{\pgfqpoint{2.944873in}{0.970743in}}{\pgfqpoint{2.940048in}{0.946708in}}{\pgfqpoint{2.937975in}{0.922360in}}%
\pgfpathlineto{\pgfqpoint{3.195782in}{0.900417in}}%
\pgfpathcurveto{\pgfqpoint{3.196197in}{0.905287in}}{\pgfqpoint{3.197162in}{0.910094in}}{\pgfqpoint{3.198659in}{0.914746in}}%
\pgfpathcurveto{\pgfqpoint{3.200156in}{0.919398in}}{\pgfqpoint{3.202175in}{0.923866in}}{\pgfqpoint{3.204679in}{0.928063in}}%
\pgfpathlineto{\pgfqpoint{2.982457in}{1.060591in}}%
\pgfpathclose%
\pgfusepath{fill}%
\end{pgfscope}%
\begin{pgfscope}%
\pgfsetbuttcap%
\pgfsetmiterjoin%
\definecolor{currentfill}{rgb}{1.000000,0.996078,0.639216}%
\pgfsetfillcolor{currentfill}%
\pgfsetlinewidth{0.000000pt}%
\definecolor{currentstroke}{rgb}{0.000000,0.000000,0.000000}%
\pgfsetstrokecolor{currentstroke}%
\pgfsetstrokeopacity{0.000000}%
\pgfsetdash{}{0pt}%
\pgfpathmoveto{\pgfqpoint{2.937975in}{0.922360in}}%
\pgfpathcurveto{\pgfqpoint{2.934341in}{0.879658in}}{\pgfqpoint{2.939230in}{0.836656in}}{\pgfqpoint{2.952358in}{0.795859in}}%
\pgfpathcurveto{\pgfqpoint{2.965486in}{0.755063in}}{\pgfqpoint{2.986593in}{0.717280in}}{\pgfqpoint{3.014449in}{0.684711in}}%
\pgfpathlineto{\pgfqpoint{3.211077in}{0.852887in}}%
\pgfpathcurveto{\pgfqpoint{3.205506in}{0.859401in}}{\pgfqpoint{3.201285in}{0.866958in}}{\pgfqpoint{3.198659in}{0.875117in}}%
\pgfpathcurveto{\pgfqpoint{3.196033in}{0.883276in}}{\pgfqpoint{3.195055in}{0.891877in}}{\pgfqpoint{3.195782in}{0.900417in}}%
\pgfpathlineto{\pgfqpoint{2.937975in}{0.922360in}}%
\pgfpathclose%
\pgfusepath{fill}%
\end{pgfscope}%
\begin{pgfscope}%
\pgfsetbuttcap%
\pgfsetmiterjoin%
\definecolor{currentfill}{rgb}{1.000000,1.000000,1.000000}%
\pgfsetfillcolor{currentfill}%
\pgfsetlinewidth{1.003750pt}%
\definecolor{currentstroke}{rgb}{1.000000,1.000000,1.000000}%
\pgfsetstrokecolor{currentstroke}%
\pgfsetdash{}{0pt}%
\pgfpathmoveto{\pgfqpoint{3.014449in}{0.684711in}}%
\pgfpathcurveto{\pgfqpoint{3.026356in}{0.670790in}}{\pgfqpoint{3.039424in}{0.657905in}}{\pgfqpoint{3.053512in}{0.646197in}}%
\pgfpathcurveto{\pgfqpoint{3.067600in}{0.634488in}}{\pgfqpoint{3.082658in}{0.623998in}}{\pgfqpoint{3.098522in}{0.614838in}}%
\pgfpathlineto{\pgfqpoint{3.227892in}{0.838913in}}%
\pgfpathcurveto{\pgfqpoint{3.224719in}{0.840745in}}{\pgfqpoint{3.221707in}{0.842843in}}{\pgfqpoint{3.218890in}{0.845185in}}%
\pgfpathcurveto{\pgfqpoint{3.216072in}{0.847526in}}{\pgfqpoint{3.213458in}{0.850103in}}{\pgfqpoint{3.211077in}{0.852887in}}%
\pgfpathlineto{\pgfqpoint{3.014449in}{0.684711in}}%
\pgfpathclose%
\pgfusepath{stroke,fill}%
\end{pgfscope}%
\begin{pgfscope}%
\definecolor{textcolor}{rgb}{0.000000,0.000000,0.000000}%
\pgfsetstrokecolor{textcolor}%
\pgfsetfillcolor{textcolor}%
\pgftext[x=3.352467in,y=0.724197in,,]{\color{textcolor}\rmfamily\fontsize{6.000000}{7.200000}\selectfont 32\%}%
\end{pgfscope}%
\begin{pgfscope}%
\definecolor{textcolor}{rgb}{0.000000,0.000000,0.000000}%
\pgfsetstrokecolor{textcolor}%
\pgfsetfillcolor{textcolor}%
\pgftext[x=3.417870in,y=1.008103in,,]{\color{textcolor}\rmfamily\fontsize{6.000000}{7.200000}\selectfont 22\%}%
\end{pgfscope}%
\begin{pgfscope}%
\definecolor{textcolor}{rgb}{0.000000,0.000000,0.000000}%
\pgfsetstrokecolor{textcolor}%
\pgfsetfillcolor{textcolor}%
\pgftext[x=3.187880in,y=1.074993in,,]{\color{textcolor}\rmfamily\fontsize{6.000000}{7.200000}\selectfont 21\%}%
\end{pgfscope}%
\begin{pgfscope}%
\definecolor{textcolor}{rgb}{0.000000,0.000000,0.000000}%
\pgfsetstrokecolor{textcolor}%
\pgfsetfillcolor{textcolor}%
\pgftext[x=3.075508in,y=0.954375in,,]{\color{textcolor}\rmfamily\fontsize{6.000000}{7.200000}\selectfont 7\%}%
\end{pgfscope}%
\begin{pgfscope}%
\definecolor{textcolor}{rgb}{0.000000,0.000000,0.000000}%
\pgfsetstrokecolor{textcolor}%
\pgfsetfillcolor{textcolor}%
\pgftext[x=3.075508in,y=0.835488in,,]{\color{textcolor}\rmfamily\fontsize{6.000000}{7.200000}\selectfont 13\%}%
\end{pgfscope}%
\begin{pgfscope}%
\definecolor{textcolor}{rgb}{0.000000,0.000000,0.000000}%
\pgfsetstrokecolor{textcolor}%
\pgfsetfillcolor{textcolor}%
\pgftext[x=3.136201in,y=0.745691in,,]{\color{textcolor}\rmfamily\fontsize{6.000000}{7.200000}\selectfont 5\%}%
\end{pgfscope}%
\begin{pgfscope}%
\definecolor{textcolor}{rgb}{0.000000,0.000000,0.000000}%
\pgfsetstrokecolor{textcolor}%
\pgfsetfillcolor{textcolor}%
\pgftext[x=3.260234in,y=0.129277in,,base]{\color{textcolor}\rmfamily\fontsize{11.000000}{13.200000}\selectfont Transition}%
\end{pgfscope}%
\begin{pgfscope}%
\pgfsetbuttcap%
\pgfsetmiterjoin%
\definecolor{currentfill}{rgb}{1.000000,1.000000,1.000000}%
\pgfsetfillcolor{currentfill}%
\pgfsetlinewidth{0.000000pt}%
\definecolor{currentstroke}{rgb}{0.000000,0.000000,0.000000}%
\pgfsetstrokecolor{currentstroke}%
\pgfsetstrokeopacity{0.000000}%
\pgfsetdash{}{0pt}%
\pgfpathmoveto{\pgfqpoint{5.271468in}{0.467175in}}%
\pgfpathlineto{\pgfqpoint{5.918315in}{0.467175in}}%
\pgfpathlineto{\pgfqpoint{5.918315in}{1.322688in}}%
\pgfpathlineto{\pgfqpoint{5.271468in}{1.322688in}}%
\pgfpathclose%
\pgfusepath{fill}%
\end{pgfscope}%
\begin{pgfscope}%
\pgfpathrectangle{\pgfqpoint{5.271468in}{0.467175in}}{\pgfqpoint{0.646848in}{0.855513in}}%
\pgfusepath{clip}%
\pgfsetbuttcap%
\pgfsetmiterjoin%
\definecolor{currentfill}{rgb}{0.672059,0.789706,0.916176}%
\pgfsetfillcolor{currentfill}%
\pgfsetlinewidth{0.000000pt}%
\definecolor{currentstroke}{rgb}{0.000000,0.000000,0.000000}%
\pgfsetstrokecolor{currentstroke}%
\pgfsetstrokeopacity{0.000000}%
\pgfsetdash{}{0pt}%
\pgfpathmoveto{\pgfqpoint{5.271468in}{1.308429in}}%
\pgfpathlineto{\pgfqpoint{5.844501in}{1.308429in}}%
\pgfpathlineto{\pgfqpoint{5.844501in}{1.194361in}}%
\pgfpathlineto{\pgfqpoint{5.271468in}{1.194361in}}%
\pgfpathclose%
\pgfusepath{fill}%
\end{pgfscope}%
\begin{pgfscope}%
\pgfpathrectangle{\pgfqpoint{5.271468in}{0.467175in}}{\pgfqpoint{0.646848in}{0.855513in}}%
\pgfusepath{clip}%
\pgfsetbuttcap%
\pgfsetmiterjoin%
\definecolor{currentfill}{rgb}{0.938725,0.718137,0.571078}%
\pgfsetfillcolor{currentfill}%
\pgfsetlinewidth{0.000000pt}%
\definecolor{currentstroke}{rgb}{0.000000,0.000000,0.000000}%
\pgfsetstrokecolor{currentstroke}%
\pgfsetstrokeopacity{0.000000}%
\pgfsetdash{}{0pt}%
\pgfpathmoveto{\pgfqpoint{5.271468in}{1.165844in}}%
\pgfpathlineto{\pgfqpoint{5.640540in}{1.165844in}}%
\pgfpathlineto{\pgfqpoint{5.640540in}{1.051776in}}%
\pgfpathlineto{\pgfqpoint{5.271468in}{1.051776in}}%
\pgfpathclose%
\pgfusepath{fill}%
\end{pgfscope}%
\begin{pgfscope}%
\pgfpathrectangle{\pgfqpoint{5.271468in}{0.467175in}}{\pgfqpoint{0.646848in}{0.855513in}}%
\pgfusepath{clip}%
\pgfsetbuttcap%
\pgfsetmiterjoin%
\definecolor{currentfill}{rgb}{0.596078,0.854902,0.654902}%
\pgfsetfillcolor{currentfill}%
\pgfsetlinewidth{0.000000pt}%
\definecolor{currentstroke}{rgb}{0.000000,0.000000,0.000000}%
\pgfsetstrokecolor{currentstroke}%
\pgfsetstrokeopacity{0.000000}%
\pgfsetdash{}{0pt}%
\pgfpathmoveto{\pgfqpoint{5.271468in}{1.023259in}}%
\pgfpathlineto{\pgfqpoint{5.446291in}{1.023259in}}%
\pgfpathlineto{\pgfqpoint{5.446291in}{0.909190in}}%
\pgfpathlineto{\pgfqpoint{5.271468in}{0.909190in}}%
\pgfpathclose%
\pgfusepath{fill}%
\end{pgfscope}%
\begin{pgfscope}%
\pgfpathrectangle{\pgfqpoint{5.271468in}{0.467175in}}{\pgfqpoint{0.646848in}{0.855513in}}%
\pgfusepath{clip}%
\pgfsetbuttcap%
\pgfsetmiterjoin%
\definecolor{currentfill}{rgb}{0.950980,0.668627,0.656863}%
\pgfsetfillcolor{currentfill}%
\pgfsetlinewidth{0.000000pt}%
\definecolor{currentstroke}{rgb}{0.000000,0.000000,0.000000}%
\pgfsetstrokecolor{currentstroke}%
\pgfsetstrokeopacity{0.000000}%
\pgfsetdash{}{0pt}%
\pgfpathmoveto{\pgfqpoint{5.271468in}{0.880673in}}%
\pgfpathlineto{\pgfqpoint{5.368592in}{0.880673in}}%
\pgfpathlineto{\pgfqpoint{5.368592in}{0.766605in}}%
\pgfpathlineto{\pgfqpoint{5.271468in}{0.766605in}}%
\pgfpathclose%
\pgfusepath{fill}%
\end{pgfscope}%
\begin{pgfscope}%
\pgfpathrectangle{\pgfqpoint{5.271468in}{0.467175in}}{\pgfqpoint{0.646848in}{0.855513in}}%
\pgfusepath{clip}%
\pgfsetbuttcap%
\pgfsetmiterjoin%
\definecolor{currentfill}{rgb}{0.828431,0.766667,0.966667}%
\pgfsetfillcolor{currentfill}%
\pgfsetlinewidth{0.000000pt}%
\definecolor{currentstroke}{rgb}{0.000000,0.000000,0.000000}%
\pgfsetstrokecolor{currentstroke}%
\pgfsetstrokeopacity{0.000000}%
\pgfsetdash{}{0pt}%
\pgfpathmoveto{\pgfqpoint{5.271468in}{0.738088in}}%
\pgfpathlineto{\pgfqpoint{5.349167in}{0.738088in}}%
\pgfpathlineto{\pgfqpoint{5.349167in}{0.624019in}}%
\pgfpathlineto{\pgfqpoint{5.271468in}{0.624019in}}%
\pgfpathclose%
\pgfusepath{fill}%
\end{pgfscope}%
\begin{pgfscope}%
\pgfpathrectangle{\pgfqpoint{5.271468in}{0.467175in}}{\pgfqpoint{0.646848in}{0.855513in}}%
\pgfusepath{clip}%
\pgfsetbuttcap%
\pgfsetmiterjoin%
\definecolor{currentfill}{rgb}{0.837745,0.734804,0.640686}%
\pgfsetfillcolor{currentfill}%
\pgfsetlinewidth{0.000000pt}%
\definecolor{currentstroke}{rgb}{0.000000,0.000000,0.000000}%
\pgfsetstrokecolor{currentstroke}%
\pgfsetstrokeopacity{0.000000}%
\pgfsetdash{}{0pt}%
\pgfpathmoveto{\pgfqpoint{5.271468in}{0.595502in}}%
\pgfpathlineto{\pgfqpoint{5.543416in}{0.595502in}}%
\pgfpathlineto{\pgfqpoint{5.543416in}{0.481434in}}%
\pgfpathlineto{\pgfqpoint{5.271468in}{0.481434in}}%
\pgfpathclose%
\pgfusepath{fill}%
\end{pgfscope}%
\begin{pgfscope}%
\pgfsetbuttcap%
\pgfsetroundjoin%
\definecolor{currentfill}{rgb}{0.000000,0.000000,0.000000}%
\pgfsetfillcolor{currentfill}%
\pgfsetlinewidth{0.803000pt}%
\definecolor{currentstroke}{rgb}{0.000000,0.000000,0.000000}%
\pgfsetstrokecolor{currentstroke}%
\pgfsetdash{}{0pt}%
\pgfsys@defobject{currentmarker}{\pgfqpoint{0.000000in}{-0.048611in}}{\pgfqpoint{0.000000in}{0.000000in}}{%
\pgfpathmoveto{\pgfqpoint{0.000000in}{0.000000in}}%
\pgfpathlineto{\pgfqpoint{0.000000in}{-0.048611in}}%
\pgfusepath{stroke,fill}%
}%
\begin{pgfscope}%
\pgfsys@transformshift{5.271468in}{0.467175in}%
\pgfsys@useobject{currentmarker}{}%
\end{pgfscope}%
\end{pgfscope}%
\begin{pgfscope}%
\definecolor{textcolor}{rgb}{0.000000,0.000000,0.000000}%
\pgfsetstrokecolor{textcolor}%
\pgfsetfillcolor{textcolor}%
\pgftext[x=5.271468in,y=0.369953in,,top]{\color{textcolor}\rmfamily\fontsize{10.000000}{12.000000}\selectfont \(\displaystyle {0}\)}%
\end{pgfscope}%
\begin{pgfscope}%
\pgfsetbuttcap%
\pgfsetroundjoin%
\definecolor{currentfill}{rgb}{0.000000,0.000000,0.000000}%
\pgfsetfillcolor{currentfill}%
\pgfsetlinewidth{0.803000pt}%
\definecolor{currentstroke}{rgb}{0.000000,0.000000,0.000000}%
\pgfsetstrokecolor{currentstroke}%
\pgfsetdash{}{0pt}%
\pgfsys@defobject{currentmarker}{\pgfqpoint{0.000000in}{-0.048611in}}{\pgfqpoint{0.000000in}{0.000000in}}{%
\pgfpathmoveto{\pgfqpoint{0.000000in}{0.000000in}}%
\pgfpathlineto{\pgfqpoint{0.000000in}{-0.048611in}}%
\pgfusepath{stroke,fill}%
}%
\begin{pgfscope}%
\pgfsys@transformshift{5.810507in}{0.467175in}%
\pgfsys@useobject{currentmarker}{}%
\end{pgfscope}%
\end{pgfscope}%
\begin{pgfscope}%
\definecolor{textcolor}{rgb}{0.000000,0.000000,0.000000}%
\pgfsetstrokecolor{textcolor}%
\pgfsetfillcolor{textcolor}%
\pgftext[x=5.810507in,y=0.369953in,,top]{\color{textcolor}\rmfamily\fontsize{10.000000}{12.000000}\selectfont \(\displaystyle {50}\)}%
\end{pgfscope}%
\begin{pgfscope}%
\pgfsetbuttcap%
\pgfsetroundjoin%
\definecolor{currentfill}{rgb}{0.000000,0.000000,0.000000}%
\pgfsetfillcolor{currentfill}%
\pgfsetlinewidth{0.803000pt}%
\definecolor{currentstroke}{rgb}{0.000000,0.000000,0.000000}%
\pgfsetstrokecolor{currentstroke}%
\pgfsetdash{}{0pt}%
\pgfsys@defobject{currentmarker}{\pgfqpoint{-0.048611in}{0.000000in}}{\pgfqpoint{-0.000000in}{0.000000in}}{%
\pgfpathmoveto{\pgfqpoint{-0.000000in}{0.000000in}}%
\pgfpathlineto{\pgfqpoint{-0.048611in}{0.000000in}}%
\pgfusepath{stroke,fill}%
}%
\begin{pgfscope}%
\pgfsys@transformshift{5.271468in}{1.251395in}%
\pgfsys@useobject{currentmarker}{}%
\end{pgfscope}%
\end{pgfscope}%
\begin{pgfscope}%
\definecolor{textcolor}{rgb}{0.000000,0.000000,0.000000}%
\pgfsetstrokecolor{textcolor}%
\pgfsetfillcolor{textcolor}%
\pgftext[x=4.508349in, y=1.203170in, left, base]{\color{textcolor}\rmfamily\fontsize{10.000000}{12.000000}\selectfont Experience}%
\end{pgfscope}%
\begin{pgfscope}%
\pgfsetbuttcap%
\pgfsetroundjoin%
\definecolor{currentfill}{rgb}{0.000000,0.000000,0.000000}%
\pgfsetfillcolor{currentfill}%
\pgfsetlinewidth{0.803000pt}%
\definecolor{currentstroke}{rgb}{0.000000,0.000000,0.000000}%
\pgfsetstrokecolor{currentstroke}%
\pgfsetdash{}{0pt}%
\pgfsys@defobject{currentmarker}{\pgfqpoint{-0.048611in}{0.000000in}}{\pgfqpoint{-0.000000in}{0.000000in}}{%
\pgfpathmoveto{\pgfqpoint{-0.000000in}{0.000000in}}%
\pgfpathlineto{\pgfqpoint{-0.048611in}{0.000000in}}%
\pgfusepath{stroke,fill}%
}%
\begin{pgfscope}%
\pgfsys@transformshift{5.271468in}{1.108810in}%
\pgfsys@useobject{currentmarker}{}%
\end{pgfscope}%
\end{pgfscope}%
\begin{pgfscope}%
\definecolor{textcolor}{rgb}{0.000000,0.000000,0.000000}%
\pgfsetstrokecolor{textcolor}%
\pgfsetfillcolor{textcolor}%
\pgftext[x=4.688133in, y=1.060585in, left, base]{\color{textcolor}\rmfamily\fontsize{10.000000}{12.000000}\selectfont Opinion}%
\end{pgfscope}%
\begin{pgfscope}%
\pgfsetbuttcap%
\pgfsetroundjoin%
\definecolor{currentfill}{rgb}{0.000000,0.000000,0.000000}%
\pgfsetfillcolor{currentfill}%
\pgfsetlinewidth{0.803000pt}%
\definecolor{currentstroke}{rgb}{0.000000,0.000000,0.000000}%
\pgfsetstrokecolor{currentstroke}%
\pgfsetdash{}{0pt}%
\pgfsys@defobject{currentmarker}{\pgfqpoint{-0.048611in}{0.000000in}}{\pgfqpoint{-0.000000in}{0.000000in}}{%
\pgfpathmoveto{\pgfqpoint{-0.000000in}{0.000000in}}%
\pgfpathlineto{\pgfqpoint{-0.048611in}{0.000000in}}%
\pgfusepath{stroke,fill}%
}%
\begin{pgfscope}%
\pgfsys@transformshift{5.271468in}{0.966224in}%
\pgfsys@useobject{currentmarker}{}%
\end{pgfscope}%
\end{pgfscope}%
\begin{pgfscope}%
\definecolor{textcolor}{rgb}{0.000000,0.000000,0.000000}%
\pgfsetstrokecolor{textcolor}%
\pgfsetfillcolor{textcolor}%
\pgftext[x=4.725557in, y=0.917999in, left, base]{\color{textcolor}\rmfamily\fontsize{10.000000}{12.000000}\selectfont Answer}%
\end{pgfscope}%
\begin{pgfscope}%
\pgfsetbuttcap%
\pgfsetroundjoin%
\definecolor{currentfill}{rgb}{0.000000,0.000000,0.000000}%
\pgfsetfillcolor{currentfill}%
\pgfsetlinewidth{0.803000pt}%
\definecolor{currentstroke}{rgb}{0.000000,0.000000,0.000000}%
\pgfsetstrokecolor{currentstroke}%
\pgfsetdash{}{0pt}%
\pgfsys@defobject{currentmarker}{\pgfqpoint{-0.048611in}{0.000000in}}{\pgfqpoint{-0.000000in}{0.000000in}}{%
\pgfpathmoveto{\pgfqpoint{-0.000000in}{0.000000in}}%
\pgfpathlineto{\pgfqpoint{-0.048611in}{0.000000in}}%
\pgfusepath{stroke,fill}%
}%
\begin{pgfscope}%
\pgfsys@transformshift{5.271468in}{0.823639in}%
\pgfsys@useobject{currentmarker}{}%
\end{pgfscope}%
\end{pgfscope}%
\begin{pgfscope}%
\definecolor{textcolor}{rgb}{0.000000,0.000000,0.000000}%
\pgfsetstrokecolor{textcolor}%
\pgfsetfillcolor{textcolor}%
\pgftext[x=4.633349in, y=0.775414in, left, base]{\color{textcolor}\rmfamily\fontsize{10.000000}{12.000000}\selectfont Question}%
\end{pgfscope}%
\begin{pgfscope}%
\pgfsetbuttcap%
\pgfsetroundjoin%
\definecolor{currentfill}{rgb}{0.000000,0.000000,0.000000}%
\pgfsetfillcolor{currentfill}%
\pgfsetlinewidth{0.803000pt}%
\definecolor{currentstroke}{rgb}{0.000000,0.000000,0.000000}%
\pgfsetstrokecolor{currentstroke}%
\pgfsetdash{}{0pt}%
\pgfsys@defobject{currentmarker}{\pgfqpoint{-0.048611in}{0.000000in}}{\pgfqpoint{-0.000000in}{0.000000in}}{%
\pgfpathmoveto{\pgfqpoint{-0.000000in}{0.000000in}}%
\pgfpathlineto{\pgfqpoint{-0.048611in}{0.000000in}}%
\pgfusepath{stroke,fill}%
}%
\begin{pgfscope}%
\pgfsys@transformshift{5.271468in}{0.681053in}%
\pgfsys@useobject{currentmarker}{}%
\end{pgfscope}%
\end{pgfscope}%
\begin{pgfscope}%
\definecolor{textcolor}{rgb}{0.000000,0.000000,0.000000}%
\pgfsetstrokecolor{textcolor}%
\pgfsetfillcolor{textcolor}%
\pgftext[x=4.764137in, y=0.632828in, left, base]{\color{textcolor}\rmfamily\fontsize{10.000000}{12.000000}\selectfont Others}%
\end{pgfscope}%
\begin{pgfscope}%
\pgfsetbuttcap%
\pgfsetroundjoin%
\definecolor{currentfill}{rgb}{0.000000,0.000000,0.000000}%
\pgfsetfillcolor{currentfill}%
\pgfsetlinewidth{0.803000pt}%
\definecolor{currentstroke}{rgb}{0.000000,0.000000,0.000000}%
\pgfsetstrokecolor{currentstroke}%
\pgfsetdash{}{0pt}%
\pgfsys@defobject{currentmarker}{\pgfqpoint{-0.048611in}{0.000000in}}{\pgfqpoint{-0.000000in}{0.000000in}}{%
\pgfpathmoveto{\pgfqpoint{-0.000000in}{0.000000in}}%
\pgfpathlineto{\pgfqpoint{-0.048611in}{0.000000in}}%
\pgfusepath{stroke,fill}%
}%
\begin{pgfscope}%
\pgfsys@transformshift{5.271468in}{0.538468in}%
\pgfsys@useobject{currentmarker}{}%
\end{pgfscope}%
\end{pgfscope}%
\begin{pgfscope}%
\definecolor{textcolor}{rgb}{0.000000,0.000000,0.000000}%
\pgfsetstrokecolor{textcolor}%
\pgfsetfillcolor{textcolor}%
\pgftext[x=4.516451in, y=0.490243in, left, base]{\color{textcolor}\rmfamily\fontsize{10.000000}{12.000000}\selectfont Fact. Stat.}%
\end{pgfscope}%
\begin{pgfscope}%
\pgfpathrectangle{\pgfqpoint{5.271468in}{0.467175in}}{\pgfqpoint{0.646848in}{0.855513in}}%
\pgfusepath{clip}%
\pgfsetrectcap%
\pgfsetroundjoin%
\pgfsetlinewidth{2.710125pt}%
\definecolor{currentstroke}{rgb}{0.260000,0.260000,0.260000}%
\pgfsetstrokecolor{currentstroke}%
\pgfsetdash{}{0pt}%
\pgfusepath{stroke}%
\end{pgfscope}%
\begin{pgfscope}%
\pgfpathrectangle{\pgfqpoint{5.271468in}{0.467175in}}{\pgfqpoint{0.646848in}{0.855513in}}%
\pgfusepath{clip}%
\pgfsetrectcap%
\pgfsetroundjoin%
\pgfsetlinewidth{2.710125pt}%
\definecolor{currentstroke}{rgb}{0.260000,0.260000,0.260000}%
\pgfsetstrokecolor{currentstroke}%
\pgfsetdash{}{0pt}%
\pgfusepath{stroke}%
\end{pgfscope}%
\begin{pgfscope}%
\pgfpathrectangle{\pgfqpoint{5.271468in}{0.467175in}}{\pgfqpoint{0.646848in}{0.855513in}}%
\pgfusepath{clip}%
\pgfsetrectcap%
\pgfsetroundjoin%
\pgfsetlinewidth{2.710125pt}%
\definecolor{currentstroke}{rgb}{0.260000,0.260000,0.260000}%
\pgfsetstrokecolor{currentstroke}%
\pgfsetdash{}{0pt}%
\pgfusepath{stroke}%
\end{pgfscope}%
\begin{pgfscope}%
\pgfpathrectangle{\pgfqpoint{5.271468in}{0.467175in}}{\pgfqpoint{0.646848in}{0.855513in}}%
\pgfusepath{clip}%
\pgfsetrectcap%
\pgfsetroundjoin%
\pgfsetlinewidth{2.710125pt}%
\definecolor{currentstroke}{rgb}{0.260000,0.260000,0.260000}%
\pgfsetstrokecolor{currentstroke}%
\pgfsetdash{}{0pt}%
\pgfusepath{stroke}%
\end{pgfscope}%
\begin{pgfscope}%
\pgfpathrectangle{\pgfqpoint{5.271468in}{0.467175in}}{\pgfqpoint{0.646848in}{0.855513in}}%
\pgfusepath{clip}%
\pgfsetrectcap%
\pgfsetroundjoin%
\pgfsetlinewidth{2.710125pt}%
\definecolor{currentstroke}{rgb}{0.260000,0.260000,0.260000}%
\pgfsetstrokecolor{currentstroke}%
\pgfsetdash{}{0pt}%
\pgfusepath{stroke}%
\end{pgfscope}%
\begin{pgfscope}%
\pgfpathrectangle{\pgfqpoint{5.271468in}{0.467175in}}{\pgfqpoint{0.646848in}{0.855513in}}%
\pgfusepath{clip}%
\pgfsetrectcap%
\pgfsetroundjoin%
\pgfsetlinewidth{2.710125pt}%
\definecolor{currentstroke}{rgb}{0.260000,0.260000,0.260000}%
\pgfsetstrokecolor{currentstroke}%
\pgfsetdash{}{0pt}%
\pgfusepath{stroke}%
\end{pgfscope}%
\begin{pgfscope}%
\pgfsetrectcap%
\pgfsetmiterjoin%
\pgfsetlinewidth{0.803000pt}%
\definecolor{currentstroke}{rgb}{0.000000,0.000000,0.000000}%
\pgfsetstrokecolor{currentstroke}%
\pgfsetdash{}{0pt}%
\pgfpathmoveto{\pgfqpoint{5.271468in}{0.467175in}}%
\pgfpathlineto{\pgfqpoint{5.271468in}{1.322688in}}%
\pgfusepath{stroke}%
\end{pgfscope}%
\begin{pgfscope}%
\definecolor{textcolor}{rgb}{0.000000,0.000000,0.000000}%
\pgfsetstrokecolor{textcolor}%
\pgfsetfillcolor{textcolor}%
\pgftext[x=5.850969in,y=1.208620in,left,base]{\color{textcolor}\rmfamily\fontsize{8.000000}{9.600000}\selectfont 53\%}%
\end{pgfscope}%
\begin{pgfscope}%
\definecolor{textcolor}{rgb}{0.000000,0.000000,0.000000}%
\pgfsetstrokecolor{textcolor}%
\pgfsetfillcolor{textcolor}%
\pgftext[x=5.647008in,y=1.066034in,left,base]{\color{textcolor}\rmfamily\fontsize{8.000000}{9.600000}\selectfont 34\%}%
\end{pgfscope}%
\begin{pgfscope}%
\definecolor{textcolor}{rgb}{0.000000,0.000000,0.000000}%
\pgfsetstrokecolor{textcolor}%
\pgfsetfillcolor{textcolor}%
\pgftext[x=5.452760in,y=0.923449in,left,base]{\color{textcolor}\rmfamily\fontsize{8.000000}{9.600000}\selectfont 16\%}%
\end{pgfscope}%
\begin{pgfscope}%
\definecolor{textcolor}{rgb}{0.000000,0.000000,0.000000}%
\pgfsetstrokecolor{textcolor}%
\pgfsetfillcolor{textcolor}%
\pgftext[x=5.375060in,y=0.780863in,left,base]{\color{textcolor}\rmfamily\fontsize{8.000000}{9.600000}\selectfont 9\%}%
\end{pgfscope}%
\begin{pgfscope}%
\definecolor{textcolor}{rgb}{0.000000,0.000000,0.000000}%
\pgfsetstrokecolor{textcolor}%
\pgfsetfillcolor{textcolor}%
\pgftext[x=5.355635in,y=0.638278in,left,base]{\color{textcolor}\rmfamily\fontsize{8.000000}{9.600000}\selectfont 7\%}%
\end{pgfscope}%
\begin{pgfscope}%
\definecolor{textcolor}{rgb}{0.000000,0.000000,0.000000}%
\pgfsetstrokecolor{textcolor}%
\pgfsetfillcolor{textcolor}%
\pgftext[x=5.549884in,y=0.495692in,left,base]{\color{textcolor}\rmfamily\fontsize{8.000000}{9.600000}\selectfont 25\%}%
\end{pgfscope}%
\begin{pgfscope}%
\definecolor{textcolor}{rgb}{0.000000,0.000000,0.000000}%
\pgfsetstrokecolor{textcolor}%
\pgfsetfillcolor{textcolor}%
\pgftext[x=5.594891in,y=0.079977in,,base]{\color{textcolor}\rmfamily\fontsize{11.000000}{13.200000}\selectfont Presentation}%
\end{pgfscope}%
\end{pgfpicture}%
\makeatother%
\endgroup%

%% file: figures/PCMI_acknowledgement_ratio.pgf
\begingroup%
\makeatletter%
\begin{pgfpicture}%
\pgfpathrectangle{\pgfpointorigin}{\pgfqpoint{3.031495in}{1.049197in}}%
\pgfusepath{use as bounding box, clip}%
\begin{pgfscope}%
\pgfsetbuttcap%
\pgfsetmiterjoin%
\definecolor{currentfill}{rgb}{1.000000,1.000000,1.000000}%
\pgfsetfillcolor{currentfill}%
\pgfsetlinewidth{0.000000pt}%
\definecolor{currentstroke}{rgb}{1.000000,1.000000,1.000000}%
\pgfsetstrokecolor{currentstroke}%
\pgfsetstrokeopacity{0.000000}%
\pgfsetdash{}{0pt}%
\pgfpathmoveto{\pgfqpoint{0.000000in}{0.000000in}}%
\pgfpathlineto{\pgfqpoint{3.031495in}{0.000000in}}%
\pgfpathlineto{\pgfqpoint{3.031495in}{1.049197in}}%
\pgfpathlineto{\pgfqpoint{0.000000in}{1.049197in}}%
\pgfpathclose%
\pgfusepath{fill}%
\end{pgfscope}%
\begin{pgfscope}%
\pgfsetbuttcap%
\pgfsetmiterjoin%
\definecolor{currentfill}{rgb}{1.000000,1.000000,1.000000}%
\pgfsetfillcolor{currentfill}%
\pgfsetlinewidth{0.000000pt}%
\definecolor{currentstroke}{rgb}{0.000000,0.000000,0.000000}%
\pgfsetstrokecolor{currentstroke}%
\pgfsetstrokeopacity{0.000000}%
\pgfsetdash{}{0pt}%
\pgfpathmoveto{\pgfqpoint{0.773525in}{0.419679in}}%
\pgfpathlineto{\pgfqpoint{2.887495in}{0.419679in}}%
\pgfpathlineto{\pgfqpoint{2.887495in}{0.905197in}}%
\pgfpathlineto{\pgfqpoint{0.773525in}{0.905197in}}%
\pgfpathclose%
\pgfusepath{fill}%
\end{pgfscope}%
\begin{pgfscope}%
\pgfpathrectangle{\pgfqpoint{0.773525in}{0.419679in}}{\pgfqpoint{2.113969in}{0.485518in}}%
\pgfusepath{clip}%
\pgfsetroundcap%
\pgfsetroundjoin%
\pgfsetlinewidth{0.803000pt}%
\definecolor{currentstroke}{rgb}{0.800000,0.800000,0.800000}%
\pgfsetstrokecolor{currentstroke}%
\pgfsetdash{}{0pt}%
\pgfpathmoveto{\pgfqpoint{1.285568in}{0.419679in}}%
\pgfpathlineto{\pgfqpoint{1.285568in}{0.905197in}}%
\pgfusepath{stroke}%
\end{pgfscope}%
\begin{pgfscope}%
\definecolor{textcolor}{rgb}{0.150000,0.150000,0.150000}%
\pgfsetstrokecolor{textcolor}%
\pgfsetfillcolor{textcolor}%
\pgftext[x=1.285568in,y=0.304401in,,top]{\color{textcolor}\rmfamily\fontsize{8.800000}{10.560000}\selectfont 0\%}%
\end{pgfscope}%
\begin{pgfscope}%
\pgfpathrectangle{\pgfqpoint{0.773525in}{0.419679in}}{\pgfqpoint{2.113969in}{0.485518in}}%
\pgfusepath{clip}%
\pgfsetroundcap%
\pgfsetroundjoin%
\pgfsetlinewidth{0.803000pt}%
\definecolor{currentstroke}{rgb}{0.800000,0.800000,0.800000}%
\pgfsetstrokecolor{currentstroke}%
\pgfsetdash{}{0pt}%
\pgfpathmoveto{\pgfqpoint{1.858934in}{0.419679in}}%
\pgfpathlineto{\pgfqpoint{1.858934in}{0.905197in}}%
\pgfusepath{stroke}%
\end{pgfscope}%
\begin{pgfscope}%
\definecolor{textcolor}{rgb}{0.150000,0.150000,0.150000}%
\pgfsetstrokecolor{textcolor}%
\pgfsetfillcolor{textcolor}%
\pgftext[x=1.858934in,y=0.304401in,,top]{\color{textcolor}\rmfamily\fontsize{8.800000}{10.560000}\selectfont 50\%}%
\end{pgfscope}%
\begin{pgfscope}%
\pgfpathrectangle{\pgfqpoint{0.773525in}{0.419679in}}{\pgfqpoint{2.113969in}{0.485518in}}%
\pgfusepath{clip}%
\pgfsetroundcap%
\pgfsetroundjoin%
\pgfsetlinewidth{0.803000pt}%
\definecolor{currentstroke}{rgb}{0.800000,0.800000,0.800000}%
\pgfsetstrokecolor{currentstroke}%
\pgfsetdash{}{0pt}%
\pgfpathmoveto{\pgfqpoint{2.432300in}{0.419679in}}%
\pgfpathlineto{\pgfqpoint{2.432300in}{0.905197in}}%
\pgfusepath{stroke}%
\end{pgfscope}%
\begin{pgfscope}%
\definecolor{textcolor}{rgb}{0.150000,0.150000,0.150000}%
\pgfsetstrokecolor{textcolor}%
\pgfsetfillcolor{textcolor}%
\pgftext[x=2.432300in,y=0.304401in,,top]{\color{textcolor}\rmfamily\fontsize{8.800000}{10.560000}\selectfont 100\%}%
\end{pgfscope}%
\begin{pgfscope}%
\definecolor{textcolor}{rgb}{0.150000,0.150000,0.150000}%
\pgfsetstrokecolor{textcolor}%
\pgfsetfillcolor{textcolor}%
\pgftext[x=1.830510in,y=0.137735in,,top]{\color{textcolor}\rmfamily\fontsize{9.600000}{11.520000}\selectfont \% Contribution of acknowledgement span}%
\end{pgfscope}%
\begin{pgfscope}%
\definecolor{textcolor}{rgb}{0.150000,0.150000,0.150000}%
\pgfsetstrokecolor{textcolor}%
\pgfsetfillcolor{textcolor}%
\pgftext[x=0.319708in, y=0.782459in, left, base]{\color{textcolor}\rmfamily\fontsize{8.800000}{10.560000}\selectfont \(\displaystyle \mathrm{pcmi}_k\)}%
\end{pgfscope}%
\begin{pgfscope}%
\definecolor{textcolor}{rgb}{0.150000,0.150000,0.150000}%
\pgfsetstrokecolor{textcolor}%
\pgfsetfillcolor{textcolor}%
\pgftext[x=0.316043in, y=0.620620in, left, base]{\color{textcolor}\rmfamily\fontsize{8.800000}{10.560000}\selectfont \(\displaystyle \mathrm{pcmi}_h\)}%
\end{pgfscope}%
\begin{pgfscope}%
\definecolor{textcolor}{rgb}{0.150000,0.150000,0.150000}%
\pgfsetstrokecolor{textcolor}%
\pgfsetfillcolor{textcolor}%
\pgftext[x=0.147338in, y=0.453724in, left, base]{\color{textcolor}\rmfamily\fontsize{8.800000}{10.560000}\selectfont \(\displaystyle \mathrm{pmi}(g;h)\)}%
\end{pgfscope}%
\begin{pgfscope}%
\pgfpathrectangle{\pgfqpoint{0.773525in}{0.419679in}}{\pgfqpoint{2.113969in}{0.485518in}}%
\pgfusepath{clip}%
\pgfsetbuttcap%
\pgfsetmiterjoin%
\definecolor{currentfill}{rgb}{0.672059,0.789706,0.916176}%
\pgfsetfillcolor{currentfill}%
\pgfsetlinewidth{1.204500pt}%
\definecolor{currentstroke}{rgb}{0.435294,0.435294,0.435294}%
\pgfsetstrokecolor{currentstroke}%
\pgfsetdash{}{0pt}%
\pgfpathmoveto{\pgfqpoint{1.299258in}{0.889013in}}%
\pgfpathlineto{\pgfqpoint{1.299258in}{0.759542in}}%
\pgfpathlineto{\pgfqpoint{1.468286in}{0.759542in}}%
\pgfpathlineto{\pgfqpoint{1.468286in}{0.889013in}}%
\pgfpathlineto{\pgfqpoint{1.299258in}{0.889013in}}%
\pgfpathclose%
\pgfusepath{stroke,fill}%
\end{pgfscope}%
\begin{pgfscope}%
\pgfpathrectangle{\pgfqpoint{0.773525in}{0.419679in}}{\pgfqpoint{2.113969in}{0.485518in}}%
\pgfusepath{clip}%
\pgfsetbuttcap%
\pgfsetmiterjoin%
\definecolor{currentfill}{rgb}{0.938725,0.718137,0.571078}%
\pgfsetfillcolor{currentfill}%
\pgfsetlinewidth{1.204500pt}%
\definecolor{currentstroke}{rgb}{0.435294,0.435294,0.435294}%
\pgfsetstrokecolor{currentstroke}%
\pgfsetdash{}{0pt}%
\pgfpathmoveto{\pgfqpoint{1.512048in}{0.727174in}}%
\pgfpathlineto{\pgfqpoint{1.512048in}{0.597702in}}%
\pgfpathlineto{\pgfqpoint{2.151426in}{0.597702in}}%
\pgfpathlineto{\pgfqpoint{2.151426in}{0.727174in}}%
\pgfpathlineto{\pgfqpoint{1.512048in}{0.727174in}}%
\pgfpathclose%
\pgfusepath{stroke,fill}%
\end{pgfscope}%
\begin{pgfscope}%
\pgfpathrectangle{\pgfqpoint{0.773525in}{0.419679in}}{\pgfqpoint{2.113969in}{0.485518in}}%
\pgfusepath{clip}%
\pgfsetbuttcap%
\pgfsetmiterjoin%
\definecolor{currentfill}{rgb}{0.596078,0.854902,0.654902}%
\pgfsetfillcolor{currentfill}%
\pgfsetlinewidth{1.204500pt}%
\definecolor{currentstroke}{rgb}{0.435294,0.435294,0.435294}%
\pgfsetstrokecolor{currentstroke}%
\pgfsetdash{}{0pt}%
\pgfpathmoveto{\pgfqpoint{1.397947in}{0.565334in}}%
\pgfpathlineto{\pgfqpoint{1.397947in}{0.435863in}}%
\pgfpathlineto{\pgfqpoint{2.191017in}{0.435863in}}%
\pgfpathlineto{\pgfqpoint{2.191017in}{0.565334in}}%
\pgfpathlineto{\pgfqpoint{1.397947in}{0.565334in}}%
\pgfpathclose%
\pgfusepath{stroke,fill}%
\end{pgfscope}%
\begin{pgfscope}%
\pgfpathrectangle{\pgfqpoint{0.773525in}{0.419679in}}{\pgfqpoint{2.113969in}{0.485518in}}%
\pgfusepath{clip}%
\pgfsetroundcap%
\pgfsetroundjoin%
\pgfsetlinewidth{1.204500pt}%
\definecolor{currentstroke}{rgb}{0.435294,0.435294,0.435294}%
\pgfsetstrokecolor{currentstroke}%
\pgfsetdash{}{0pt}%
\pgfpathmoveto{\pgfqpoint{1.299258in}{0.824278in}}%
\pgfpathlineto{\pgfqpoint{1.267205in}{0.824278in}}%
\pgfusepath{stroke}%
\end{pgfscope}%
\begin{pgfscope}%
\pgfpathrectangle{\pgfqpoint{0.773525in}{0.419679in}}{\pgfqpoint{2.113969in}{0.485518in}}%
\pgfusepath{clip}%
\pgfsetroundcap%
\pgfsetroundjoin%
\pgfsetlinewidth{1.204500pt}%
\definecolor{currentstroke}{rgb}{0.435294,0.435294,0.435294}%
\pgfsetstrokecolor{currentstroke}%
\pgfsetdash{}{0pt}%
\pgfpathmoveto{\pgfqpoint{1.468286in}{0.824278in}}%
\pgfpathlineto{\pgfqpoint{1.629685in}{0.824278in}}%
\pgfusepath{stroke}%
\end{pgfscope}%
\begin{pgfscope}%
\pgfpathrectangle{\pgfqpoint{0.773525in}{0.419679in}}{\pgfqpoint{2.113969in}{0.485518in}}%
\pgfusepath{clip}%
\pgfsetroundcap%
\pgfsetroundjoin%
\pgfsetlinewidth{1.204500pt}%
\definecolor{currentstroke}{rgb}{0.435294,0.435294,0.435294}%
\pgfsetstrokecolor{currentstroke}%
\pgfsetdash{}{0pt}%
\pgfpathmoveto{\pgfqpoint{1.267205in}{0.856646in}}%
\pgfpathlineto{\pgfqpoint{1.267205in}{0.791910in}}%
\pgfusepath{stroke}%
\end{pgfscope}%
\begin{pgfscope}%
\pgfpathrectangle{\pgfqpoint{0.773525in}{0.419679in}}{\pgfqpoint{2.113969in}{0.485518in}}%
\pgfusepath{clip}%
\pgfsetroundcap%
\pgfsetroundjoin%
\pgfsetlinewidth{1.204500pt}%
\definecolor{currentstroke}{rgb}{0.435294,0.435294,0.435294}%
\pgfsetstrokecolor{currentstroke}%
\pgfsetdash{}{0pt}%
\pgfpathmoveto{\pgfqpoint{1.629685in}{0.856646in}}%
\pgfpathlineto{\pgfqpoint{1.629685in}{0.791910in}}%
\pgfusepath{stroke}%
\end{pgfscope}%
\begin{pgfscope}%
\pgfpathrectangle{\pgfqpoint{0.773525in}{0.419679in}}{\pgfqpoint{2.113969in}{0.485518in}}%
\pgfusepath{clip}%
\pgfsetroundcap%
\pgfsetroundjoin%
\pgfsetlinewidth{1.204500pt}%
\definecolor{currentstroke}{rgb}{0.435294,0.435294,0.435294}%
\pgfsetstrokecolor{currentstroke}%
\pgfsetdash{}{0pt}%
\pgfpathmoveto{\pgfqpoint{1.512048in}{0.662438in}}%
\pgfpathlineto{\pgfqpoint{1.046127in}{0.662438in}}%
\pgfusepath{stroke}%
\end{pgfscope}%
\begin{pgfscope}%
\pgfpathrectangle{\pgfqpoint{0.773525in}{0.419679in}}{\pgfqpoint{2.113969in}{0.485518in}}%
\pgfusepath{clip}%
\pgfsetroundcap%
\pgfsetroundjoin%
\pgfsetlinewidth{1.204500pt}%
\definecolor{currentstroke}{rgb}{0.435294,0.435294,0.435294}%
\pgfsetstrokecolor{currentstroke}%
\pgfsetdash{}{0pt}%
\pgfpathmoveto{\pgfqpoint{2.151426in}{0.662438in}}%
\pgfpathlineto{\pgfqpoint{2.733030in}{0.662438in}}%
\pgfusepath{stroke}%
\end{pgfscope}%
\begin{pgfscope}%
\pgfpathrectangle{\pgfqpoint{0.773525in}{0.419679in}}{\pgfqpoint{2.113969in}{0.485518in}}%
\pgfusepath{clip}%
\pgfsetroundcap%
\pgfsetroundjoin%
\pgfsetlinewidth{1.204500pt}%
\definecolor{currentstroke}{rgb}{0.435294,0.435294,0.435294}%
\pgfsetstrokecolor{currentstroke}%
\pgfsetdash{}{0pt}%
\pgfpathmoveto{\pgfqpoint{1.046127in}{0.694806in}}%
\pgfpathlineto{\pgfqpoint{1.046127in}{0.630070in}}%
\pgfusepath{stroke}%
\end{pgfscope}%
\begin{pgfscope}%
\pgfpathrectangle{\pgfqpoint{0.773525in}{0.419679in}}{\pgfqpoint{2.113969in}{0.485518in}}%
\pgfusepath{clip}%
\pgfsetroundcap%
\pgfsetroundjoin%
\pgfsetlinewidth{1.204500pt}%
\definecolor{currentstroke}{rgb}{0.435294,0.435294,0.435294}%
\pgfsetstrokecolor{currentstroke}%
\pgfsetdash{}{0pt}%
\pgfpathmoveto{\pgfqpoint{2.733030in}{0.694806in}}%
\pgfpathlineto{\pgfqpoint{2.733030in}{0.630070in}}%
\pgfusepath{stroke}%
\end{pgfscope}%
\begin{pgfscope}%
\pgfpathrectangle{\pgfqpoint{0.773525in}{0.419679in}}{\pgfqpoint{2.113969in}{0.485518in}}%
\pgfusepath{clip}%
\pgfsetroundcap%
\pgfsetroundjoin%
\pgfsetlinewidth{1.204500pt}%
\definecolor{currentstroke}{rgb}{0.435294,0.435294,0.435294}%
\pgfsetstrokecolor{currentstroke}%
\pgfsetdash{}{0pt}%
\pgfpathmoveto{\pgfqpoint{1.397947in}{0.500599in}}%
\pgfpathlineto{\pgfqpoint{0.869615in}{0.500599in}}%
\pgfusepath{stroke}%
\end{pgfscope}%
\begin{pgfscope}%
\pgfpathrectangle{\pgfqpoint{0.773525in}{0.419679in}}{\pgfqpoint{2.113969in}{0.485518in}}%
\pgfusepath{clip}%
\pgfsetroundcap%
\pgfsetroundjoin%
\pgfsetlinewidth{1.204500pt}%
\definecolor{currentstroke}{rgb}{0.435294,0.435294,0.435294}%
\pgfsetstrokecolor{currentstroke}%
\pgfsetdash{}{0pt}%
\pgfpathmoveto{\pgfqpoint{2.191017in}{0.500599in}}%
\pgfpathlineto{\pgfqpoint{2.791405in}{0.500599in}}%
\pgfusepath{stroke}%
\end{pgfscope}%
\begin{pgfscope}%
\pgfpathrectangle{\pgfqpoint{0.773525in}{0.419679in}}{\pgfqpoint{2.113969in}{0.485518in}}%
\pgfusepath{clip}%
\pgfsetroundcap%
\pgfsetroundjoin%
\pgfsetlinewidth{1.204500pt}%
\definecolor{currentstroke}{rgb}{0.435294,0.435294,0.435294}%
\pgfsetstrokecolor{currentstroke}%
\pgfsetdash{}{0pt}%
\pgfpathmoveto{\pgfqpoint{0.869615in}{0.532967in}}%
\pgfpathlineto{\pgfqpoint{0.869615in}{0.468231in}}%
\pgfusepath{stroke}%
\end{pgfscope}%
\begin{pgfscope}%
\pgfpathrectangle{\pgfqpoint{0.773525in}{0.419679in}}{\pgfqpoint{2.113969in}{0.485518in}}%
\pgfusepath{clip}%
\pgfsetroundcap%
\pgfsetroundjoin%
\pgfsetlinewidth{1.204500pt}%
\definecolor{currentstroke}{rgb}{0.435294,0.435294,0.435294}%
\pgfsetstrokecolor{currentstroke}%
\pgfsetdash{}{0pt}%
\pgfpathmoveto{\pgfqpoint{2.791405in}{0.532967in}}%
\pgfpathlineto{\pgfqpoint{2.791405in}{0.468231in}}%
\pgfusepath{stroke}%
\end{pgfscope}%
\begin{pgfscope}%
\pgfpathrectangle{\pgfqpoint{0.773525in}{0.419679in}}{\pgfqpoint{2.113969in}{0.485518in}}%
\pgfusepath{clip}%
\pgfsetroundcap%
\pgfsetroundjoin%
\pgfsetlinewidth{1.204500pt}%
\definecolor{currentstroke}{rgb}{0.435294,0.435294,0.435294}%
\pgfsetstrokecolor{currentstroke}%
\pgfsetdash{}{0pt}%
\pgfpathmoveto{\pgfqpoint{1.338723in}{0.889013in}}%
\pgfpathlineto{\pgfqpoint{1.338723in}{0.759542in}}%
\pgfusepath{stroke}%
\end{pgfscope}%
\begin{pgfscope}%
\pgfpathrectangle{\pgfqpoint{0.773525in}{0.419679in}}{\pgfqpoint{2.113969in}{0.485518in}}%
\pgfusepath{clip}%
\pgfsetroundcap%
\pgfsetroundjoin%
\pgfsetlinewidth{1.204500pt}%
\definecolor{currentstroke}{rgb}{0.435294,0.435294,0.435294}%
\pgfsetstrokecolor{currentstroke}%
\pgfsetdash{}{0pt}%
\pgfpathmoveto{\pgfqpoint{1.825283in}{0.727174in}}%
\pgfpathlineto{\pgfqpoint{1.825283in}{0.597702in}}%
\pgfusepath{stroke}%
\end{pgfscope}%
\begin{pgfscope}%
\pgfpathrectangle{\pgfqpoint{0.773525in}{0.419679in}}{\pgfqpoint{2.113969in}{0.485518in}}%
\pgfusepath{clip}%
\pgfsetroundcap%
\pgfsetroundjoin%
\pgfsetlinewidth{1.204500pt}%
\definecolor{currentstroke}{rgb}{0.435294,0.435294,0.435294}%
\pgfsetstrokecolor{currentstroke}%
\pgfsetdash{}{0pt}%
\pgfpathmoveto{\pgfqpoint{1.692062in}{0.565334in}}%
\pgfpathlineto{\pgfqpoint{1.692062in}{0.435863in}}%
\pgfusepath{stroke}%
\end{pgfscope}%
\begin{pgfscope}%
\pgfsetrectcap%
\pgfsetmiterjoin%
\pgfsetlinewidth{1.003750pt}%
\definecolor{currentstroke}{rgb}{0.800000,0.800000,0.800000}%
\pgfsetstrokecolor{currentstroke}%
\pgfsetdash{}{0pt}%
\pgfpathmoveto{\pgfqpoint{0.773525in}{0.419679in}}%
\pgfpathlineto{\pgfqpoint{2.887495in}{0.419679in}}%
\pgfusepath{stroke}%
\end{pgfscope}%
\end{pgfpicture}%
\makeatother%
\endgroup%

%% file: figures/PCMI_skew_only_univariate.pgf
\begingroup%
\makeatletter%
\begin{pgfpicture}%
\pgfpathrectangle{\pgfpointorigin}{\pgfqpoint{3.031495in}{1.873567in}}%
\pgfusepath{use as bounding box, clip}%
\begin{pgfscope}%
\pgfsetbuttcap%
\pgfsetmiterjoin%
\definecolor{currentfill}{rgb}{1.000000,1.000000,1.000000}%
\pgfsetfillcolor{currentfill}%
\pgfsetlinewidth{0.000000pt}%
\definecolor{currentstroke}{rgb}{1.000000,1.000000,1.000000}%
\pgfsetstrokecolor{currentstroke}%
\pgfsetstrokeopacity{0.000000}%
\pgfsetdash{}{0pt}%
\pgfpathmoveto{\pgfqpoint{0.000000in}{0.000000in}}%
\pgfpathlineto{\pgfqpoint{3.031495in}{0.000000in}}%
\pgfpathlineto{\pgfqpoint{3.031495in}{1.873567in}}%
\pgfpathlineto{\pgfqpoint{0.000000in}{1.873567in}}%
\pgfpathclose%
\pgfusepath{fill}%
\end{pgfscope}%
\begin{pgfscope}%
\pgfsetbuttcap%
\pgfsetmiterjoin%
\definecolor{currentfill}{rgb}{1.000000,1.000000,1.000000}%
\pgfsetfillcolor{currentfill}%
\pgfsetlinewidth{0.000000pt}%
\definecolor{currentstroke}{rgb}{0.000000,0.000000,0.000000}%
\pgfsetstrokecolor{currentstroke}%
\pgfsetstrokeopacity{0.000000}%
\pgfsetdash{}{0pt}%
\pgfpathmoveto{\pgfqpoint{0.634134in}{0.187357in}}%
\pgfpathlineto{\pgfqpoint{1.476092in}{0.187357in}}%
\pgfpathlineto{\pgfqpoint{1.476092in}{1.546607in}}%
\pgfpathlineto{\pgfqpoint{0.634134in}{1.546607in}}%
\pgfpathclose%
\pgfusepath{fill}%
\end{pgfscope}%
\begin{pgfscope}%
\pgfpathrectangle{\pgfqpoint{0.634134in}{0.187357in}}{\pgfqpoint{0.841957in}{1.359250in}}%
\pgfusepath{clip}%
\pgfsetroundcap%
\pgfsetroundjoin%
\pgfsetlinewidth{0.803000pt}%
\definecolor{currentstroke}{rgb}{0.800000,0.800000,0.800000}%
\pgfsetstrokecolor{currentstroke}%
\pgfsetdash{}{0pt}%
\pgfpathmoveto{\pgfqpoint{0.634134in}{0.256936in}}%
\pgfpathlineto{\pgfqpoint{1.476092in}{0.256936in}}%
\pgfusepath{stroke}%
\end{pgfscope}%
\begin{pgfscope}%
\pgfsetbuttcap%
\pgfsetroundjoin%
\definecolor{currentfill}{rgb}{0.150000,0.150000,0.150000}%
\pgfsetfillcolor{currentfill}%
\pgfsetlinewidth{1.003750pt}%
\definecolor{currentstroke}{rgb}{0.150000,0.150000,0.150000}%
\pgfsetstrokecolor{currentstroke}%
\pgfsetdash{}{0pt}%
\pgfsys@defobject{currentmarker}{\pgfqpoint{-0.069444in}{0.000000in}}{\pgfqpoint{-0.000000in}{0.000000in}}{%
\pgfpathmoveto{\pgfqpoint{-0.000000in}{0.000000in}}%
\pgfpathlineto{\pgfqpoint{-0.069444in}{0.000000in}}%
\pgfusepath{stroke,fill}%
}%
\begin{pgfscope}%
\pgfsys@transformshift{0.634134in}{0.256936in}%
\pgfsys@useobject{currentmarker}{}%
\end{pgfscope}%
\end{pgfscope}%
\begin{pgfscope}%
\definecolor{textcolor}{rgb}{0.150000,0.150000,0.150000}%
\pgfsetstrokecolor{textcolor}%
\pgfsetfillcolor{textcolor}%
\pgftext[x=0.451843in, y=0.213533in, left, base]{\color{textcolor}\sffamily\fontsize{8.800000}{10.560000}\selectfont \(\displaystyle {0}\)}%
\end{pgfscope}%
\begin{pgfscope}%
\pgfpathrectangle{\pgfqpoint{0.634134in}{0.187357in}}{\pgfqpoint{0.841957in}{1.359250in}}%
\pgfusepath{clip}%
\pgfsetroundcap%
\pgfsetroundjoin%
\pgfsetlinewidth{0.803000pt}%
\definecolor{currentstroke}{rgb}{0.800000,0.800000,0.800000}%
\pgfsetstrokecolor{currentstroke}%
\pgfsetdash{}{0pt}%
\pgfpathmoveto{\pgfqpoint{0.634134in}{0.671757in}}%
\pgfpathlineto{\pgfqpoint{1.476092in}{0.671757in}}%
\pgfusepath{stroke}%
\end{pgfscope}%
\begin{pgfscope}%
\pgfsetbuttcap%
\pgfsetroundjoin%
\definecolor{currentfill}{rgb}{0.150000,0.150000,0.150000}%
\pgfsetfillcolor{currentfill}%
\pgfsetlinewidth{1.003750pt}%
\definecolor{currentstroke}{rgb}{0.150000,0.150000,0.150000}%
\pgfsetstrokecolor{currentstroke}%
\pgfsetdash{}{0pt}%
\pgfsys@defobject{currentmarker}{\pgfqpoint{-0.069444in}{0.000000in}}{\pgfqpoint{-0.000000in}{0.000000in}}{%
\pgfpathmoveto{\pgfqpoint{-0.000000in}{0.000000in}}%
\pgfpathlineto{\pgfqpoint{-0.069444in}{0.000000in}}%
\pgfusepath{stroke,fill}%
}%
\begin{pgfscope}%
\pgfsys@transformshift{0.634134in}{0.671757in}%
\pgfsys@useobject{currentmarker}{}%
\end{pgfscope}%
\end{pgfscope}%
\begin{pgfscope}%
\definecolor{textcolor}{rgb}{0.150000,0.150000,0.150000}%
\pgfsetstrokecolor{textcolor}%
\pgfsetfillcolor{textcolor}%
\pgftext[x=0.387607in, y=0.628354in, left, base]{\color{textcolor}\sffamily\fontsize{8.800000}{10.560000}\selectfont \(\displaystyle {50}\)}%
\end{pgfscope}%
\begin{pgfscope}%
\pgfpathrectangle{\pgfqpoint{0.634134in}{0.187357in}}{\pgfqpoint{0.841957in}{1.359250in}}%
\pgfusepath{clip}%
\pgfsetroundcap%
\pgfsetroundjoin%
\pgfsetlinewidth{0.803000pt}%
\definecolor{currentstroke}{rgb}{0.800000,0.800000,0.800000}%
\pgfsetstrokecolor{currentstroke}%
\pgfsetdash{}{0pt}%
\pgfpathmoveto{\pgfqpoint{0.634134in}{1.086578in}}%
\pgfpathlineto{\pgfqpoint{1.476092in}{1.086578in}}%
\pgfusepath{stroke}%
\end{pgfscope}%
\begin{pgfscope}%
\pgfsetbuttcap%
\pgfsetroundjoin%
\definecolor{currentfill}{rgb}{0.150000,0.150000,0.150000}%
\pgfsetfillcolor{currentfill}%
\pgfsetlinewidth{1.003750pt}%
\definecolor{currentstroke}{rgb}{0.150000,0.150000,0.150000}%
\pgfsetstrokecolor{currentstroke}%
\pgfsetdash{}{0pt}%
\pgfsys@defobject{currentmarker}{\pgfqpoint{-0.069444in}{0.000000in}}{\pgfqpoint{-0.000000in}{0.000000in}}{%
\pgfpathmoveto{\pgfqpoint{-0.000000in}{0.000000in}}%
\pgfpathlineto{\pgfqpoint{-0.069444in}{0.000000in}}%
\pgfusepath{stroke,fill}%
}%
\begin{pgfscope}%
\pgfsys@transformshift{0.634134in}{1.086578in}%
\pgfsys@useobject{currentmarker}{}%
\end{pgfscope}%
\end{pgfscope}%
\begin{pgfscope}%
\definecolor{textcolor}{rgb}{0.150000,0.150000,0.150000}%
\pgfsetstrokecolor{textcolor}%
\pgfsetfillcolor{textcolor}%
\pgftext[x=0.323372in, y=1.043175in, left, base]{\color{textcolor}\sffamily\fontsize{8.800000}{10.560000}\selectfont \(\displaystyle {100}\)}%
\end{pgfscope}%
\begin{pgfscope}%
\pgfpathrectangle{\pgfqpoint{0.634134in}{0.187357in}}{\pgfqpoint{0.841957in}{1.359250in}}%
\pgfusepath{clip}%
\pgfsetroundcap%
\pgfsetroundjoin%
\pgfsetlinewidth{0.803000pt}%
\definecolor{currentstroke}{rgb}{0.800000,0.800000,0.800000}%
\pgfsetstrokecolor{currentstroke}%
\pgfsetdash{}{0pt}%
\pgfpathmoveto{\pgfqpoint{0.634134in}{1.501399in}}%
\pgfpathlineto{\pgfqpoint{1.476092in}{1.501399in}}%
\pgfusepath{stroke}%
\end{pgfscope}%
\begin{pgfscope}%
\pgfsetbuttcap%
\pgfsetroundjoin%
\definecolor{currentfill}{rgb}{0.150000,0.150000,0.150000}%
\pgfsetfillcolor{currentfill}%
\pgfsetlinewidth{1.003750pt}%
\definecolor{currentstroke}{rgb}{0.150000,0.150000,0.150000}%
\pgfsetstrokecolor{currentstroke}%
\pgfsetdash{}{0pt}%
\pgfsys@defobject{currentmarker}{\pgfqpoint{-0.069444in}{0.000000in}}{\pgfqpoint{-0.000000in}{0.000000in}}{%
\pgfpathmoveto{\pgfqpoint{-0.000000in}{0.000000in}}%
\pgfpathlineto{\pgfqpoint{-0.069444in}{0.000000in}}%
\pgfusepath{stroke,fill}%
}%
\begin{pgfscope}%
\pgfsys@transformshift{0.634134in}{1.501399in}%
\pgfsys@useobject{currentmarker}{}%
\end{pgfscope}%
\end{pgfscope}%
\begin{pgfscope}%
\definecolor{textcolor}{rgb}{0.150000,0.150000,0.150000}%
\pgfsetstrokecolor{textcolor}%
\pgfsetfillcolor{textcolor}%
\pgftext[x=0.323372in, y=1.457997in, left, base]{\color{textcolor}\sffamily\fontsize{8.800000}{10.560000}\selectfont \(\displaystyle {150}\)}%
\end{pgfscope}%
\begin{pgfscope}%
\definecolor{textcolor}{rgb}{0.150000,0.150000,0.150000}%
\pgfsetstrokecolor{textcolor}%
\pgfsetfillcolor{textcolor}%
\pgftext[x=0.267816in,y=0.866982in,,bottom,rotate=90.000000]{\color{textcolor}\sffamily\fontsize{9.600000}{11.520000}\selectfont \(\displaystyle \mathrm{pcmi}_k\)}%
\end{pgfscope}%
\begin{pgfscope}%
\pgfpathrectangle{\pgfqpoint{0.634134in}{0.187357in}}{\pgfqpoint{0.841957in}{1.359250in}}%
\pgfusepath{clip}%
\pgfsetbuttcap%
\pgfsetmiterjoin%
\definecolor{currentfill}{rgb}{0.580392,0.580392,0.580392}%
\pgfsetfillcolor{currentfill}%
\pgfsetlinewidth{1.204500pt}%
\definecolor{currentstroke}{rgb}{0.188235,0.188235,0.188235}%
\pgfsetstrokecolor{currentstroke}%
\pgfsetdash{}{0pt}%
\pgfpathmoveto{\pgfqpoint{0.662200in}{0.498213in}}%
\pgfpathlineto{\pgfqpoint{0.886722in}{0.498213in}}%
\pgfpathlineto{\pgfqpoint{0.886722in}{0.755992in}}%
\pgfpathlineto{\pgfqpoint{0.662200in}{0.755992in}}%
\pgfpathlineto{\pgfqpoint{0.662200in}{0.498213in}}%
\pgfpathclose%
\pgfusepath{stroke,fill}%
\end{pgfscope}%
\begin{pgfscope}%
\pgfpathrectangle{\pgfqpoint{0.634134in}{0.187357in}}{\pgfqpoint{0.841957in}{1.359250in}}%
\pgfusepath{clip}%
\pgfsetbuttcap%
\pgfsetmiterjoin%
\definecolor{currentfill}{rgb}{0.740686,0.573039,0.431863}%
\pgfsetfillcolor{currentfill}%
\pgfsetlinewidth{1.204500pt}%
\definecolor{currentstroke}{rgb}{0.188235,0.188235,0.188235}%
\pgfsetstrokecolor{currentstroke}%
\pgfsetdash{}{0pt}%
\pgfpathmoveto{\pgfqpoint{0.942852in}{0.670879in}}%
\pgfpathlineto{\pgfqpoint{1.167374in}{0.670879in}}%
\pgfpathlineto{\pgfqpoint{1.167374in}{0.999523in}}%
\pgfpathlineto{\pgfqpoint{0.942852in}{0.999523in}}%
\pgfpathlineto{\pgfqpoint{0.942852in}{0.670879in}}%
\pgfpathclose%
\pgfusepath{stroke,fill}%
\end{pgfscope}%
\begin{pgfscope}%
\pgfpathrectangle{\pgfqpoint{0.634134in}{0.187357in}}{\pgfqpoint{0.841957in}{1.359250in}}%
\pgfusepath{clip}%
\pgfsetbuttcap%
\pgfsetmiterjoin%
\definecolor{currentfill}{rgb}{0.084314,0.543137,0.416667}%
\pgfsetfillcolor{currentfill}%
\pgfsetlinewidth{1.204500pt}%
\definecolor{currentstroke}{rgb}{0.188235,0.188235,0.188235}%
\pgfsetstrokecolor{currentstroke}%
\pgfsetdash{}{0pt}%
\pgfpathmoveto{\pgfqpoint{1.223505in}{0.654806in}}%
\pgfpathlineto{\pgfqpoint{1.448027in}{0.654806in}}%
\pgfpathlineto{\pgfqpoint{1.448027in}{0.954091in}}%
\pgfpathlineto{\pgfqpoint{1.223505in}{0.954091in}}%
\pgfpathlineto{\pgfqpoint{1.223505in}{0.654806in}}%
\pgfpathclose%
\pgfusepath{stroke,fill}%
\end{pgfscope}%
\begin{pgfscope}%
\pgfpathrectangle{\pgfqpoint{0.634134in}{0.187357in}}{\pgfqpoint{0.841957in}{1.359250in}}%
\pgfusepath{clip}%
\pgfsetroundcap%
\pgfsetroundjoin%
\pgfsetlinewidth{1.204500pt}%
\definecolor{currentstroke}{rgb}{0.188235,0.188235,0.188235}%
\pgfsetstrokecolor{currentstroke}%
\pgfsetdash{}{0pt}%
\pgfpathmoveto{\pgfqpoint{0.774461in}{0.498213in}}%
\pgfpathlineto{\pgfqpoint{0.774461in}{0.249141in}}%
\pgfusepath{stroke}%
\end{pgfscope}%
\begin{pgfscope}%
\pgfpathrectangle{\pgfqpoint{0.634134in}{0.187357in}}{\pgfqpoint{0.841957in}{1.359250in}}%
\pgfusepath{clip}%
\pgfsetroundcap%
\pgfsetroundjoin%
\pgfsetlinewidth{1.204500pt}%
\definecolor{currentstroke}{rgb}{0.188235,0.188235,0.188235}%
\pgfsetstrokecolor{currentstroke}%
\pgfsetdash{}{0pt}%
\pgfpathmoveto{\pgfqpoint{0.774461in}{0.755992in}}%
\pgfpathlineto{\pgfqpoint{0.774461in}{1.139883in}}%
\pgfusepath{stroke}%
\end{pgfscope}%
\begin{pgfscope}%
\pgfpathrectangle{\pgfqpoint{0.634134in}{0.187357in}}{\pgfqpoint{0.841957in}{1.359250in}}%
\pgfusepath{clip}%
\pgfsetroundcap%
\pgfsetroundjoin%
\pgfsetlinewidth{1.204500pt}%
\definecolor{currentstroke}{rgb}{0.188235,0.188235,0.188235}%
\pgfsetstrokecolor{currentstroke}%
\pgfsetdash{}{0pt}%
\pgfpathmoveto{\pgfqpoint{0.718330in}{0.249141in}}%
\pgfpathlineto{\pgfqpoint{0.830591in}{0.249141in}}%
\pgfusepath{stroke}%
\end{pgfscope}%
\begin{pgfscope}%
\pgfpathrectangle{\pgfqpoint{0.634134in}{0.187357in}}{\pgfqpoint{0.841957in}{1.359250in}}%
\pgfusepath{clip}%
\pgfsetroundcap%
\pgfsetroundjoin%
\pgfsetlinewidth{1.204500pt}%
\definecolor{currentstroke}{rgb}{0.188235,0.188235,0.188235}%
\pgfsetstrokecolor{currentstroke}%
\pgfsetdash{}{0pt}%
\pgfpathmoveto{\pgfqpoint{0.718330in}{1.139883in}}%
\pgfpathlineto{\pgfqpoint{0.830591in}{1.139883in}}%
\pgfusepath{stroke}%
\end{pgfscope}%
\begin{pgfscope}%
\pgfpathrectangle{\pgfqpoint{0.634134in}{0.187357in}}{\pgfqpoint{0.841957in}{1.359250in}}%
\pgfusepath{clip}%
\pgfsetroundcap%
\pgfsetroundjoin%
\pgfsetlinewidth{1.204500pt}%
\definecolor{currentstroke}{rgb}{0.188235,0.188235,0.188235}%
\pgfsetstrokecolor{currentstroke}%
\pgfsetdash{}{0pt}%
\pgfpathmoveto{\pgfqpoint{1.055113in}{0.670879in}}%
\pgfpathlineto{\pgfqpoint{1.055113in}{0.357445in}}%
\pgfusepath{stroke}%
\end{pgfscope}%
\begin{pgfscope}%
\pgfpathrectangle{\pgfqpoint{0.634134in}{0.187357in}}{\pgfqpoint{0.841957in}{1.359250in}}%
\pgfusepath{clip}%
\pgfsetroundcap%
\pgfsetroundjoin%
\pgfsetlinewidth{1.204500pt}%
\definecolor{currentstroke}{rgb}{0.188235,0.188235,0.188235}%
\pgfsetstrokecolor{currentstroke}%
\pgfsetdash{}{0pt}%
\pgfpathmoveto{\pgfqpoint{1.055113in}{0.999523in}}%
\pgfpathlineto{\pgfqpoint{1.055113in}{1.484823in}}%
\pgfusepath{stroke}%
\end{pgfscope}%
\begin{pgfscope}%
\pgfpathrectangle{\pgfqpoint{0.634134in}{0.187357in}}{\pgfqpoint{0.841957in}{1.359250in}}%
\pgfusepath{clip}%
\pgfsetroundcap%
\pgfsetroundjoin%
\pgfsetlinewidth{1.204500pt}%
\definecolor{currentstroke}{rgb}{0.188235,0.188235,0.188235}%
\pgfsetstrokecolor{currentstroke}%
\pgfsetdash{}{0pt}%
\pgfpathmoveto{\pgfqpoint{0.998983in}{0.357445in}}%
\pgfpathlineto{\pgfqpoint{1.111244in}{0.357445in}}%
\pgfusepath{stroke}%
\end{pgfscope}%
\begin{pgfscope}%
\pgfpathrectangle{\pgfqpoint{0.634134in}{0.187357in}}{\pgfqpoint{0.841957in}{1.359250in}}%
\pgfusepath{clip}%
\pgfsetroundcap%
\pgfsetroundjoin%
\pgfsetlinewidth{1.204500pt}%
\definecolor{currentstroke}{rgb}{0.188235,0.188235,0.188235}%
\pgfsetstrokecolor{currentstroke}%
\pgfsetdash{}{0pt}%
\pgfpathmoveto{\pgfqpoint{0.998983in}{1.484823in}}%
\pgfpathlineto{\pgfqpoint{1.111244in}{1.484823in}}%
\pgfusepath{stroke}%
\end{pgfscope}%
\begin{pgfscope}%
\pgfpathrectangle{\pgfqpoint{0.634134in}{0.187357in}}{\pgfqpoint{0.841957in}{1.359250in}}%
\pgfusepath{clip}%
\pgfsetroundcap%
\pgfsetroundjoin%
\pgfsetlinewidth{1.204500pt}%
\definecolor{currentstroke}{rgb}{0.188235,0.188235,0.188235}%
\pgfsetstrokecolor{currentstroke}%
\pgfsetdash{}{0pt}%
\pgfpathmoveto{\pgfqpoint{1.335766in}{0.654806in}}%
\pgfpathlineto{\pgfqpoint{1.335766in}{0.336686in}}%
\pgfusepath{stroke}%
\end{pgfscope}%
\begin{pgfscope}%
\pgfpathrectangle{\pgfqpoint{0.634134in}{0.187357in}}{\pgfqpoint{0.841957in}{1.359250in}}%
\pgfusepath{clip}%
\pgfsetroundcap%
\pgfsetroundjoin%
\pgfsetlinewidth{1.204500pt}%
\definecolor{currentstroke}{rgb}{0.188235,0.188235,0.188235}%
\pgfsetstrokecolor{currentstroke}%
\pgfsetdash{}{0pt}%
\pgfpathmoveto{\pgfqpoint{1.335766in}{0.954091in}}%
\pgfpathlineto{\pgfqpoint{1.335766in}{1.400347in}}%
\pgfusepath{stroke}%
\end{pgfscope}%
\begin{pgfscope}%
\pgfpathrectangle{\pgfqpoint{0.634134in}{0.187357in}}{\pgfqpoint{0.841957in}{1.359250in}}%
\pgfusepath{clip}%
\pgfsetroundcap%
\pgfsetroundjoin%
\pgfsetlinewidth{1.204500pt}%
\definecolor{currentstroke}{rgb}{0.188235,0.188235,0.188235}%
\pgfsetstrokecolor{currentstroke}%
\pgfsetdash{}{0pt}%
\pgfpathmoveto{\pgfqpoint{1.279635in}{0.336686in}}%
\pgfpathlineto{\pgfqpoint{1.391896in}{0.336686in}}%
\pgfusepath{stroke}%
\end{pgfscope}%
\begin{pgfscope}%
\pgfpathrectangle{\pgfqpoint{0.634134in}{0.187357in}}{\pgfqpoint{0.841957in}{1.359250in}}%
\pgfusepath{clip}%
\pgfsetroundcap%
\pgfsetroundjoin%
\pgfsetlinewidth{1.204500pt}%
\definecolor{currentstroke}{rgb}{0.188235,0.188235,0.188235}%
\pgfsetstrokecolor{currentstroke}%
\pgfsetdash{}{0pt}%
\pgfpathmoveto{\pgfqpoint{1.279635in}{1.400347in}}%
\pgfpathlineto{\pgfqpoint{1.391896in}{1.400347in}}%
\pgfusepath{stroke}%
\end{pgfscope}%
\begin{pgfscope}%
\pgfpathrectangle{\pgfqpoint{0.634134in}{0.187357in}}{\pgfqpoint{0.841957in}{1.359250in}}%
\pgfusepath{clip}%
\pgfsetbuttcap%
\pgfsetroundjoin%
\pgfsetlinewidth{1.204500pt}%
\definecolor{currentstroke}{rgb}{1.000000,0.714000,0.757000}%
\pgfsetstrokecolor{currentstroke}%
\pgfsetdash{{4.440000pt}{1.920000pt}}{0.000000pt}%
\pgfpathmoveto{\pgfqpoint{0.634134in}{0.755966in}}%
\pgfpathlineto{\pgfqpoint{1.476092in}{0.755966in}}%
\pgfusepath{stroke}%
\end{pgfscope}%
\begin{pgfscope}%
\pgfpathrectangle{\pgfqpoint{0.634134in}{0.187357in}}{\pgfqpoint{0.841957in}{1.359250in}}%
\pgfusepath{clip}%
\pgfsetroundcap%
\pgfsetroundjoin%
\pgfsetlinewidth{1.204500pt}%
\definecolor{currentstroke}{rgb}{0.188235,0.188235,0.188235}%
\pgfsetstrokecolor{currentstroke}%
\pgfsetdash{}{0pt}%
\pgfpathmoveto{\pgfqpoint{0.662200in}{0.612247in}}%
\pgfpathlineto{\pgfqpoint{0.886722in}{0.612247in}}%
\pgfusepath{stroke}%
\end{pgfscope}%
\begin{pgfscope}%
\pgfpathrectangle{\pgfqpoint{0.634134in}{0.187357in}}{\pgfqpoint{0.841957in}{1.359250in}}%
\pgfusepath{clip}%
\pgfsetroundcap%
\pgfsetroundjoin%
\pgfsetlinewidth{1.204500pt}%
\definecolor{currentstroke}{rgb}{0.188235,0.188235,0.188235}%
\pgfsetstrokecolor{currentstroke}%
\pgfsetdash{}{0pt}%
\pgfpathmoveto{\pgfqpoint{0.942852in}{0.830286in}}%
\pgfpathlineto{\pgfqpoint{1.167374in}{0.830286in}}%
\pgfusepath{stroke}%
\end{pgfscope}%
\begin{pgfscope}%
\pgfpathrectangle{\pgfqpoint{0.634134in}{0.187357in}}{\pgfqpoint{0.841957in}{1.359250in}}%
\pgfusepath{clip}%
\pgfsetroundcap%
\pgfsetroundjoin%
\pgfsetlinewidth{1.204500pt}%
\definecolor{currentstroke}{rgb}{0.188235,0.188235,0.188235}%
\pgfsetstrokecolor{currentstroke}%
\pgfsetdash{}{0pt}%
\pgfpathmoveto{\pgfqpoint{1.223505in}{0.789654in}}%
\pgfpathlineto{\pgfqpoint{1.448027in}{0.789654in}}%
\pgfusepath{stroke}%
\end{pgfscope}%
\begin{pgfscope}%
\pgfsetrectcap%
\pgfsetmiterjoin%
\pgfsetlinewidth{1.003750pt}%
\definecolor{currentstroke}{rgb}{0.000000,0.000000,0.000000}%
\pgfsetstrokecolor{currentstroke}%
\pgfsetdash{}{0pt}%
\pgfpathmoveto{\pgfqpoint{0.634134in}{0.187357in}}%
\pgfpathlineto{\pgfqpoint{0.634134in}{1.546607in}}%
\pgfusepath{stroke}%
\end{pgfscope}%
\begin{pgfscope}%
\definecolor{textcolor}{rgb}{0.150000,0.150000,0.150000}%
\pgfsetstrokecolor{textcolor}%
\pgfsetfillcolor{textcolor}%
\pgftext[x=1.055113in,y=1.629940in,,base]{\color{textcolor}\sffamily\fontsize{9.600000}{11.520000}\selectfont (a)}%
\end{pgfscope}%
\begin{pgfscope}%
\pgfsetbuttcap%
\pgfsetmiterjoin%
\definecolor{currentfill}{rgb}{1.000000,1.000000,1.000000}%
\pgfsetfillcolor{currentfill}%
\pgfsetlinewidth{0.000000pt}%
\definecolor{currentstroke}{rgb}{0.000000,0.000000,0.000000}%
\pgfsetstrokecolor{currentstroke}%
\pgfsetstrokeopacity{0.000000}%
\pgfsetdash{}{0pt}%
\pgfpathmoveto{\pgfqpoint{2.045537in}{0.187357in}}%
\pgfpathlineto{\pgfqpoint{2.887495in}{0.187357in}}%
\pgfpathlineto{\pgfqpoint{2.887495in}{1.546607in}}%
\pgfpathlineto{\pgfqpoint{2.045537in}{1.546607in}}%
\pgfpathclose%
\pgfusepath{fill}%
\end{pgfscope}%
\begin{pgfscope}%
\pgfpathrectangle{\pgfqpoint{2.045537in}{0.187357in}}{\pgfqpoint{0.841957in}{1.359250in}}%
\pgfusepath{clip}%
\pgfsetroundcap%
\pgfsetroundjoin%
\pgfsetlinewidth{0.803000pt}%
\definecolor{currentstroke}{rgb}{0.800000,0.800000,0.800000}%
\pgfsetstrokecolor{currentstroke}%
\pgfsetdash{}{0pt}%
\pgfpathmoveto{\pgfqpoint{2.045537in}{0.422306in}}%
\pgfpathlineto{\pgfqpoint{2.887495in}{0.422306in}}%
\pgfusepath{stroke}%
\end{pgfscope}%
\begin{pgfscope}%
\pgfsetbuttcap%
\pgfsetroundjoin%
\definecolor{currentfill}{rgb}{0.150000,0.150000,0.150000}%
\pgfsetfillcolor{currentfill}%
\pgfsetlinewidth{1.003750pt}%
\definecolor{currentstroke}{rgb}{0.150000,0.150000,0.150000}%
\pgfsetstrokecolor{currentstroke}%
\pgfsetdash{}{0pt}%
\pgfsys@defobject{currentmarker}{\pgfqpoint{-0.069444in}{0.000000in}}{\pgfqpoint{-0.000000in}{0.000000in}}{%
\pgfpathmoveto{\pgfqpoint{-0.000000in}{0.000000in}}%
\pgfpathlineto{\pgfqpoint{-0.069444in}{0.000000in}}%
\pgfusepath{stroke,fill}%
}%
\begin{pgfscope}%
\pgfsys@transformshift{2.045537in}{0.422306in}%
\pgfsys@useobject{currentmarker}{}%
\end{pgfscope}%
\end{pgfscope}%
\begin{pgfscope}%
\definecolor{textcolor}{rgb}{0.150000,0.150000,0.150000}%
\pgfsetstrokecolor{textcolor}%
\pgfsetfillcolor{textcolor}%
\pgftext[x=1.863246in, y=0.378903in, left, base]{\color{textcolor}\sffamily\fontsize{8.800000}{10.560000}\selectfont \(\displaystyle {0}\)}%
\end{pgfscope}%
\begin{pgfscope}%
\pgfpathrectangle{\pgfqpoint{2.045537in}{0.187357in}}{\pgfqpoint{0.841957in}{1.359250in}}%
\pgfusepath{clip}%
\pgfsetroundcap%
\pgfsetroundjoin%
\pgfsetlinewidth{0.803000pt}%
\definecolor{currentstroke}{rgb}{0.800000,0.800000,0.800000}%
\pgfsetstrokecolor{currentstroke}%
\pgfsetdash{}{0pt}%
\pgfpathmoveto{\pgfqpoint{2.045537in}{0.821728in}}%
\pgfpathlineto{\pgfqpoint{2.887495in}{0.821728in}}%
\pgfusepath{stroke}%
\end{pgfscope}%
\begin{pgfscope}%
\pgfsetbuttcap%
\pgfsetroundjoin%
\definecolor{currentfill}{rgb}{0.150000,0.150000,0.150000}%
\pgfsetfillcolor{currentfill}%
\pgfsetlinewidth{1.003750pt}%
\definecolor{currentstroke}{rgb}{0.150000,0.150000,0.150000}%
\pgfsetstrokecolor{currentstroke}%
\pgfsetdash{}{0pt}%
\pgfsys@defobject{currentmarker}{\pgfqpoint{-0.069444in}{0.000000in}}{\pgfqpoint{-0.000000in}{0.000000in}}{%
\pgfpathmoveto{\pgfqpoint{-0.000000in}{0.000000in}}%
\pgfpathlineto{\pgfqpoint{-0.069444in}{0.000000in}}%
\pgfusepath{stroke,fill}%
}%
\begin{pgfscope}%
\pgfsys@transformshift{2.045537in}{0.821728in}%
\pgfsys@useobject{currentmarker}{}%
\end{pgfscope}%
\end{pgfscope}%
\begin{pgfscope}%
\definecolor{textcolor}{rgb}{0.150000,0.150000,0.150000}%
\pgfsetstrokecolor{textcolor}%
\pgfsetfillcolor{textcolor}%
\pgftext[x=1.799010in, y=0.778325in, left, base]{\color{textcolor}\sffamily\fontsize{8.800000}{10.560000}\selectfont \(\displaystyle {10}\)}%
\end{pgfscope}%
\begin{pgfscope}%
\pgfpathrectangle{\pgfqpoint{2.045537in}{0.187357in}}{\pgfqpoint{0.841957in}{1.359250in}}%
\pgfusepath{clip}%
\pgfsetroundcap%
\pgfsetroundjoin%
\pgfsetlinewidth{0.803000pt}%
\definecolor{currentstroke}{rgb}{0.800000,0.800000,0.800000}%
\pgfsetstrokecolor{currentstroke}%
\pgfsetdash{}{0pt}%
\pgfpathmoveto{\pgfqpoint{2.045537in}{1.221150in}}%
\pgfpathlineto{\pgfqpoint{2.887495in}{1.221150in}}%
\pgfusepath{stroke}%
\end{pgfscope}%
\begin{pgfscope}%
\pgfsetbuttcap%
\pgfsetroundjoin%
\definecolor{currentfill}{rgb}{0.150000,0.150000,0.150000}%
\pgfsetfillcolor{currentfill}%
\pgfsetlinewidth{1.003750pt}%
\definecolor{currentstroke}{rgb}{0.150000,0.150000,0.150000}%
\pgfsetstrokecolor{currentstroke}%
\pgfsetdash{}{0pt}%
\pgfsys@defobject{currentmarker}{\pgfqpoint{-0.069444in}{0.000000in}}{\pgfqpoint{-0.000000in}{0.000000in}}{%
\pgfpathmoveto{\pgfqpoint{-0.000000in}{0.000000in}}%
\pgfpathlineto{\pgfqpoint{-0.069444in}{0.000000in}}%
\pgfusepath{stroke,fill}%
}%
\begin{pgfscope}%
\pgfsys@transformshift{2.045537in}{1.221150in}%
\pgfsys@useobject{currentmarker}{}%
\end{pgfscope}%
\end{pgfscope}%
\begin{pgfscope}%
\definecolor{textcolor}{rgb}{0.150000,0.150000,0.150000}%
\pgfsetstrokecolor{textcolor}%
\pgfsetfillcolor{textcolor}%
\pgftext[x=1.799010in, y=1.177747in, left, base]{\color{textcolor}\sffamily\fontsize{8.800000}{10.560000}\selectfont \(\displaystyle {20}\)}%
\end{pgfscope}%
\begin{pgfscope}%
\definecolor{textcolor}{rgb}{0.150000,0.150000,0.150000}%
\pgfsetstrokecolor{textcolor}%
\pgfsetfillcolor{textcolor}%
\pgftext[x=1.743455in,y=0.866982in,,bottom,rotate=90.000000]{\color{textcolor}\sffamily\fontsize{9.600000}{11.520000}\selectfont \(\displaystyle \mathrm{pcmi}_h\)}%
\end{pgfscope}%
\begin{pgfscope}%
\pgfpathrectangle{\pgfqpoint{2.045537in}{0.187357in}}{\pgfqpoint{0.841957in}{1.359250in}}%
\pgfusepath{clip}%
\pgfsetbuttcap%
\pgfsetmiterjoin%
\definecolor{currentfill}{rgb}{0.580392,0.580392,0.580392}%
\pgfsetfillcolor{currentfill}%
\pgfsetlinewidth{1.204500pt}%
\definecolor{currentstroke}{rgb}{0.188235,0.188235,0.188235}%
\pgfsetstrokecolor{currentstroke}%
\pgfsetdash{}{0pt}%
\pgfpathmoveto{\pgfqpoint{2.073603in}{0.574979in}}%
\pgfpathlineto{\pgfqpoint{2.298125in}{0.574979in}}%
\pgfpathlineto{\pgfqpoint{2.298125in}{0.845746in}}%
\pgfpathlineto{\pgfqpoint{2.073603in}{0.845746in}}%
\pgfpathlineto{\pgfqpoint{2.073603in}{0.574979in}}%
\pgfpathclose%
\pgfusepath{stroke,fill}%
\end{pgfscope}%
\begin{pgfscope}%
\pgfpathrectangle{\pgfqpoint{2.045537in}{0.187357in}}{\pgfqpoint{0.841957in}{1.359250in}}%
\pgfusepath{clip}%
\pgfsetbuttcap%
\pgfsetmiterjoin%
\definecolor{currentfill}{rgb}{0.740686,0.573039,0.431863}%
\pgfsetfillcolor{currentfill}%
\pgfsetlinewidth{1.204500pt}%
\definecolor{currentstroke}{rgb}{0.188235,0.188235,0.188235}%
\pgfsetstrokecolor{currentstroke}%
\pgfsetdash{}{0pt}%
\pgfpathmoveto{\pgfqpoint{2.354255in}{0.609281in}}%
\pgfpathlineto{\pgfqpoint{2.578777in}{0.609281in}}%
\pgfpathlineto{\pgfqpoint{2.578777in}{0.949999in}}%
\pgfpathlineto{\pgfqpoint{2.354255in}{0.949999in}}%
\pgfpathlineto{\pgfqpoint{2.354255in}{0.609281in}}%
\pgfpathclose%
\pgfusepath{stroke,fill}%
\end{pgfscope}%
\begin{pgfscope}%
\pgfpathrectangle{\pgfqpoint{2.045537in}{0.187357in}}{\pgfqpoint{0.841957in}{1.359250in}}%
\pgfusepath{clip}%
\pgfsetbuttcap%
\pgfsetmiterjoin%
\definecolor{currentfill}{rgb}{0.084314,0.543137,0.416667}%
\pgfsetfillcolor{currentfill}%
\pgfsetlinewidth{1.204500pt}%
\definecolor{currentstroke}{rgb}{0.188235,0.188235,0.188235}%
\pgfsetstrokecolor{currentstroke}%
\pgfsetdash{}{0pt}%
\pgfpathmoveto{\pgfqpoint{2.634907in}{0.688506in}}%
\pgfpathlineto{\pgfqpoint{2.859429in}{0.688506in}}%
\pgfpathlineto{\pgfqpoint{2.859429in}{1.012553in}}%
\pgfpathlineto{\pgfqpoint{2.634907in}{1.012553in}}%
\pgfpathlineto{\pgfqpoint{2.634907in}{0.688506in}}%
\pgfpathclose%
\pgfusepath{stroke,fill}%
\end{pgfscope}%
\begin{pgfscope}%
\pgfpathrectangle{\pgfqpoint{2.045537in}{0.187357in}}{\pgfqpoint{0.841957in}{1.359250in}}%
\pgfusepath{clip}%
\pgfsetroundcap%
\pgfsetroundjoin%
\pgfsetlinewidth{1.204500pt}%
\definecolor{currentstroke}{rgb}{0.188235,0.188235,0.188235}%
\pgfsetstrokecolor{currentstroke}%
\pgfsetdash{}{0pt}%
\pgfpathmoveto{\pgfqpoint{2.185864in}{0.574979in}}%
\pgfpathlineto{\pgfqpoint{2.185864in}{0.249141in}}%
\pgfusepath{stroke}%
\end{pgfscope}%
\begin{pgfscope}%
\pgfpathrectangle{\pgfqpoint{2.045537in}{0.187357in}}{\pgfqpoint{0.841957in}{1.359250in}}%
\pgfusepath{clip}%
\pgfsetroundcap%
\pgfsetroundjoin%
\pgfsetlinewidth{1.204500pt}%
\definecolor{currentstroke}{rgb}{0.188235,0.188235,0.188235}%
\pgfsetstrokecolor{currentstroke}%
\pgfsetdash{}{0pt}%
\pgfpathmoveto{\pgfqpoint{2.185864in}{0.845746in}}%
\pgfpathlineto{\pgfqpoint{2.185864in}{1.250988in}}%
\pgfusepath{stroke}%
\end{pgfscope}%
\begin{pgfscope}%
\pgfpathrectangle{\pgfqpoint{2.045537in}{0.187357in}}{\pgfqpoint{0.841957in}{1.359250in}}%
\pgfusepath{clip}%
\pgfsetroundcap%
\pgfsetroundjoin%
\pgfsetlinewidth{1.204500pt}%
\definecolor{currentstroke}{rgb}{0.188235,0.188235,0.188235}%
\pgfsetstrokecolor{currentstroke}%
\pgfsetdash{}{0pt}%
\pgfpathmoveto{\pgfqpoint{2.129733in}{0.249141in}}%
\pgfpathlineto{\pgfqpoint{2.241994in}{0.249141in}}%
\pgfusepath{stroke}%
\end{pgfscope}%
\begin{pgfscope}%
\pgfpathrectangle{\pgfqpoint{2.045537in}{0.187357in}}{\pgfqpoint{0.841957in}{1.359250in}}%
\pgfusepath{clip}%
\pgfsetroundcap%
\pgfsetroundjoin%
\pgfsetlinewidth{1.204500pt}%
\definecolor{currentstroke}{rgb}{0.188235,0.188235,0.188235}%
\pgfsetstrokecolor{currentstroke}%
\pgfsetdash{}{0pt}%
\pgfpathmoveto{\pgfqpoint{2.129733in}{1.250988in}}%
\pgfpathlineto{\pgfqpoint{2.241994in}{1.250988in}}%
\pgfusepath{stroke}%
\end{pgfscope}%
\begin{pgfscope}%
\pgfpathrectangle{\pgfqpoint{2.045537in}{0.187357in}}{\pgfqpoint{0.841957in}{1.359250in}}%
\pgfusepath{clip}%
\pgfsetroundcap%
\pgfsetroundjoin%
\pgfsetlinewidth{1.204500pt}%
\definecolor{currentstroke}{rgb}{0.188235,0.188235,0.188235}%
\pgfsetstrokecolor{currentstroke}%
\pgfsetdash{}{0pt}%
\pgfpathmoveto{\pgfqpoint{2.466516in}{0.609281in}}%
\pgfpathlineto{\pgfqpoint{2.466516in}{0.299434in}}%
\pgfusepath{stroke}%
\end{pgfscope}%
\begin{pgfscope}%
\pgfpathrectangle{\pgfqpoint{2.045537in}{0.187357in}}{\pgfqpoint{0.841957in}{1.359250in}}%
\pgfusepath{clip}%
\pgfsetroundcap%
\pgfsetroundjoin%
\pgfsetlinewidth{1.204500pt}%
\definecolor{currentstroke}{rgb}{0.188235,0.188235,0.188235}%
\pgfsetstrokecolor{currentstroke}%
\pgfsetdash{}{0pt}%
\pgfpathmoveto{\pgfqpoint{2.466516in}{0.949999in}}%
\pgfpathlineto{\pgfqpoint{2.466516in}{1.439695in}}%
\pgfusepath{stroke}%
\end{pgfscope}%
\begin{pgfscope}%
\pgfpathrectangle{\pgfqpoint{2.045537in}{0.187357in}}{\pgfqpoint{0.841957in}{1.359250in}}%
\pgfusepath{clip}%
\pgfsetroundcap%
\pgfsetroundjoin%
\pgfsetlinewidth{1.204500pt}%
\definecolor{currentstroke}{rgb}{0.188235,0.188235,0.188235}%
\pgfsetstrokecolor{currentstroke}%
\pgfsetdash{}{0pt}%
\pgfpathmoveto{\pgfqpoint{2.410385in}{0.299434in}}%
\pgfpathlineto{\pgfqpoint{2.522646in}{0.299434in}}%
\pgfusepath{stroke}%
\end{pgfscope}%
\begin{pgfscope}%
\pgfpathrectangle{\pgfqpoint{2.045537in}{0.187357in}}{\pgfqpoint{0.841957in}{1.359250in}}%
\pgfusepath{clip}%
\pgfsetroundcap%
\pgfsetroundjoin%
\pgfsetlinewidth{1.204500pt}%
\definecolor{currentstroke}{rgb}{0.188235,0.188235,0.188235}%
\pgfsetstrokecolor{currentstroke}%
\pgfsetdash{}{0pt}%
\pgfpathmoveto{\pgfqpoint{2.410385in}{1.439695in}}%
\pgfpathlineto{\pgfqpoint{2.522646in}{1.439695in}}%
\pgfusepath{stroke}%
\end{pgfscope}%
\begin{pgfscope}%
\pgfpathrectangle{\pgfqpoint{2.045537in}{0.187357in}}{\pgfqpoint{0.841957in}{1.359250in}}%
\pgfusepath{clip}%
\pgfsetroundcap%
\pgfsetroundjoin%
\pgfsetlinewidth{1.204500pt}%
\definecolor{currentstroke}{rgb}{0.188235,0.188235,0.188235}%
\pgfsetstrokecolor{currentstroke}%
\pgfsetdash{}{0pt}%
\pgfpathmoveto{\pgfqpoint{2.747168in}{0.688506in}}%
\pgfpathlineto{\pgfqpoint{2.747168in}{0.432101in}}%
\pgfusepath{stroke}%
\end{pgfscope}%
\begin{pgfscope}%
\pgfpathrectangle{\pgfqpoint{2.045537in}{0.187357in}}{\pgfqpoint{0.841957in}{1.359250in}}%
\pgfusepath{clip}%
\pgfsetroundcap%
\pgfsetroundjoin%
\pgfsetlinewidth{1.204500pt}%
\definecolor{currentstroke}{rgb}{0.188235,0.188235,0.188235}%
\pgfsetstrokecolor{currentstroke}%
\pgfsetdash{}{0pt}%
\pgfpathmoveto{\pgfqpoint{2.747168in}{1.012553in}}%
\pgfpathlineto{\pgfqpoint{2.747168in}{1.484823in}}%
\pgfusepath{stroke}%
\end{pgfscope}%
\begin{pgfscope}%
\pgfpathrectangle{\pgfqpoint{2.045537in}{0.187357in}}{\pgfqpoint{0.841957in}{1.359250in}}%
\pgfusepath{clip}%
\pgfsetroundcap%
\pgfsetroundjoin%
\pgfsetlinewidth{1.204500pt}%
\definecolor{currentstroke}{rgb}{0.188235,0.188235,0.188235}%
\pgfsetstrokecolor{currentstroke}%
\pgfsetdash{}{0pt}%
\pgfpathmoveto{\pgfqpoint{2.691038in}{0.432101in}}%
\pgfpathlineto{\pgfqpoint{2.803299in}{0.432101in}}%
\pgfusepath{stroke}%
\end{pgfscope}%
\begin{pgfscope}%
\pgfpathrectangle{\pgfqpoint{2.045537in}{0.187357in}}{\pgfqpoint{0.841957in}{1.359250in}}%
\pgfusepath{clip}%
\pgfsetroundcap%
\pgfsetroundjoin%
\pgfsetlinewidth{1.204500pt}%
\definecolor{currentstroke}{rgb}{0.188235,0.188235,0.188235}%
\pgfsetstrokecolor{currentstroke}%
\pgfsetdash{}{0pt}%
\pgfpathmoveto{\pgfqpoint{2.691038in}{1.484823in}}%
\pgfpathlineto{\pgfqpoint{2.803299in}{1.484823in}}%
\pgfusepath{stroke}%
\end{pgfscope}%
\begin{pgfscope}%
\pgfpathrectangle{\pgfqpoint{2.045537in}{0.187357in}}{\pgfqpoint{0.841957in}{1.359250in}}%
\pgfusepath{clip}%
\pgfsetbuttcap%
\pgfsetroundjoin%
\pgfsetlinewidth{1.204500pt}%
\definecolor{currentstroke}{rgb}{1.000000,0.714000,0.757000}%
\pgfsetstrokecolor{currentstroke}%
\pgfsetdash{{4.440000pt}{1.920000pt}}{0.000000pt}%
\pgfpathmoveto{\pgfqpoint{2.045537in}{0.845693in}}%
\pgfpathlineto{\pgfqpoint{2.887495in}{0.845693in}}%
\pgfusepath{stroke}%
\end{pgfscope}%
\begin{pgfscope}%
\pgfpathrectangle{\pgfqpoint{2.045537in}{0.187357in}}{\pgfqpoint{0.841957in}{1.359250in}}%
\pgfusepath{clip}%
\pgfsetroundcap%
\pgfsetroundjoin%
\pgfsetlinewidth{1.204500pt}%
\definecolor{currentstroke}{rgb}{0.188235,0.188235,0.188235}%
\pgfsetstrokecolor{currentstroke}%
\pgfsetdash{}{0pt}%
\pgfpathmoveto{\pgfqpoint{2.073603in}{0.688248in}}%
\pgfpathlineto{\pgfqpoint{2.298125in}{0.688248in}}%
\pgfusepath{stroke}%
\end{pgfscope}%
\begin{pgfscope}%
\pgfpathrectangle{\pgfqpoint{2.045537in}{0.187357in}}{\pgfqpoint{0.841957in}{1.359250in}}%
\pgfusepath{clip}%
\pgfsetroundcap%
\pgfsetroundjoin%
\pgfsetlinewidth{1.204500pt}%
\definecolor{currentstroke}{rgb}{0.188235,0.188235,0.188235}%
\pgfsetstrokecolor{currentstroke}%
\pgfsetdash{}{0pt}%
\pgfpathmoveto{\pgfqpoint{2.354255in}{0.763026in}}%
\pgfpathlineto{\pgfqpoint{2.578777in}{0.763026in}}%
\pgfusepath{stroke}%
\end{pgfscope}%
\begin{pgfscope}%
\pgfpathrectangle{\pgfqpoint{2.045537in}{0.187357in}}{\pgfqpoint{0.841957in}{1.359250in}}%
\pgfusepath{clip}%
\pgfsetroundcap%
\pgfsetroundjoin%
\pgfsetlinewidth{1.204500pt}%
\definecolor{currentstroke}{rgb}{0.188235,0.188235,0.188235}%
\pgfsetstrokecolor{currentstroke}%
\pgfsetdash{}{0pt}%
\pgfpathmoveto{\pgfqpoint{2.634907in}{0.844332in}}%
\pgfpathlineto{\pgfqpoint{2.859429in}{0.844332in}}%
\pgfusepath{stroke}%
\end{pgfscope}%
\begin{pgfscope}%
\pgfsetrectcap%
\pgfsetmiterjoin%
\pgfsetlinewidth{1.003750pt}%
\definecolor{currentstroke}{rgb}{0.000000,0.000000,0.000000}%
\pgfsetstrokecolor{currentstroke}%
\pgfsetdash{}{0pt}%
\pgfpathmoveto{\pgfqpoint{2.045537in}{0.187357in}}%
\pgfpathlineto{\pgfqpoint{2.045537in}{1.546607in}}%
\pgfusepath{stroke}%
\end{pgfscope}%
\begin{pgfscope}%
\definecolor{textcolor}{rgb}{0.150000,0.150000,0.150000}%
\pgfsetstrokecolor{textcolor}%
\pgfsetfillcolor{textcolor}%
\pgftext[x=2.466516in,y=1.629940in,,base]{\color{textcolor}\sffamily\fontsize{9.600000}{11.520000}\selectfont (b)}%
\end{pgfscope}%
\begin{pgfscope}%
\pgfsetbuttcap%
\pgfsetmiterjoin%
\definecolor{currentfill}{rgb}{1.000000,1.000000,1.000000}%
\pgfsetfillcolor{currentfill}%
\pgfsetfillopacity{0.800000}%
\pgfsetlinewidth{0.803000pt}%
\definecolor{currentstroke}{rgb}{0.800000,0.800000,0.800000}%
\pgfsetstrokecolor{currentstroke}%
\pgfsetstrokeopacity{0.800000}%
\pgfsetdash{}{0pt}%
\pgfpathmoveto{\pgfqpoint{0.476946in}{0.000000in}}%
\pgfpathlineto{\pgfqpoint{2.922457in}{0.000000in}}%
\pgfpathquadraticcurveto{\pgfqpoint{2.944680in}{0.000000in}}{\pgfqpoint{2.944680in}{0.022222in}}%
\pgfpathlineto{\pgfqpoint{2.944680in}{0.166049in}}%
\pgfpathquadraticcurveto{\pgfqpoint{2.944680in}{0.188272in}}{\pgfqpoint{2.922457in}{0.188272in}}%
\pgfpathlineto{\pgfqpoint{0.476946in}{0.188272in}}%
\pgfpathquadraticcurveto{\pgfqpoint{0.454724in}{0.188272in}}{\pgfqpoint{0.454724in}{0.166049in}}%
\pgfpathlineto{\pgfqpoint{0.454724in}{0.022222in}}%
\pgfpathquadraticcurveto{\pgfqpoint{0.454724in}{0.000000in}}{\pgfqpoint{0.476946in}{0.000000in}}%
\pgfpathclose%
\pgfusepath{stroke,fill}%
\end{pgfscope}%
\begin{pgfscope}%
\pgfsetbuttcap%
\pgfsetmiterjoin%
\definecolor{currentfill}{rgb}{0.580392,0.580392,0.580392}%
\pgfsetfillcolor{currentfill}%
\pgfsetlinewidth{0.803000pt}%
\definecolor{currentstroke}{rgb}{0.580392,0.580392,0.580392}%
\pgfsetstrokecolor{currentstroke}%
\pgfsetdash{}{0pt}%
\pgfpathmoveto{\pgfqpoint{0.499169in}{0.066049in}}%
\pgfpathlineto{\pgfqpoint{0.610280in}{0.066049in}}%
\pgfpathlineto{\pgfqpoint{0.610280in}{0.143827in}}%
\pgfpathlineto{\pgfqpoint{0.499169in}{0.143827in}}%
\pgfpathclose%
\pgfusepath{stroke,fill}%
\end{pgfscope}%
\begin{pgfscope}%
\definecolor{textcolor}{rgb}{0.150000,0.150000,0.150000}%
\pgfsetstrokecolor{textcolor}%
\pgfsetfillcolor{textcolor}%
\pgftext[x=0.643613in,y=0.066049in,left,base]{\color{textcolor}\sffamily\fontsize{8.000000}{9.600000}\selectfont All candidates}%
\end{pgfscope}%
\begin{pgfscope}%
\pgfsetbuttcap%
\pgfsetmiterjoin%
\definecolor{currentfill}{rgb}{0.792157,0.568627,0.380392}%
\pgfsetfillcolor{currentfill}%
\pgfsetlinewidth{0.803000pt}%
\definecolor{currentstroke}{rgb}{0.792157,0.568627,0.380392}%
\pgfsetstrokecolor{currentstroke}%
\pgfsetdash{}{0pt}%
\pgfpathmoveto{\pgfqpoint{1.413067in}{0.066049in}}%
\pgfpathlineto{\pgfqpoint{1.524178in}{0.066049in}}%
\pgfpathlineto{\pgfqpoint{1.524178in}{0.143827in}}%
\pgfpathlineto{\pgfqpoint{1.413067in}{0.143827in}}%
\pgfpathclose%
\pgfusepath{stroke,fill}%
\end{pgfscope}%
\begin{pgfscope}%
\definecolor{textcolor}{rgb}{0.150000,0.150000,0.150000}%
\pgfsetstrokecolor{textcolor}%
\pgfsetfillcolor{textcolor}%
\pgftext[x=1.557511in,y=0.066049in,left,base]{\color{textcolor}\sffamily\fontsize{8.000000}{9.600000}\selectfont Max. PMI}%
\end{pgfscope}%
\begin{pgfscope}%
\pgfsetbuttcap%
\pgfsetmiterjoin%
\definecolor{currentfill}{rgb}{0.007843,0.619608,0.450980}%
\pgfsetfillcolor{currentfill}%
\pgfsetlinewidth{0.803000pt}%
\definecolor{currentstroke}{rgb}{0.007843,0.619608,0.450980}%
\pgfsetstrokecolor{currentstroke}%
\pgfsetdash{}{0pt}%
\pgfpathmoveto{\pgfqpoint{2.145906in}{0.066049in}}%
\pgfpathlineto{\pgfqpoint{2.257017in}{0.066049in}}%
\pgfpathlineto{\pgfqpoint{2.257017in}{0.143827in}}%
\pgfpathlineto{\pgfqpoint{2.145906in}{0.143827in}}%
\pgfpathclose%
\pgfusepath{stroke,fill}%
\end{pgfscope}%
\begin{pgfscope}%
\definecolor{textcolor}{rgb}{0.150000,0.150000,0.150000}%
\pgfsetstrokecolor{textcolor}%
\pgfsetfillcolor{textcolor}%
\pgftext[x=2.290350in,y=0.066049in,left,base]{\color{textcolor}\sffamily\fontsize{8.000000}{9.600000}\selectfont Fused-PCMI}%
\end{pgfscope}%
\end{pgfpicture}%
\makeatother%
\endgroup%